\documentclass{mindlab}

\usepackage{graphicx}
\usepackage{multirow}
\usepackage{amsmath,amssymb,amsfonts}
\usepackage{amsthm}
\usepackage{mathrsfs}
\usepackage{xcolor}
\usepackage{microtype}
\usepackage{textcomp}
\usepackage{booktabs}
\usepackage{array}
\usepackage{url}
\usepackage{listings}
\usepackage{placeins}
\usepackage{subcaption}
\usepackage{tikz}
\usepackage{hyperref}
\usepackage[most]{tcolorbox}
\usepackage{fontawesome5}
\usepackage[smartEllipses]{markdown}
\usepackage{pifont}
\usepackage{xcolor}
\usepackage{bbm}
\usepackage[table]{xcolor} % For cell coloring
\usepackage{multirow}   % For multi-row cells
\usepackage{subcaption}
\usepackage{wrapfig}
\usepackage{xspace}

\usepackage[most]{tcolorbox}
\usepackage{wrapfig}
\usepackage{multirow}
\usepackage{graphicx,xcolor,float}
\usepackage{threeparttable}
\usepackage[ruled,vlined]{algorithm2e}
\usepackage{colortbl}
\usepackage{color}
\usepackage{multirow}
\usepackage{tabularx}
\usepackage{float}
\usepackage{amsmath, amssymb}
\usepackage{tikz}
\usetikzlibrary{shapes.geometric, arrows.meta, positioning, calc, fit, backgrounds}
\usepackage{graphicx}
\usepackage{booktabs}
\usepackage{arydshln}
\usepackage{enumitem}
\usepackage{wrapfig}
\usepackage{caption}
\usepackage{graphicx}
\usepackage{makecell}
\usepackage{float} 
\usepackage{soul}
\usepackage{mathtools}
\usepackage{bbding}
\usepackage{makecell}
\usepackage{tabularx}
\usepackage{amssymb,mathrsfs,amsmath}
\usepackage{pifont}
\usepackage{xcolor}
\usepackage{bbm}
\usepackage[table]{xcolor} % For cell coloring
\usepackage{multirow}
\newcolumntype{L}[1]{>{\raggedright\arraybackslash}p{#1}}
\newcolumntype{C}[1]{>{\centering\arraybackslash}p{#1}}
\newcolumntype{M}[1]{>{\centering\arraybackslash}m{#1}}
\providecommand{\apphead}{\rowcolor{mindlabbluepale!45}}
\providecommand{\appkey}[1]{{\sffamily\bfseries\color{mindlabblue}#1}}
\providecommand{\appgroup}[2]{\addlinespace[2pt]\rowcolor{mindlabbluepale!45}\multicolumn{#1}{@{}l}{\sffamily\bfseries\color{mindlabfg}#2}\\}

\providecommand{\pony}[1]{#1}
\providecommand{\fittowidth}[1]{%
  \sbox0{#1}%
  \ifdim\wd0>\textwidth
    \resizebox{\textwidth}{!}{\usebox0}%
  \else
    \usebox0%
  \fi
}

\begin{document}

\title{On the Scaling of PEFT: Towards Million Personal Models of Trillion Parameters}

\author{Mind Lab}
\correspondence{\email{contact@mindlab.ltd}}
\date{May 2026}

\abstract{
\pony{Parameter-efficient fine-tuning (PEFT) is usually evaluated as a cheaper alternative to full fine-tuning. This paper studies a broader possibility: whether small trainable adapters can serve as persistent local state on top of strong shared foundation models. In this view, a base model supplies common competence, while adapters may carry part of an instance-specific behavioral state, such as preferences, skills, tool habits, or memory-like updates. This framing is deliberately bounded: PEFT does not store the whole person or replace retrieval, but it may provide a compact unit of adaptive state that can be trained, evaluated, served, and composed at population scale.}

\pony{We study this possibility through three coupled scaling problems that must reinforce one another. \textbf{Scale Up} asks whether a stronger shared base model makes small local updates more useful. We study large-prior LoRA reinforcement learning, routing-aware correction, and training--serving consistency at trillion-scale MoE. \textbf{Scale Down} asks how small the local adaptive state can become while still learning reliably. We analyze rank regimes, low-rank instability, RL-native initialization, hyperparameter transfer, and memory-oriented adapter designs such as $\delta$-mem. \textbf{Scale Out} asks what becomes possible when many persistent adapted instances coexist. We examine LoRA memory capacity, context-to-parameter learning, per-user adapter simulation, and diversity-based majority voting. MinT \citep{mindlab2026mint} provides one concrete infrastructure example for supporting all three axes: adapter identity, policy revision, training provenance, evaluation, and serving residency are the mechanisms that let large shared priors, small local adapters, and large adapter populations coexist.}

\pony{Taken together, these results suggest that PEFT is more than a budget-conscious substitute for full fine-tuning. A personal model built on a strong shared prior can preserve continuity across repeated interaction, serve as a stable user simulator for agents that treat the user as part of their environment, and contribute to collective performance through diversity-based aggregation. The broader ambition is a world where strong foundation models support not one universal assistant, but millions of persistent personal ones.}
}

\maketitle

\section{Introduction}\label{sec:introduction}

\begin{figure}[t]
\centering
\includegraphics[width=\textwidth, page=5]{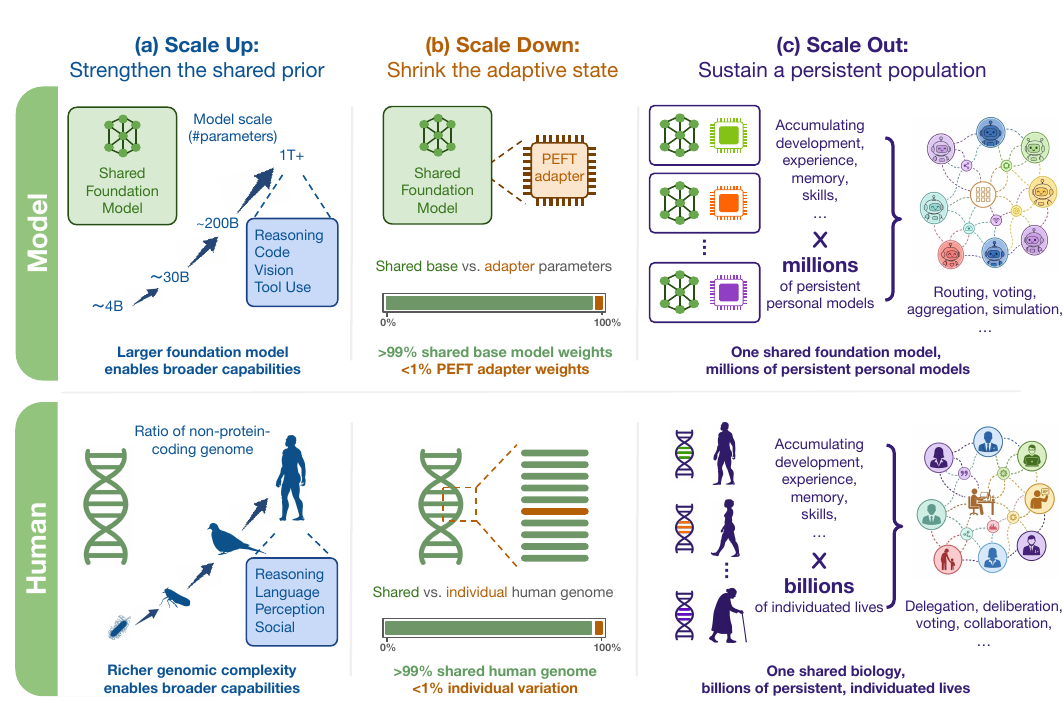}
\caption{\pony{The three scaling axes of PEFT, illustrated through a biological analogy. \textbf{(a) Scale Up.} As organisms grow more complex up the tree of life, the fraction of regulatory DNA increases, from roughly 10\% in simple organisms to 98\% in humans \citep{encode2012}, enabling richer reasoning, perception, and social behavior. Analogously, larger foundation models unlock capabilities such as reasoning, code, vision, and tool use that make small local updates more powerful. \textbf{(b) Scale Down.} Human individuals share more than 99\% of their genome \citep{1000genomes2012}; the differences that make each person distinct amount to less than 1\% of the total. A PEFT adapter occupies a similarly small fraction of the base model weights, and carries the local adaptive state that distinguishes one personal model from another. \textbf{(c) Scale Out.} One shared biology supports billions of persistent, individuated human lives, each accumulating its own development, experience, memory, and skills. One shared foundation model can support millions of persistent personal model instances in the same way, each shaped by its own history and each a member of a larger population.}}
\label{fig:peft-perspective-teaser}
\end{figure}

\pony{Frontier models can now write production code, operate tools, and reason across long contexts \citep{openai2025gpt45,glmteam2026glm5,kimiteam2026k2,qwenteam2025qwen35,anthropic2025claude47}. Agentic systems built on these models resolve real-world software engineering tasks autonomously \citep{anthropic2025claudecode,wang2024openhands,jimenez2024swebench}. But a capable assistant is not automatically a personal one. It may answer more questions and call more tools, and still fail to preserve continuity with one person over time. Long context, retrieval \citep{lewis2021rag}, prompts \citep{ji2025contextengineering}, and user profiles \citep{li2024personalagents} all help, but are not enough by themselves. A personal model needs state that can persist, adapt, and shape future behavior \citep{yao2025secondhalf,silver2025experience}.}

\pony{This paper argues that parameter-efficient fine-tuning (PEFT), especially LoRA \citep{hu2021lora}, is a practical way to represent part of that state. The goal is not PEFT itself. The goal is a persistent personal model instance: a system built from a shared base model, local adaptive state, context, tools, and external memory. The base model provides broad intelligence. The adapter carries part of the learned consequences of repeated experience, such as preferences, skills, tool habits, and some memory-like behavior. Raw facts, episodes, and documents can still live in retrieval systems or external memory \citep{packer2024memgpt,zhong2024memorybank}. The adapter is not the whole memory system.}

\pony{The biological analogy in Figure~\ref{fig:peft-perspective-teaser} frames the architecture. Any two humans share approximately 99.9\% of their genome \citep{1000genomes2012}. This small fraction of variation is sufficient to produce the full range of human individuality. One shared biology supports billions of persistent, individuated lives, each accumulating its own experience, and population diversity is itself a resource. The development of foundation models may follow a similar trajectory: shared priors growing stronger, local adaptive states remaining small, and populations of persistent personal instances becoming the unit of scale.}

\pony{We organize the technical path around three coupled scaling problems. \textbf{Scale Up} asks how to make strong shared base models repeatedly adaptable. \textbf{Scale Down} asks how small the local adaptive state can be while still carrying meaningful individuality. \textbf{Scale Out} asks what becomes possible when many persistent adapted instances coexist. The three axes are dependencies, not independent categories. Scale Up without Scale Down produces powerful priors too expensive to adapt continuously. Scale Down without Scale Up produces cheap but low-leverage adapters. Scale Out without both produces many weak or disposable variants rather than durable personal models. Section~\ref{sec:three-axes} develops this dependency structure, and Sections~\ref{sec:scale-up}--\ref{sec:infrastructure} provide the technical evidence for each axis.}

\pony{The downstream implications of personal models extend in three directions: preserving continuity for individuals, improving user simulation for agents that treat the user as part of their environment \citep{park2023generative,oasis2024}, and enabling collective intelligence through diversity among adapted models. These are directions rather than solved problems. What PEFT contributes is a practical unit of persistent individuality that is small enough to scale.}

\pony{These claims require calibration. PEFT does not store the whole person, replace retrieval, solve personal memory by itself \citep{packer2024memgpt}, reproduce society, or guarantee creativity. It does something narrower but important: it makes a portion of individuality local, persistent, efficient, and manageable. A small adapter can change without rewriting the shared base model, can be updated repeatedly at lower cost than a full checkpoint \citep{biderman2024loraforgets}, and can be named, evaluated, served, rolled back, or retired as part of a lifecycle \citep{mindlab2026mint}.}

\pony{This paper makes the following contributions. (1) We introduce a three-axis framework (Scale Up, Scale Down, Scale Out) for understanding PEFT as a scaling mechanism for personal models, and show that the axes form a dependency chain rather than a loose taxonomy. (2) We demonstrate trillion-scale LoRA RL on a 1T-parameter MoE model and identify scale-induced failure modes including training--inference mismatch in sparse architectures. (3) We characterize LoRA rank regimes under RL fine-tuning, propose OLoRA-tail as an RL-native initialization, and establish hyperparameter transfer rules across ranks. (4) We measure LoRA memory capacity laws, introduce Context Learning as a write policy for personal adapters, and show that per-user LoRA adapters produce richer social simulation structure than shared-base agents, and demonstrate in a controlled model-count experiment that diversity among distinct LoRA variants produces collective intelligence under majority voting, raising AIME24 accuracy from $0.3644$ at $k=1$ to $0.4867$ at $k=198$.}

\pony{The spine of the paper is therefore simple: PEFT makes it possible to scale from one shared foundation model to many persistent personal model instances. The near-term path is algorithmic and infrastructural. The long-term promise is personal service, better user simulation, and useful diversity among adapted models.}

\section{The Three Scaling Axes of PEFT}\label{sec:three-axes}

\pony{PEFT scales along three coupled axes because a personal model instance is both a learning problem and a systems object. Scale Up asks how strong the shared base model must be before small local updates become high leverage. Scale Down asks how small, stable, and repeatedly writable the local adaptive state can become. Scale Out asks what becomes possible when many persistent adapted instances can be trained, served, evaluated, and governed over time.}

\begin{enumerate}
    \item \pony{\textbf{Scale Up}: a stronger shared base model gives each small adapter more latent capability to redirect.}
    \item \pony{\textbf{Scale Down}: a smaller and stabler adaptive state lowers the marginal cost of repeated learning.}
    \item \pony{\textbf{Scale Out}: low marginal cost allows personalization to expand from isolated adapters to populations of persistent model instances.}
\end{enumerate}

\pony{The dependencies also define failure cases. Scale Up without Scale Down produces powerful priors that remain too expensive to adapt continuously. Scale Down without Scale Up produces cheap adapters with limited capability to specialize. Scale Out without both produces many weak or disposable variants rather than durable personal models. The thesis of this paper is that PEFT becomes transformative only when the three axes reinforce one another.}

\pony{The axes support the three visions in sequence. For individuals, they make repeated personalization plausible: a strong base model supplies broad competence, a local adapter stores part of the learned behavioral state, and lifecycle infrastructure keeps the instance persistent. For user simulation and agent environments, they make it possible for simulated users to preserve stable preferences, goals, memories, and constraints across interactions. For collective intelligence, they make diversity among adapted policies measurable and usable through voting, routing, debate, or distillation.}

\pony{Long-term memory is where the three axes meet most clearly. Context and retrieval remain essential, but a personal model also needs some learned state that persists beyond the current prompt. The adapter should not become a raw archive of the user's history. Instead, it should carry part of the behavioral consequences of repeated experience, while editable facts and documents remain in external memory. This distinction keeps the claim precise: PEFT is a mechanism for local adaptive state, not a complete memory system.}

\pony{The axes also correspond to three kinds of evidence used in this manuscript. Scale Up is grounded in trillion-parameter LoRA RL and the infrastructure needed to make large priors trainable. Scale Down is grounded in LoRA rank, initialization, and hyperparameter studies that ask how little trainable state can still learn reliably. Scale Out is grounded in memory, agent, user-simulation, and aggregation settings where the number and diversity of persistent adapted instances become objects of scaling.}

\begin{table}[tbp]
\centering
\small
\setlength{\tabcolsep}{6pt}
\renewcommand{\arraystretch}{1.20}
\caption{The three scaling axes of PEFT in the latest thesis outline.}
\label{tab:axes}
\fittowidth{%
\begin{tabular}{@{}M{2.6cm}M{4.6cm}M{6.4cm}@{}}
\toprule
\apphead Axis & Description & Function \\
\midrule
\appkey{Scale Up}   & Increase the capability of the shared prior. & Makes small updates high leverage by providing richer latent structure to adapt. \\
\addlinespace[2pt]
\appkey{Scale Down} & Reduce the marginal cost and instability of adaptation. & Makes repeated learning, storage, and serving feasible for many instances. \\
\addlinespace[2pt]
\appkey{Scale Out}  & Increase the number and diversity of persistent adapters. & Enables personalization, population diversity, and emergent population-level behavior. \\
\bottomrule
\end{tabular}%
}
\end{table}

\section{Scale Up: Scaling Model Capacity for a Stronger Shared Prior}\label{sec:scale-up}

\pony{Scale Up is the first technical requirement for persistent personal model instances. A personal adapter is useful only when the shared base model already contains broad capabilities that a small local update can redirect. The goal is therefore not to replace personalization with larger pretraining, but to make stronger shared base models repeatedly adaptable under realistic learning budgets.}

\pony{This question is especially sharp in reinforcement learning. RL can reinforce reasoning strategies, tool-use policies, and long-horizon behaviors, but it can only reinforce trajectories that the current policy can sample with sufficient probability. A stronger shared base model changes the effective search space by making useful but unstable behaviors reachable. LoRA changes the economics by allowing the learning loop to operate on that stronger base without updating all parameters. From the perspective of personal models, Scale Up asks why a small local adaptive state becomes more valuable as the shared base becomes stronger.}

\pony{This section develops the Scale Up axis in five steps. First, we argue that RL is prior-limited: the base model determines which trajectories exploration can discover and which behaviors credit assignment can reinforce. Second, we explain how LoRA changes the scaling trade-off by turning PEFT into budgeted access to stronger priors. Third, we discuss how trillion-scale LoRA RL makes such priors operationally reachable in real learning loops. This is an engineering feasibility argument, distinct from the capability argument above, and both are necessary for the Scale Up thesis. Fourth, we analyze scale-induced failure modes that emerge when large-prior adaptation spans sparse architectures, distributed rollout, sharded optimization, adapter semantics, and serving runtimes. Finally, we connect Scale Up to the next axis: once strong priors can be adapted, the next question is how small, stable, and efficiently trainable the local adaptive state can become.}

\subsection{Why RL is Prior-Limited}\label{subsec:prior-limited}

Recent progress in reasoning-oriented reinforcement learning has made one point increasingly difficult to ignore: under realistic budgets, RL is often more effective at amplifying behaviors that already exist in weak or unstable form than at creating sophisticated capabilities \emph{de novo} \citep{deepseek2025r1,hu2025openreasonerzero}. DeepSeek-R1-Zero showed that large-scale RL can elicit reasoning behaviors such as self-reflection, verification, and long-chain exploration without relying solely on conventional supervised reasoning traces \citep{deepseek2025r1}. Open-Reasoner-Zero further showed that relatively simple on-policy RL recipes can improve reasoning performance when applied to a sufficiently capable base model \citep{hu2025openreasonerzero}. These results support the view that RL can unlock latent capabilities, but they do not imply that RL is independent of the base model. On the contrary, they make the role of the base-model prior more visible.

RL does not search over an abstract space of possible capabilities. It searches through the trajectory distribution induced by the current policy. In language-model RL, the action space is the vocabulary-conditioned continuation distribution over long sequences, and the probability of sampling a useful reasoning trajectory is strongly shaped by the pretrained policy. The bottleneck is therefore not only reward quality, but trajectory support: the base policy must assign sufficient probability mass to behaviors that the reward can later select and reinforce.

This prior-limited view explains why stronger models can make RL more productive. Exploration improves when the model can already propose partially useful trajectories. Credit assignment improves when competent trajectories occur often enough to be distinguished from noise. Transfer improves when the base model already encodes broad latent structure. A weak model may rarely visit high-reward reasoning patterns, making policy-gradient updates sparse, high-variance, or overly dependent on reward shaping. By contrast, a strong model can assign non-negligible probability mass to useful but unstable behaviors, allowing RL to reinforce and regularize them.

This observation reframes the role of scale. Larger models are not valuable only because they have higher static benchmark performance. They are valuable because they alter the distribution of trajectories available to post-training. A stronger prior can make reasoning paths, tool-use strategies, and long-horizon behaviors reachable before they are stable. RL can then act less as a mechanism for inventing capability from scratch and more as a mechanism for selecting, sharpening, and stabilizing behaviors already latent in the model.

For personal models, this point is fundamental. A local adaptive state can only be high-leverage if the shared prior already contains useful structure to redirect. Without such a prior, personalization risks collapsing into shallow memorization, narrow task fitting, or brittle behavioral patches. With a strong prior, small updates can produce disproportionately large behavioral changes because they operate on a model that already encodes much of the world, the task format, and the relevant reasoning patterns.

\subsection{LoRA as Budgeted Access to Strong Priors}\label{subsec:lora-budgeted-prior}

LoRA changes the scaling trade-off from how many parameters can be updated to how much prior can be brought into the learning loop under a fixed budget. LoRA was originally proposed as a low-rank adaptation method that freezes pretrained weights and injects trainable low-rank matrices, thereby reducing the number of trainable parameters and the deployment cost by orders of magnitude while preserving the shared base model \citep{hu2021lora}. In ordinary supervised fine-tuning, LoRA is often interpreted as a storage- and memory-saving technique. In the Scale Up axis, however, LoRA has a more structural role: it makes stronger priors economically accessible to repeated optimization.

This distinction matters because the central comparison is no longer simply `full fine-tuning versus adapter tuning''. It is instead `how much prior can be brought into the learning loop under a fixed adaptation budget''. A smaller fully trainable model may expose more trainable parameters relative to its size, but it may still lack the latent reasoning substrate that makes RL productive. A larger LoRA-adapted model may expose fewer trainable parameters, but those parameters act on a stronger prior. In this regime, the adapter does not need to carry the full capability. It only needs to steer the prior.

LoRA and full-parameter training should therefore not be treated as identical optimization regimes. Prior work has shown that low-rank adaptation differs from full fine-tuning in forgetting behavior, representation movement, and update geometry \citep{biderman2024loraforgets,shuttleworth2025loraillusion}. Notably, \citet{biderman2024loraforgets} find that LoRA forgets less than full fine-tuning, a property that is advantageous for personal models that must preserve base capabilities while acquiring new ones. \citet{shuttleworth2025loraillusion} further show that LoRA and full fine-tuning are not equivalent in representation movement, cautioning against the simplistic claim that LoRA is always a substitute for full training. Together, these differences explain why LoRA can be especially effective when the goal is not to relearn a task from scratch, but to modulate a strong pretrained representation. The rank constraint is a limitation when new capability must be built from little support, but it can be an advantage when adaptation only needs to redirect latent structure.

The motivating comparison in this manuscript illustrates this point. Under broadly comparable RL budgets, larger base models adapted with LoRA achieved larger headroom-normalized gains than a much smaller model trained with full-parameter RL, despite using substantially fewer trainable parameters. This comparison separates three quantities that are often conflated: total model capacity, activated computation, and trainable parameter count. Its significance is not a universal ranking of LoRA over full training. Rather, it shows that, when learning budgets are fixed, the strength of the prior can matter more than the size of the trainable surface.

\begin{table}[tbp]
\centering
\small
\setlength{\tabcolsep}{6pt}
\renewcommand{\arraystretch}{1.20}
\caption{Large prior plus small LoRA update can outperform full RL on a smaller model under comparable RL budgets. Values are summarized from the motivating Mind Lab trillion-parameter LoRA RL study.}
\label{tab:prior}
\fittowidth{%
\begin{tabular}{@{}M{5.6cm}M{2.4cm}M{2.6cm}M{2.6cm}@{}}
\toprule
\apphead Model and adaptation & Trainable parameters & AIME 2025 normalized gain & GPQA Diamond normalized gain \\
\midrule
\appkey{DS-Distill-Qwen-1.5B}, full RL          & 1.5B  & 8.33\%  & 25.00\% \\
\addlinespace[2pt]
\appkey{DS-Distill-Qwen-7B}, LoRA $r{=}64$      & 0.16B & 11.31\% & 27.23\% \\
\addlinespace[2pt]
\appkey{DS-Distill-Qwen-32B}, LoRA $r{=}8$      & 0.07B & 20.61\% & 33.02\% \\
\bottomrule
\end{tabular}%
}
\end{table}

This is the core principle behind Scale Up. The objective is to make stronger priors repeatedly accessible to PEFT-based learning, rather than simply to make models larger. If a frontier prior can only be used as a static checkpoint, its value for personalization and continual adaptation remains limited. If it can be updated cheaply and reliably, it becomes a reusable substrate for reasoning optimization, tool-use refinement, domain specialization, and, ultimately, memory formation in persistent personal models.

A note on interpretation: the motivating comparison above varies model size and training method simultaneously, so it does not cleanly isolate prior strength from trainable parameter count. Its significance is not a universal ranking of LoRA over full training. Rather, it illustrates that, when learning budgets are fixed, the strength of the prior can matter more than the size of the trainable surface. The subsequent engineering evidence in Section~\ref{subsec:trillion-practical} supports this claim by showing that large-prior LoRA RL can be made operationally stable.

\subsection{Operationalizing Trillion-Scale LoRA RL}\label{subsec:trillion-practical}

If strong priors make adaptation high-leverage, the next question is whether such adaptation can be made operationally viable at frontier scale. At this scale, LoRA is no longer merely a low-rank parameterization. It becomes part of a distributed learning system. The adapted policy must remain coherent across rollout, optimization, checkpointing, model conversion, and serving. Frontier-scale PEFT therefore depends as much on systems alignment as on parameter efficiency.

\begin{figure}[tbp]
\centering
\includegraphics[width=0.9\textwidth]{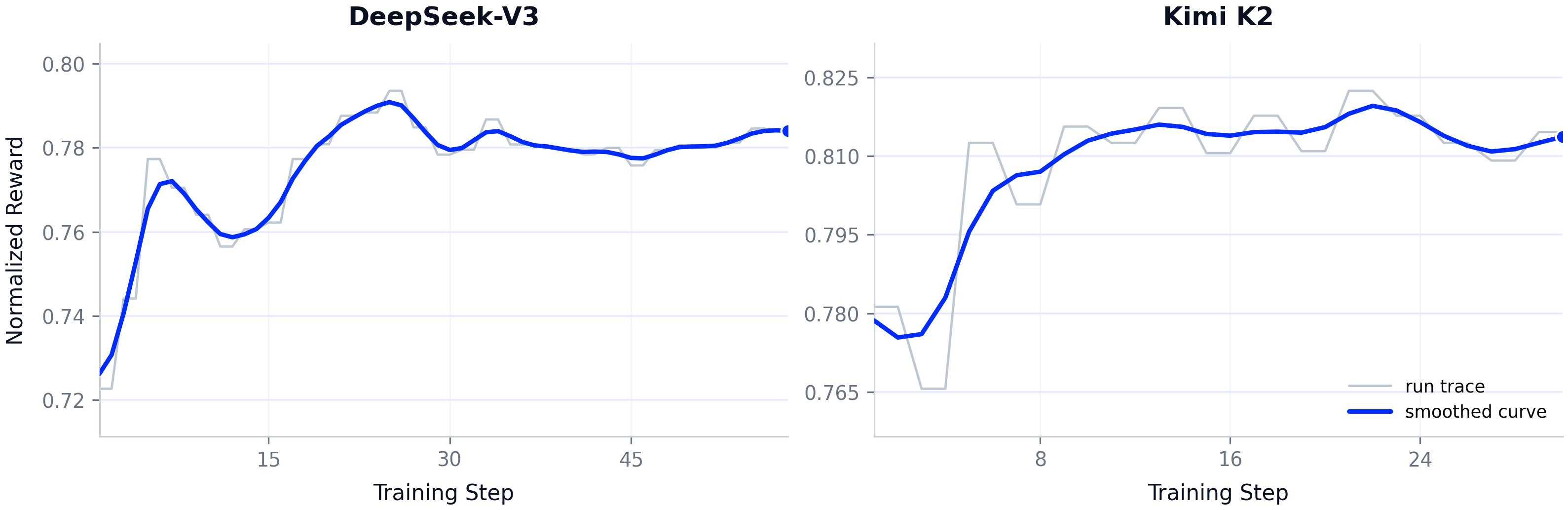}
\caption{\pony{GRPO LoRA training on large-scale LLMs. Stable reward and task-success curves support the feasibility of adapting a trillion-parameter MoE reasoning model with LoRA-based RL when rollout, training, and hybrid parallelism are jointly designed.}}
\label{fig:scale-up-trillion-lora-rl}
\end{figure}

Trillion-scale LoRA RL combines several difficulties that are usually treated separately. The base model is too large for naive replication. Sparse MoE structure introduces expert routing and all-to-all communication. The RL loop requires frequent alternation between inference-style rollout and training-style optimization. The adapter state is small relative to the base model, but it must still be sharded, synchronized, merged, and served correctly. A naive data-parallel LoRA implementation fails because it assumes that the base model can be treated as a monolithic frozen object and that the adapter can be attached as a local modification. At trillion scale, neither assumption holds.

This challenge is especially acute in mixture-of-experts architectures. In dense models, numerical differences between training and inference may perturb activations while leaving the computational pathway unchanged. In MoE models, the same perturbations can alter routing decisions, causing tokens to traverse different experts and thereby changing the effective computation itself. Moreover, expert layers and dense layers play different roles in the computation. If adapters are attached only to dense submodules, the RL signal may not adequately affect expert-specific behavior. If adapters are attached naively to expert submodules, communication and checkpoint handling can become infeasible. Practical trillion-scale LoRA RL therefore requires adapter placement to be co-designed with the model's sparse computation structure.

The Kimi K2 LoRA RL case serves as a proof of existence for this regime: a trillion-parameter sparse prior can be brought into an on-policy RL loop without full-parameter updates, provided that adapter placement, rollout generation, and distributed optimization are co-designed. The system applies RL to a trillion-parameter MoE reasoning model with 32.6B activated parameters and 1.04T total parameters. It uses LoRA on selected dense and expert layers, a GRPO-style on-policy optimization loop, and hybrid tensor, pipeline, expert, and sequence parallelism. This configuration is important not because each component is individually novel, but because all components must operate as a single adaptation system.

The use of GRPO-style on-policy optimization is central to the Scale Up argument because it makes the system a reinforcement learning loop rather than a static supervised adapter update. The adapter is optimized from sampled reasoning trajectories and reward feedback, so the quality of the base prior directly affects both exploration and credit assignment. In other words, the system tests the central claim under an actual RL regime: a strong prior is useful only if it can be sampled, evaluated, updated, and re-sampled without breaking policy consistency.

The core design principle is to treat parallelism as a schedulable resource rather than a fixed layout. Rollout and training have different computational profiles. Serving-oriented rollout benefits from high-throughput decoding and efficient KV-cache management, whereas training-oriented optimization requires sharded gradients, optimizer states, and backward computation. A practical system must bridge these two regimes without allowing the policy used for sampling to drift from the policy used for optimization. This is why integrating a rollout engine with a Megatron-style training backend is conceptually central. Such integration converts a frontier prior from an expensive static object into an adaptation target that can repeatedly enter the RL loop.

LoRA contributes to this system in three ways. First, it reduces the amount of trainable state, making optimizer memory and gradient communication manageable. Second, it enables multiple RL runs or downstream variants to share the same frozen base model, which is essential for amortizing the cost of frontier-scale priors. Third, when attached to both dense and expert components, it allows the RL signal to influence global reasoning behavior as well as expert-specific computation. In the Kimi K2 system, this design reduces the compute and communication footprint to approximately 10\% of conventional full-parameter RL while preserving the ability to adapt a trillion-scale prior.

Empirically, the Kimi K2 experiment provides three pieces of evidence for this claim. First, LoRA RL reduces the required GPU budget to roughly 10\% of conventional full-parameter RL on the same class of model. Second, the training curves show smooth improvement in reward and task success rate without catastrophic divergence. Third, held-out evaluations suggest that the adapter improves task-specific behavior while preserving the general capabilities of the base model. These observations show that trillion-scale LoRA RL is not merely memory-efficient. It can also remain stable and behaviorally useful when the distributed system is designed around MoE parallelism.

This result establishes the positive side of Scale Up: strong priors can be made reachable by PEFT-based RL. It also reveals the other side of Scale Up. Once a frontier prior becomes trainable, the learning problem no longer resides only in the optimizer or the adapter. It is distributed across the execution path that generates rollouts, the backend that computes gradients, the sparse architecture that routes tokens, and the runtime that later serves the adapted model. Scale therefore introduces not only larger capability, but also new failure modes.

\subsection{Scale-Induced Failure Modes}\label{subsec:scale-failures}

\begin{figure}[tbp]
\centering
\includegraphics[width=0.7\textwidth]{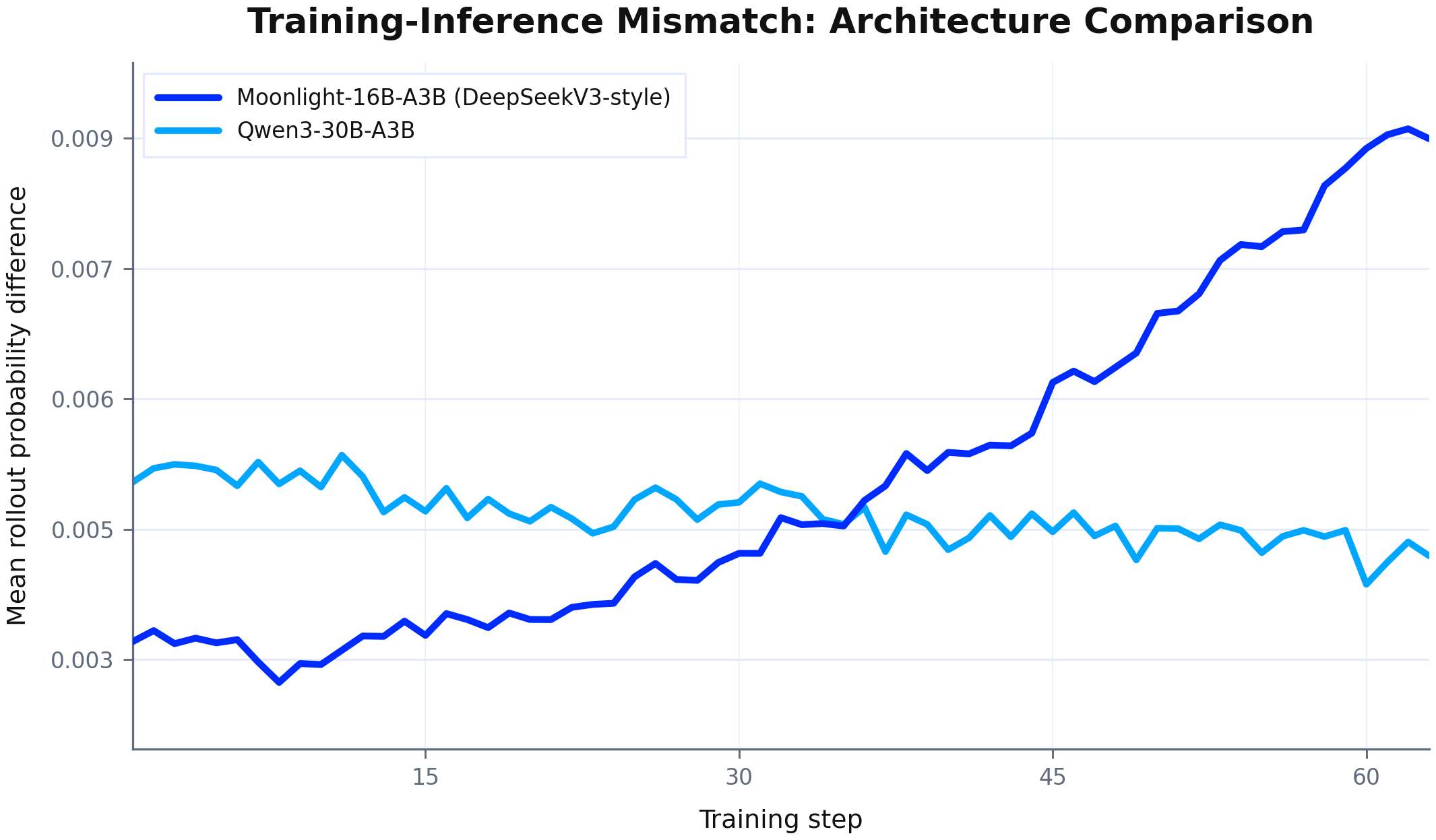}
\caption{TIM comparison between Moonlight-16B-A3B (DeepSeekV3-style) and Qwen3-30B-A3B. The DeepSeekV3-style architecture shows substantially higher training--inference mismatch.}
\label{fig:r3-mismatch-routing}
\end{figure}

Scaling up the prior does not merely increase the number of parameters in an otherwise unchanged training recipe. It changes the failure surface of PEFT-based reinforcement learning. In small and medium-scale models, discrepancies among rollout, training, checkpointing, and serving are often absent, numerically negligible, or hidden by a simple execution path. At frontier scale, especially in sparse MoE reasoning models, the same discrepancies can become algorithmic, architectural, adapter-semantic, and lifecycle-level failures.

We organize these scale-induced failures into four categories. First, \textbf{algorithmic mismatch failures} arise when the policy that generates rollouts differs from the policy optimized during training, violating the assumptions of on-policy RL. Second, \textbf{sparse-architecture failures} arise when small execution differences change expert routing, sparse-attention selection, or other discrete computation paths. Third, \textbf{adapter-semantics failures} arise when a low-rank adapter is attached, transformed, or interpreted in a way that changes the intended meaning of the adapted update. Fourth, \textbf{lifecycle and serving failures} arise when the trained adapter is saved, merged, quantized, or served under a runtime that no longer instantiates the same effective computation. This taxonomy is not meant to separate independent bugs. Rather, it identifies the coupled failure surface that emerges when large-prior adaptation spans reinforcement learning, sparse architectures, distributed execution, adapter semantics, and serving infrastructure.

\begin{figure}[tbp]
\centering
\begin{subfigure}[t]{0.32\textwidth}
\centering
\includegraphics[width=\linewidth]{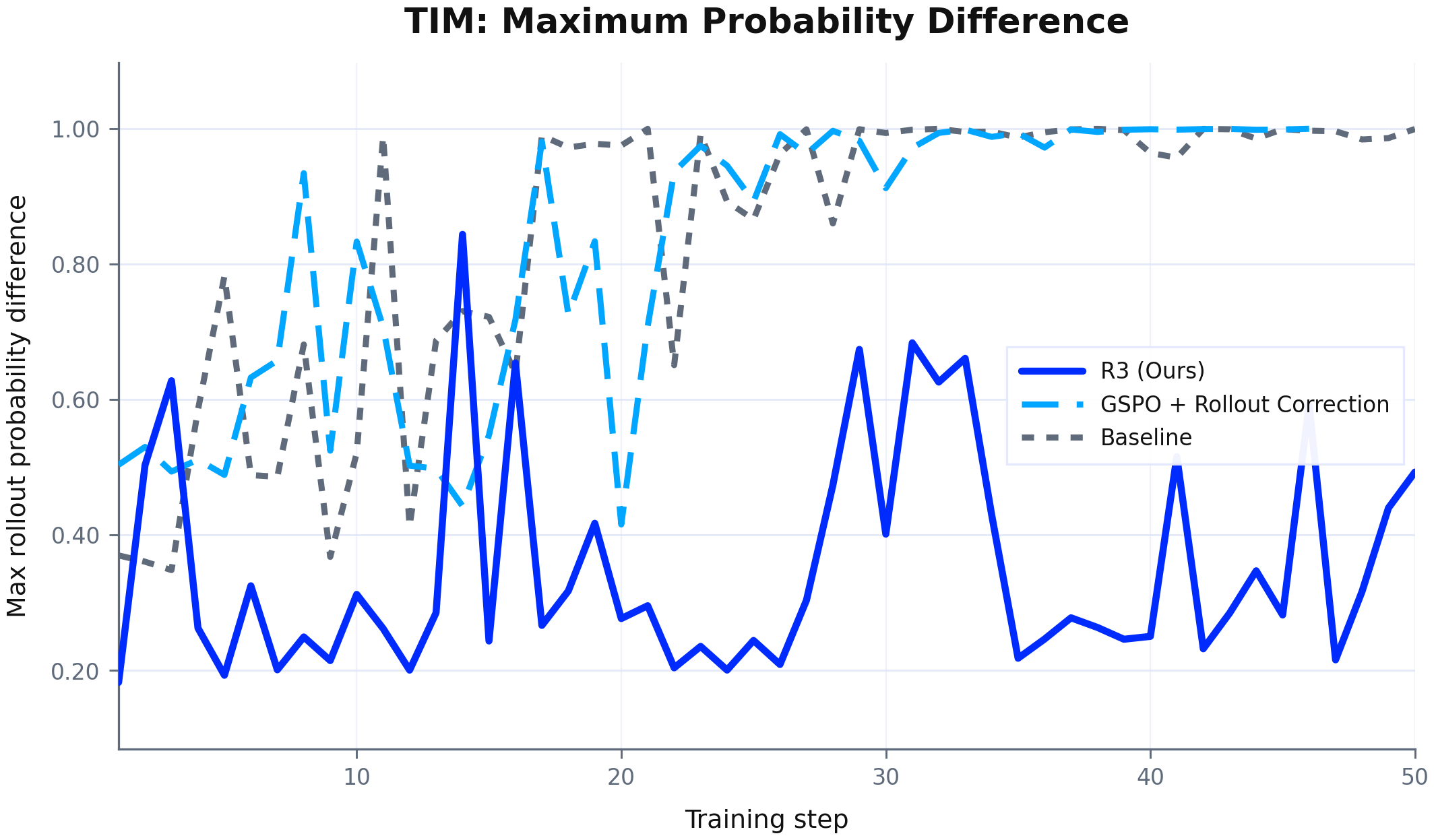}
\caption{Maximum rollout probability difference per step. The original and rollout-corrected runs climb rapidly and remain high, whereas the R3-fixed run stays consistently lower.}
\label{fig:r3-tim-max-prob-diff}
\end{subfigure}
\hfill
\begin{subfigure}[t]{0.32\textwidth}
\centering
\includegraphics[width=\linewidth]{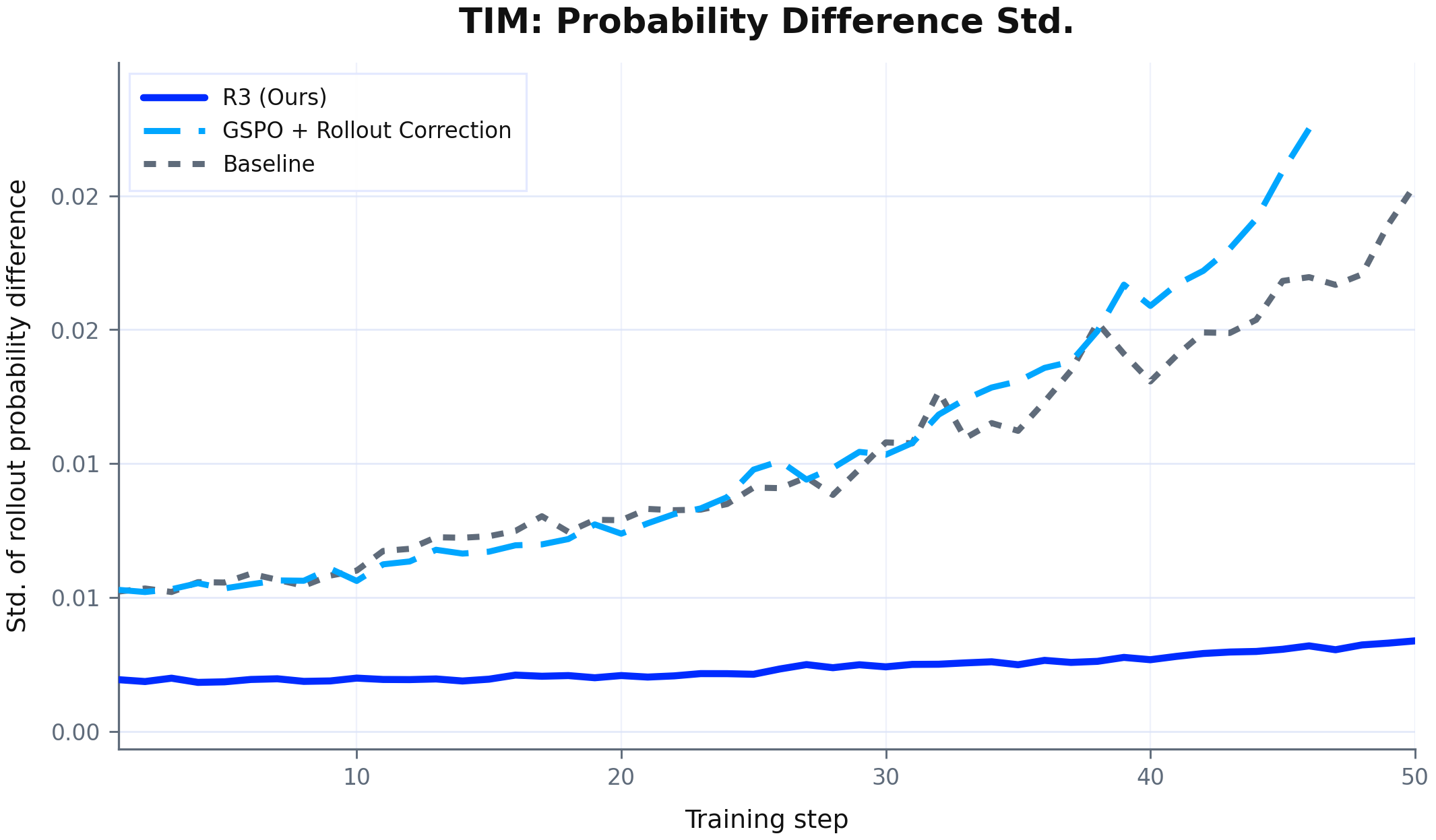}
\caption{Standard deviation of rollout probability difference. R3 maintains lower variance throughout training.}
\label{fig:r3-tim-std}
\end{subfigure}
\hfill
\begin{subfigure}[t]{0.32\textwidth}
\centering
\includegraphics[width=\linewidth]{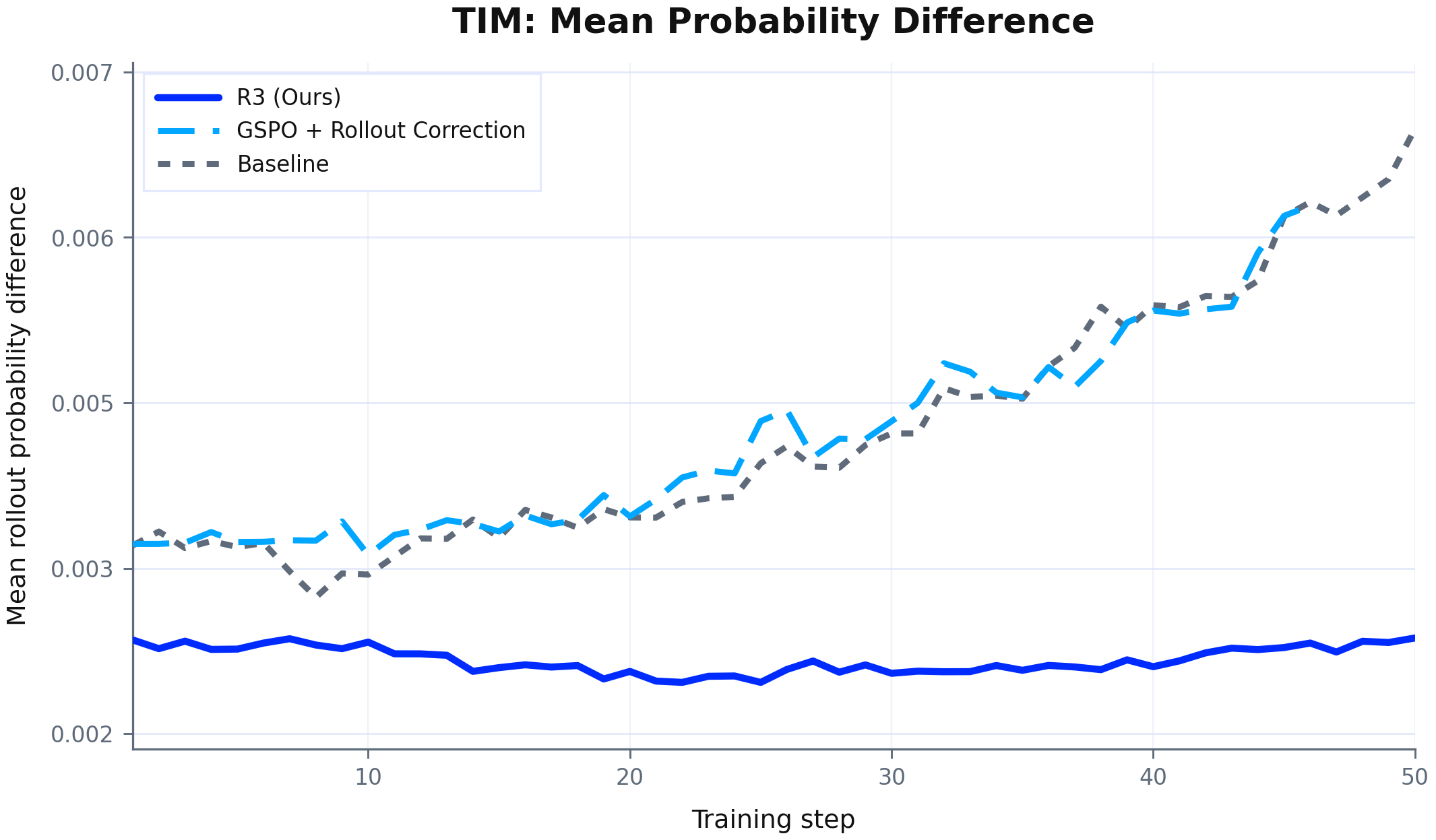}
\caption{Mean rollout probability difference. R3 substantially reduces TIM relative to the baselines.}
\label{fig:r3-tim-mean-prob-diff}
\end{subfigure}
\caption{TIM metrics: rollout probability mismatch. Across maximum difference, standard deviation, and mean difference, Router Replay R3 consistently reduces the mismatch between rollout-side and training-side probabilities relative to the original and rollout-corrected baselines.}
\label{fig:r3-tim-rollout-probability-mismatch}
\end{figure}

The first case is training--inference mismatch (TIM), an algorithmic mismatch failure. In a standard policy-gradient view, trajectories are sampled from a behavior policy, and gradients are computed to improve the corresponding policy under some form of probability-ratio correction. This view assumes that the policy used for rollout and the policy used for optimization are well defined and comparable. In large-model RL systems, however, rollout and training are often executed by different engines: a serving-oriented inference stack generates trajectories, while a training-oriented distributed backend computes losses and updates. In dense models, the difference between these two executions may remain a small numerical perturbation. In MoE models, the same perturbation can change discrete routing decisions, causing tokens to pass through different experts. Once this occurs, the training engine is no longer optimizing the same effective computation that produced the rollout.

\begin{figure}[tbp]
\centering
\begin{subfigure}[t]{0.48\textwidth}
\centering
\includegraphics[width=\linewidth]{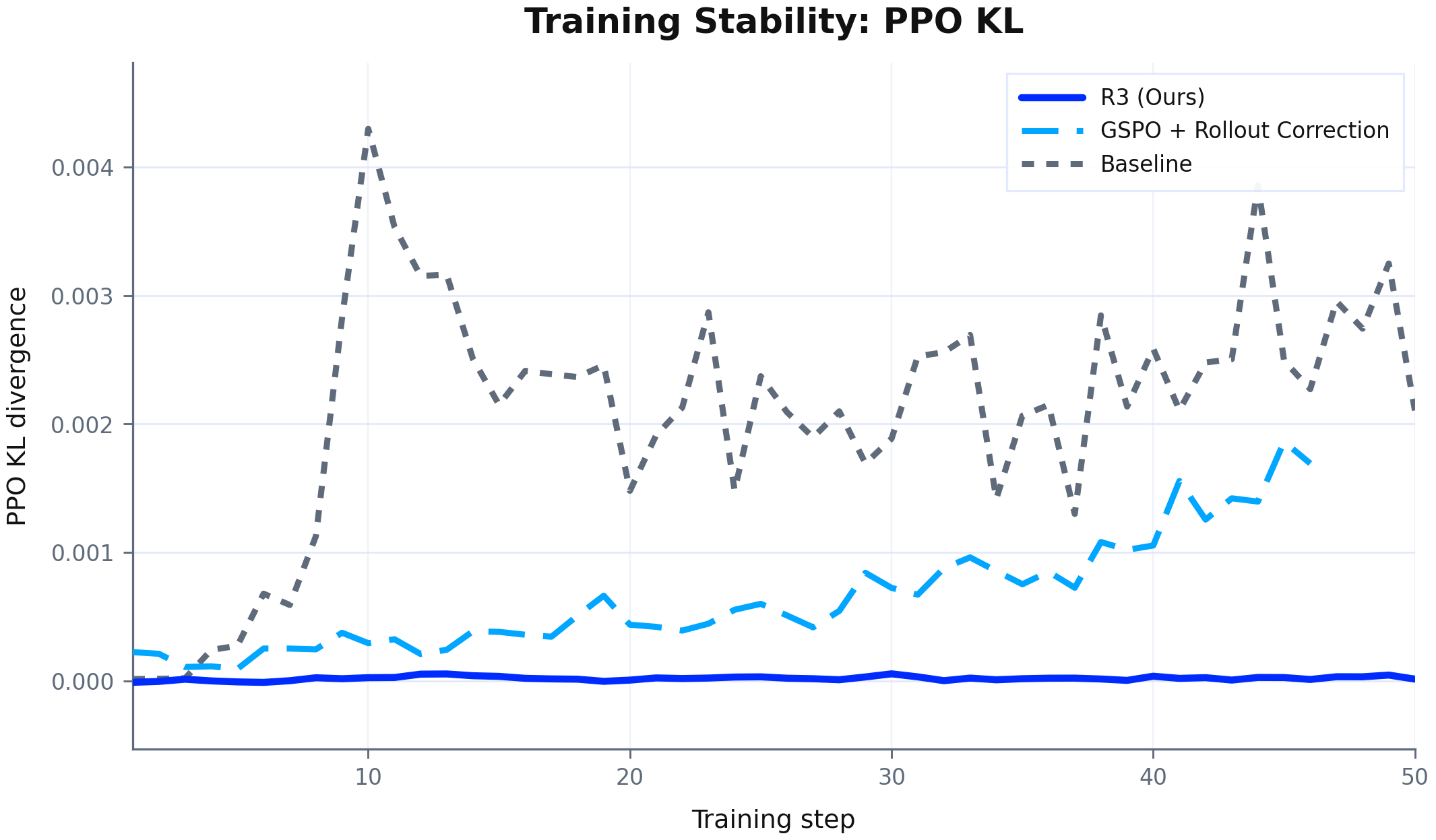}
\caption{PPO KL divergence. The R3 run maintains near-zero KL divergence (0.000026 at step 46), while other runs increase substantially.}
\label{fig:r3-train-stable-ppo-kl}
\end{subfigure}
\hfill
\begin{subfigure}[t]{0.48\textwidth}
\centering
\includegraphics[width=\linewidth]{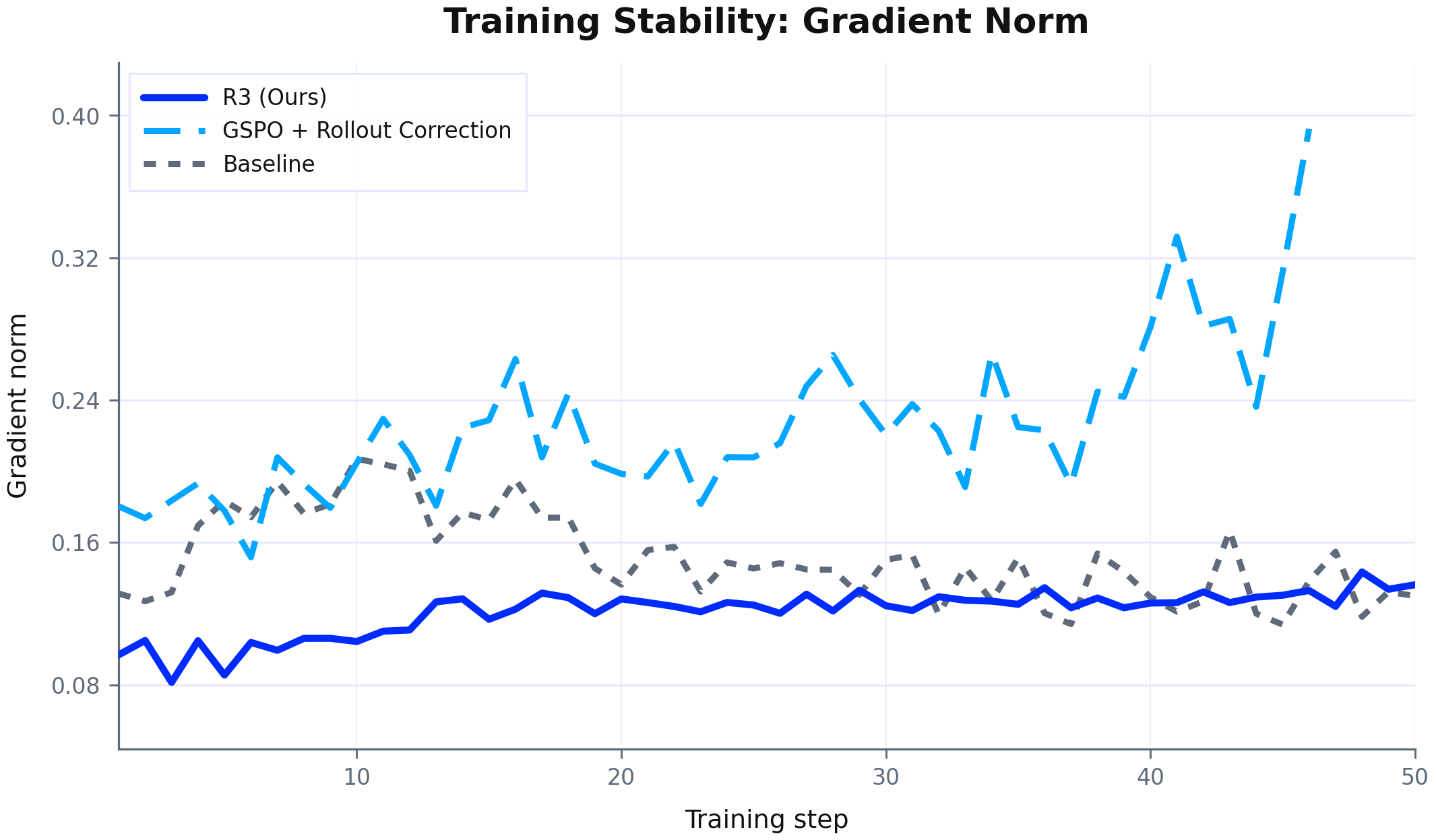}
\caption{Gradient norm. R3 exhibits the most stable gradients, whereas the baselines show higher variance and drift.}
\label{fig:r3-train-stable-grad-norm}
\end{subfigure}
\caption{Training stability signals. Router Replay R3 stabilizes the RL update by maintaining lower KL divergence and smoother gradient norms than the original and rollout-corrected baselines.}
\label{fig:r3-training-stability-signals}
\end{figure}

TIM is therefore more than a numerical precision issue. It becomes an algorithmic failure mode because it changes the object of the policy update. The nominal policy may be the same checkpoint with the same adapter, but the effective policy differs if the rollout path and the training path activate different experts. Importance correction can mitigate moderate probability mismatch, but it assumes that the two policies remain comparable distributions over the same underlying computation. When routing divergence changes the computation graph itself, this assumption becomes fragile. This is why large-scale MoE RL requires mechanisms that are not central in small-model training, such as routing-aware correction, routing replay, or stricter rollout--training equivalence.

Router Replay R3 illustrates how a sparse-architecture failure can become an RL failure. Its significance is not merely that it fixes a particular implementation problem, but that it treats routing as part of the effective computation to be preserved. In routing-sensitive MoE architectures, preserving the sampled computation requires preserving the routing decisions that determine which experts participate in the forward pass. If inference-side and training-side routing diverge, gradients are computed through a sparse path different from the one that generated the samples. R3 addresses this issue by recording routing information during rollout and replaying it during training, thereby reducing the semantic gap between the sampled policy and the optimized policy. The broader lesson is that, at scale, the RL algorithm can no longer be defined independently of the execution path that realizes the policy.

\begin{figure}[tbp]
\centering
\begin{subfigure}[t]{0.48\textwidth}
\centering
\includegraphics[width=\linewidth]{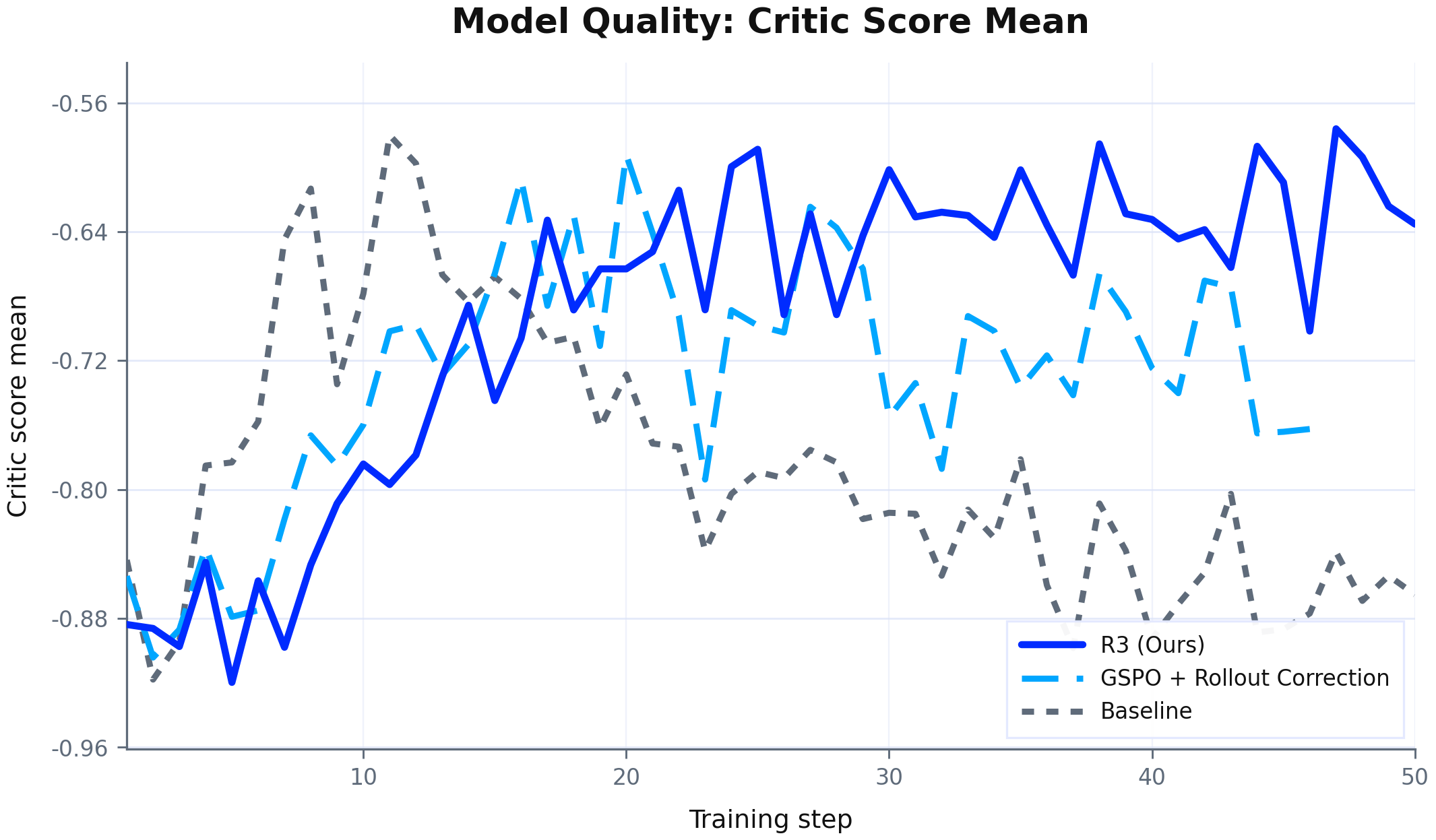}
\caption{Critic score mean. The R3 run sustains higher scores than the baselines, which trend downward.}
\label{fig:r3-eval-critic-mean}
\end{subfigure}
\hfill
\begin{subfigure}[t]{0.48\textwidth}
\centering
\includegraphics[width=\linewidth]{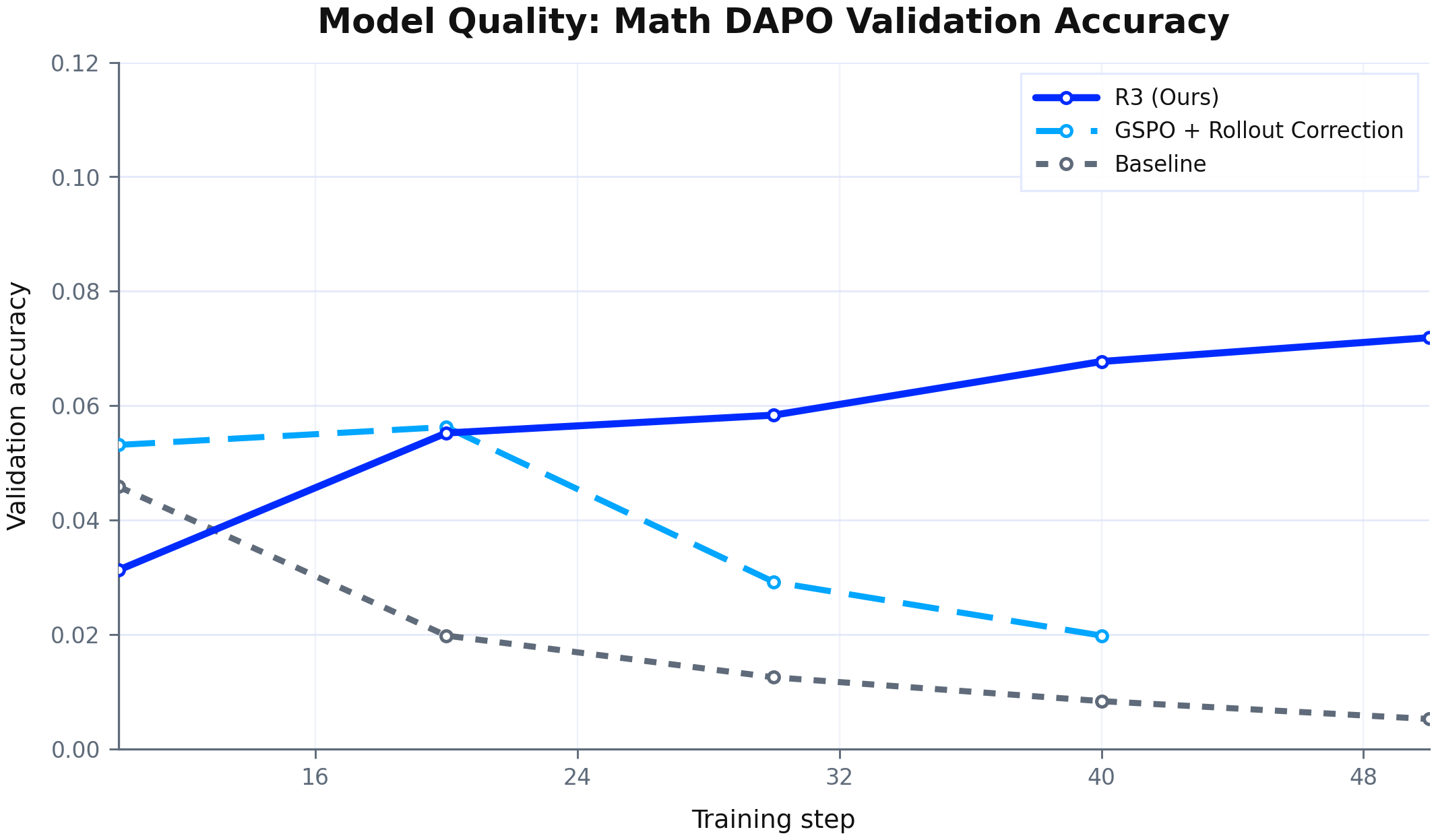}
\caption{Validation accuracy on the math DAPO task. The R3 run improves monotonically and finishes strongest, while the original and rollout-corrected runs degrade.}
\label{fig:r3-eval-val-acc}
\end{subfigure}
\caption{Quality metrics. Router Replay R3 improves downstream quality by sustaining higher critic scores and achieving stronger validation accuracy on the math DAPO task.}
\label{fig:r3-quality-metrics}
\end{figure}

The second case is GLM5 and GLM5.1 support, which illustrates how sparse-architecture and adapter-semantics failures can appear together. Supporting a small dense model often means implementing the architecture, loading the checkpoint, attaching adapters, and running training or inference with relatively direct correspondence among these steps. Supporting a frontier MoE model is fundamentally different. GLM5-style models combine MoE, Multi-Head Latent Attention (MLA), DeepSeek Sparse Attention (DSA), Multi-Token Prediction (MTP), LoRA adaptation, training-time distributed execution, inference-time serving kernels, and checkpoint bridge logic. Each component can be locally correct, yet the full system can still be globally inconsistent if the training stack, inference stack, and bridge do not interpret the model in the same way.

This problem appears in several forms. DSA makes token selection part of the computation semantics: a small difference in the indexer or top-$k$ behavior can change which tokens are included in sparse attention. MTP touches model structure, loss computation, output heads, and checkpoint conversion simultaneously, so an adapter trained under one interpretation of the MTP path may not be equivalent when loaded under another runtime. MLA and other specialized projection modules may contain custom forward logic, dtype behavior, fused kernels, or tensor-parallel communication patterns. A generic LoRA wrapper that is correct for ordinary linear layers may therefore be semantically incorrect for these modules. In such cases, the adapter may load successfully, but the served model is no longer the same adapted computation that was trained.

The GLM5/GLM5.1 case therefore shows that, at frontier scale, infrastructure is not a passive substrate for algorithms. It becomes part of the model definition. The bridge from training to inference is a semantic preservation problem rather than a file-format conversion. The serving runtime must preserve the same sparse attention, latent attention, expert, and adapter semantics. The training backend must instantiate the same computation that the inference engine will later serve. This is why model-family support at scale often becomes a full-stack alignment problem spanning training, inference, and checkpoint conversion.

\begin{figure}[tbp]
\centering
\includegraphics[width=0.9\textwidth]{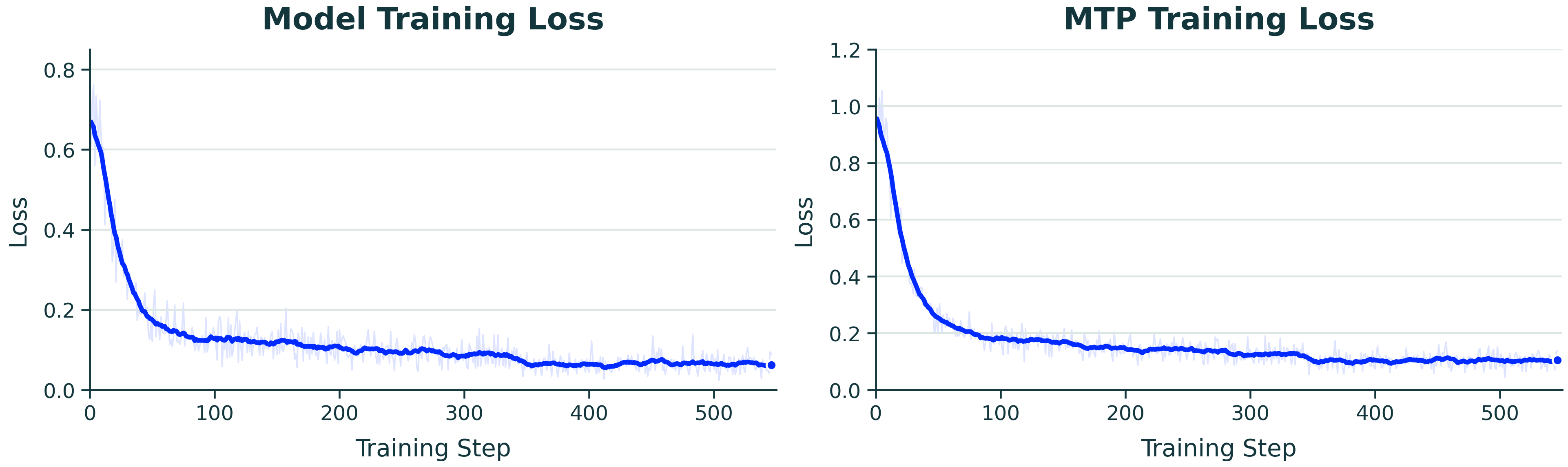}
\caption{Training loss curves for the model and MTP components of GLM5.1 with LoRA adapters.}
\label{fig:glm5-mint-stack}
\end{figure}

Together, TIM, R3, and GLM5 support show that scale changes not only what the model can learn, but also how it can fail. At frontier scale, PEFT-based RL fails not only through poor rewards, unstable gradients, or insufficient adapter capacity, but also through mismatches among routing, sparse attention, adapter interpretation, checkpoint conversion, and serving execution. These failures are distinctive to large-prior adaptation because the effective computation is distributed across multiple systems rather than contained in a single local model object.

This reframes failure modes at scale. They are not simply bugs, numerical instabilities, or inconvenient engineering details. They indicate that the training regime itself has changed. Small-model experience often assumes that model execution is simple enough for training, inference, and serving to be treated as interchangeable realizations of the same policy. Frontier-scale MoE PEFT breaks this assumption. Routing, sparse selection, adapter interpretation, checkpoint conversion, and serving kernels all become part of the operational meaning of the adapted model. As a result, correctness must be defined not only by whether training loss decreases, but also by whether the adapted computation remains consistent across the full lifecycle from rollout and training to export, merge, and serving.

For persistent personal models, this issue becomes even more important. Adapters are not disposable optimization artifacts. They may carry task-specific behavior, user-specific preferences, domain experience, or long-term memory. If the infrastructure reinterprets an adapter during loading or serving, the system may preserve the file while losing the learned behavior. If the training algorithm optimizes an effective computation different from the rollout policy, the system may improve a surrogate that was never actually deployed. Scale Up therefore requires more than access to a large prior. It requires mechanisms that preserve the meaning of adaptation across the full lifecycle of the adapted model.

\subsection{From Scale Up to Scale Down}\label{subsec:scale-up-to-down}

Scale Up establishes the first condition for PEFT-based personal models: the adaptive boundary must sit on top of a sufficiently strong shared prior. RL is prior-limited because it can only reinforce trajectories that the current policy can sample. LoRA changes the economics of this process by allowing stronger priors to enter the learning loop under a fixed adaptation budget, and trillion-scale LoRA RL shows that such priors can be adapted in real on-policy training systems. Scale-induced failure modes then show that this adaptation must remain consistent across sparse architectures, distributed execution, adapter semantics, and serving lifecycles.

However, Scale Up alone is not enough. A strong prior can make each update more powerful, but it does not by itself make adaptation continuous, stable, or cheap enough to support persistent model instances. If the adaptive unit is too large, unstable, brittle, or expensive to train and serve, then large-prior adaptation remains occasional rather than continuous. Powerful priors would remain impressive shared artifacts, but not writable substrates for long-term personalization.

This is where the next axis begins. Scale Down asks how small, stable, and efficiently trainable the adaptive unit can become. It shifts attention from the strength of the shared prior to the cost, reliability, and repeatability of the local update. In the three-axis framework, Scale Up supplies the capability substrate. Scale Down then makes the adaptive unit cheap and reliable enough for repeated learning, and Scale Out expands these adaptive units into populations of persistent model instances.

These observations foreshadow the infrastructure problem developed in Chapter~6, where we introduce MinT~\citep{mindlab2026mint} as a LoRA-based framework for multi-train and multi-serve learning over shared base models. For the purposes of the present Scale Up axis, the point is narrower: once a strong prior can be adapted, the system must make that adaptation repeatable across training, evaluation, transfer to inference, and serving. This requirement exposes the next bottleneck. The adaptive unit itself must be compact, stable, and efficiently writable; otherwise, large-prior adaptation remains an isolated event rather than a sustained learning regime.

\section{Scale Down: The Operating Regime of Efficient Adaptation}\label{sec:scale-down}

If Scale Up gives PEFT its leverage, Scale Down decides whether that leverage can be exercised widely enough to sustain persistent personal model instances. The goal is to identify the smallest reliable unit of persistent learning: a local adaptive state that is expressive yet stable, cheap to update and store, and practical to serve across many instances at once.

How small can this adaptive state become before it stops learning reliably? The answer is a regime, not a fixed threshold. Under standard recipes, middle-rank adapters already learn dependably across seeds. Extremely low-rank adapters are different: they reach the same best-case performance, but they do not reach it reliably, and closing that reliability gap still depends on better initialization, tighter variance control, and transferable hyperparameters. This leads to a deliberately limited claim. The smaller an adapter can become while staying stable, the cheaper it is to train, store, and serve the many instances a personal-model population requires.

This section develops Scale Down in two movements. \Cref{subsec:inside-lora} stays within the standard LoRA setting, in which the adaptive state is a static low-rank update, and examines three coupled questions: how far rank can be reduced before across-seed reliability collapses, how RL-native initialization can rescue the tiny-adapter regime, and how learning-rate recipes can transfer across ranks so that adapter populations do not require per-instance tuning. \Cref{subsec:beyond-lora} then looks beyond static LoRA and asks whether the adaptive unit can itself be redesigned as a compact, writable state that accumulates interaction history, taking $\delta$-mem and related stateful adapters as the representative case. Together, the two movements extend Scale Down from parameter reduction to adaptive-state design.

\subsection{Inside LoRA: Finding the Efficient Operating Regime}\label{subsec:inside-lora}

LoRA provides the clearest laboratory for studying efficient adaptation because it exposes the central tradeoff directly: how much learning can survive as the trainable surface becomes extremely small. The design space is not only about low rank in the abstract, but about the joint control of expressive capacity, optimization stability, and operational simplicity. The three subsections below form a progression. Rank reduction asks how small the adaptive state can become before the across-seed mean collapses, even when the best run remains competitive. Initialization asks how such tiny adapters can be made reliable rather than merely occasionally lucky under RL optimization. Hyperparameter reuse asks whether the resulting recipe can be reused across the large populations of adapters that personal-model deployment requires.

\subsubsection{Rank Reduction Under Minimal Parameters}\label{subsubsec:rank-reduction}

A central question in PEFT is whether useful learning survives under extreme rank reduction. From this perspective, LoRA rank is not simply a monotonic capacity knob. It defines an adaptation regime: how small can the adaptive state become before the across-seed mean degrades, even though the best run remains competitive with much larger adapters?

\begin{figure}[tbp]
\centering
\begin{subfigure}[t]{0.33\textwidth}
\centering
\includegraphics[width=\linewidth]{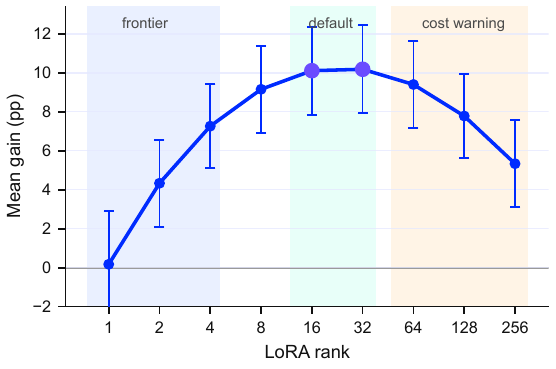}
\caption{Mean gain by rank.}
\label{fig:rank-overview-gain}
\end{subfigure}\hfill
\begin{subfigure}[t]{0.33\textwidth}
\centering
\includegraphics[width=\linewidth]{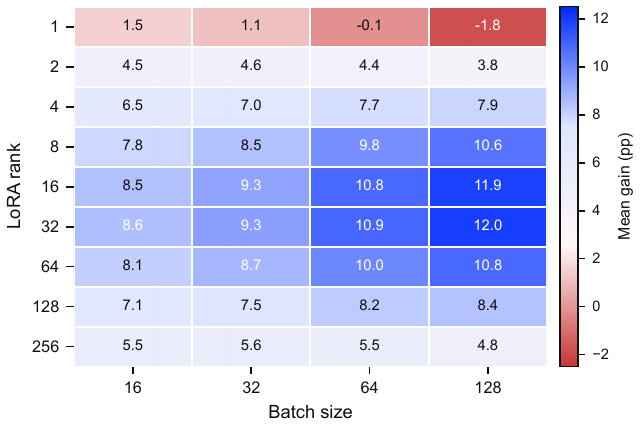}
\caption{Rank--batch gain map.}
\label{fig:rank-overview-heatmap}
\end{subfigure}\hfill
\begin{subfigure}[t]{0.33\textwidth}
\centering
\includegraphics[width=\linewidth]{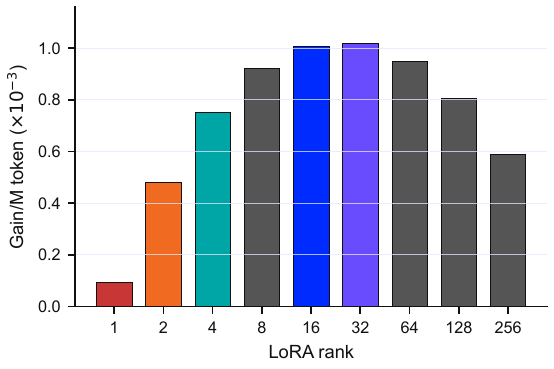}
\caption{Token efficiency.}
\label{fig:rank-overview-efficiency}
\end{subfigure}
\caption{Rank-sweep overview. The Qwen$3$-$8$B PPO sweep separates LoRA rank into low-rank research frontier, middle-rank deployment default, and high-rank cost-warning regimes rather than a monotonic capacity curve.}
\label{fig:rank-sweep-overview}
\end{figure}

A Qwen$3$-$8$B PPO sweep provides the main evidence for this question. The sweep contains $216$ runs across nine LoRA ranks, four batch sizes, and six seeds per configuration, using a fixed $500$-step PPO schedule on a mixed mathematics corpus with verifiable rewards. As summarized in \Cref{fig:rank-sweep-overview}, the resulting behavior separates into three distinct regions rather than a single monotonic scaling law.

\begin{figure}[tbp]
\centering
\begin{subfigure}[t]{0.48\textwidth}
\centering
\includegraphics[width=\linewidth]{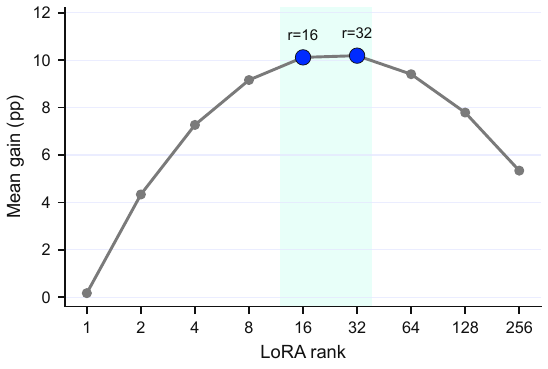}
\caption{Middle-rank peak.}
\label{fig:rank-middle-curve}
\end{subfigure}\hfill
\begin{subfigure}[t]{0.48\textwidth}
\centering
\includegraphics[width=\linewidth]{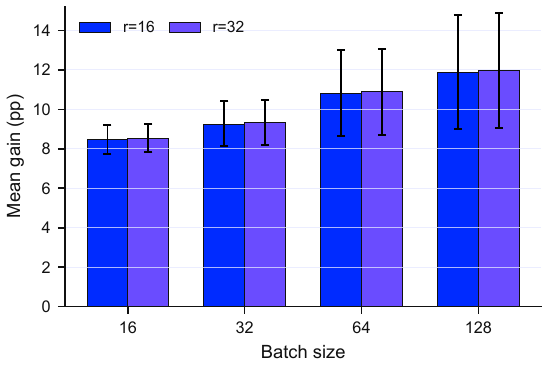}
\caption{Batch behavior at ranks $16$ and $32$.}
\label{fig:rank-middle-batch}
\end{subfigure}

\medskip
\begin{subfigure}[t]{0.48\textwidth}
\centering
\includegraphics[width=\linewidth]{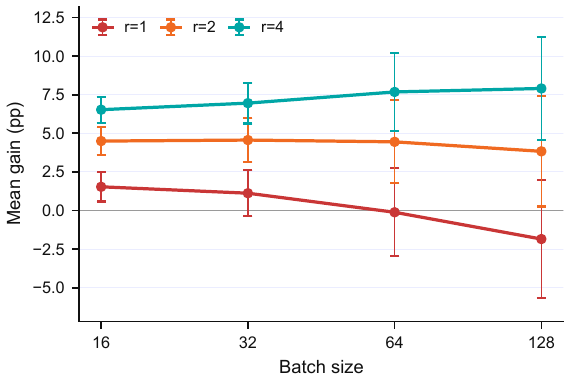}
\caption{Batch sensitivity at ranks $1$--$4$.}
\label{fig:rank-low-batch}
\end{subfigure}\hfill
\begin{subfigure}[t]{0.48\textwidth}
\centering
\includegraphics[width=\linewidth]{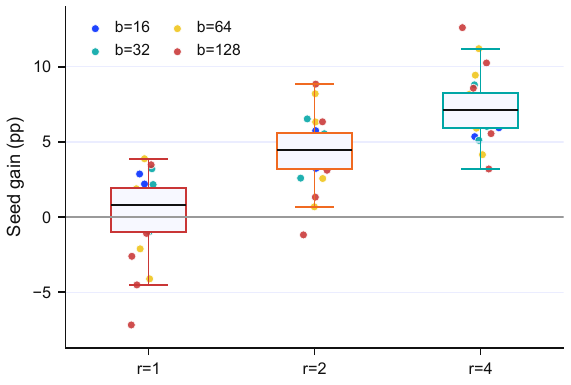}
\caption{Seed-level spread at ranks $1$--$4$.}
\label{fig:rank-low-seed}
\end{subfigure}
\caption{Middle- and low-rank operating regions. \textbf{(a, b)} Ranks $16$ and $32$ provide the strongest practical balance of mean gain and downside risk under the observed PPO recipe. \textbf{(c, d)} Ranks $1$--$4$ still carry positive reinforcement-learning signal, but rank $1$ becomes sharply batch- and seed-sensitive.}
\label{fig:rank-region-detail}
\end{figure}

Ranks $16$ and $32$ form the strongest practical region under the current recipe. They achieve the highest mean gains, maintain relatively low downside risk, and provide strong token efficiency. \Cref{fig:rank-region-detail}~(a, b) isolates this middle-rank band across batch sizes, making it the current deployment default.

The more important implication, however, appears at the low-rank end of the curve. Ranks $1$ to $4$ do not behave like a uniformly failed region. Their best-seed runs are close to those at ranks $16$ and $32$, while their across-seed means dip and their seed-to-seed spread widens. As shown in \Cref{fig:rank-region-detail}~(c, d), extremely small adapters can already carry the same level of reinforcement-learning signal that larger adapters reach, but current training recipes cannot access that signal consistently across seeds.

This distinction substantially changes the interpretation of low-rank PEFT. If the low-rank region were uniformly weak across seeds, the natural conclusion would be that the adapter is simply too small for the task. Instead, the sweep suggests that low rank is under-optimized rather than under-capacity. The best run at rank $1$ already matches the best runs at ranks $16$ and $32$; what collapses at low rank is reliability across seeds, not the height of the attainable ceiling. The dominant failure mode is therefore not insufficient expressivity but insufficient stability, which is the problem the next subsubsection takes up.

\begin{figure}[tbp]
\centering
\begin{subfigure}[t]{0.48\textwidth}
    \centering
    \includegraphics[width=\linewidth]{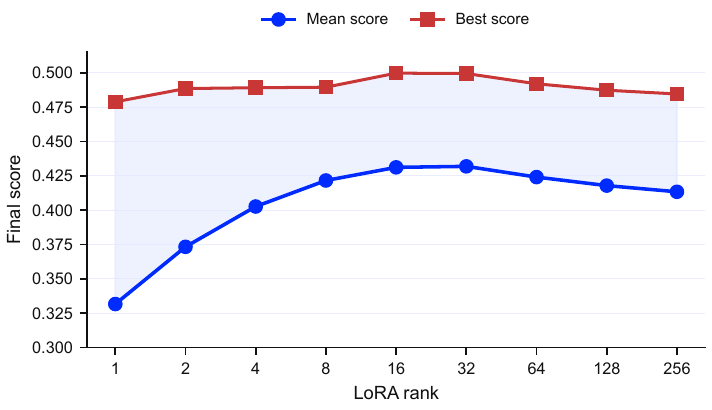}
    \caption{Mean--best separation. The best-seed frontier is nearly flat across ranks, while the across-seed mean degrades at low rank, indicating that the limitation is reliability rather than capacity.}
    \label{fig:rank-mean-best}
\end{subfigure}
\hfill
\begin{subfigure}[t]{0.48\textwidth}
    \centering
    \includegraphics[width=\linewidth]{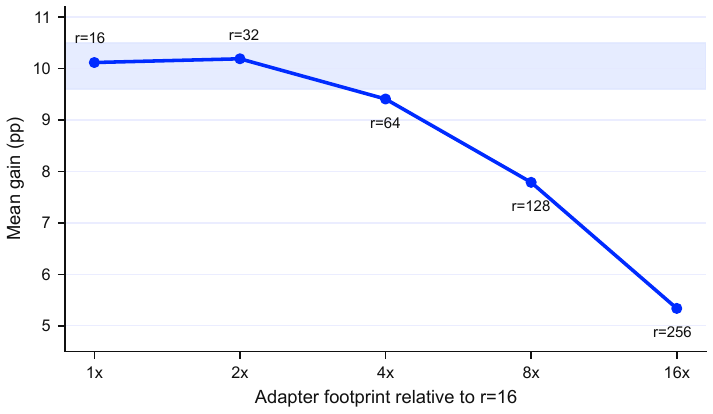}
    \caption{High-rank opportunity cost. Beyond the middle-rank region, adapter footprint increases while the observed performance frontier flattens.}
    \label{fig:rank-high-cost}
\end{subfigure}

\medskip
\begin{subfigure}[t]{0.48\textwidth}
    \centering
    \includegraphics[width=\linewidth]{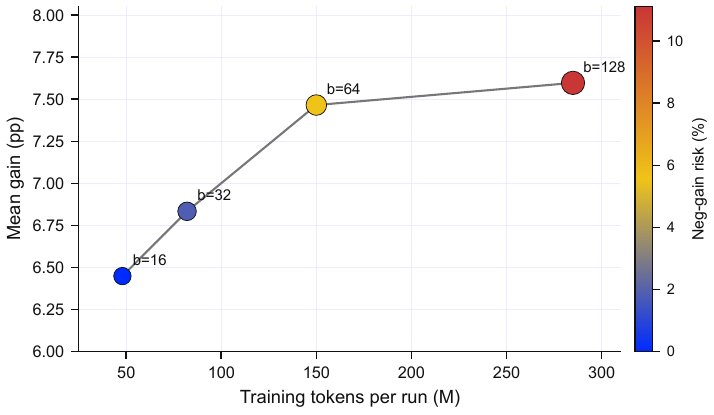}
    \caption{Batch size as a budget variable. Larger batches slightly raise average gain, but they also increase token consumption under the fixed-step PPO schedule.}
    \label{fig:rank-batch-cost}
\end{subfigure}
\hfill
\begin{subfigure}[t]{0.48\textwidth}
    \centering
    \includegraphics[width=\linewidth]{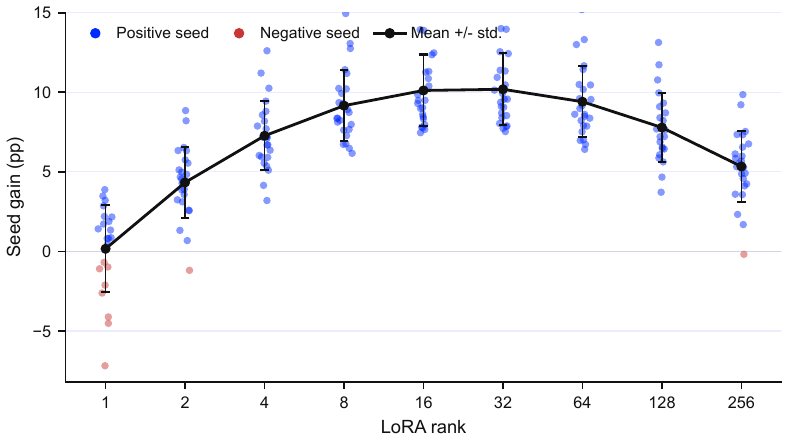}
    \caption{Seed-level reliability. Cheaper adapter configurations make seed-heavy evaluation feasible, which is necessary for separating stable operating regimes from lucky runs.}
    \label{fig:rank-seed-reliability}
\end{subfigure}
\caption{Reliability and cost structure of the rank sweep. Mean--best separation and high-rank opportunity cost characterize where added rank stops paying off, while batch-cost and seed-reliability views show why cheaper adapters enable the seed-heavy evaluation that low-rank reliability requires.}
\label{fig:rank-reliability-cost}
\end{figure}

% \begin{figure}[tbp]
% \centering
% \includegraphics[width=0.76\textwidth]{figures/scale_down/rank_reduction_mean_best.pdf}
% \caption{Mean--best separation. Low-rank adapters can occasionally reach strong runs even when their mean behavior remains weak, making reliability a recipe problem as much as a capacity problem.}
% \label{fig:rank-mean-best}
% \end{figure}

This interpretation rests on a careful reading of mean versus best-run performance. Mean score measures whether a configuration is dependable across seeds, while best score measures whether the regime can ever reach a strong solution under the current recipe. \Cref{fig:rank-mean-best} shows that the best-seed frontier remains nearly flat across ranks $1$ through $32$, while the across-seed mean drops at low rank. Strong best runs with weak mean runs indicate an optimization and initialization bottleneck rather than a hard capacity limit, and motivate spending the next subsubsection on RL-native initialization rather than on simply raising rank.

% \begin{figure}[tbp]
% \centering
% \includegraphics[width=0.72\textwidth]{figures/scale_down/rank_reduction_high_cost.pdf}
% \caption{High-rank opportunity cost. Beyond the middle-rank region, adapter footprint increases while the observed performance frontier flattens.}
% \label{fig:rank-high-cost}
% \end{figure}

The high-rank region exposes the opposite problem. Increasing rank from the middle region to $64$, $128$, or $256$ inflates trainable parameters, optimizer state, checkpoint size, and memory pressure, but does not extend the observed best-run frontier in the sweep. \Cref{fig:rank-high-cost} makes this tradeoff explicit: beyond the middle-rank region, larger adapters add footprint without raising the ceiling. Under a fixed research or deployment budget, unnecessary rank displaces more useful expenditures such as additional seeds, broader ablations, tighter variance control, reward debugging, and rank-specific optimization.

% \begin{figure}[tbp]
% \centering
% \includegraphics[width=0.76\textwidth]{figures/scale_down/rank_reduction_batch_cost.pdf}
% \caption{Batch size as a budget variable. Larger batches slightly raise average gain, but they also increase token consumption under the fixed-step PPO schedule.}
% \label{fig:rank-batch-cost}
% \end{figure}

Batch size must be interpreted as part of the same budgeted adaptation regime. Because the sweep fixes PPO steps rather than total training tokens, larger batches directly increase token consumption. \Cref{fig:rank-batch-cost} shows that larger batches slightly improve average gain, but at sharply higher token cost and with increased downside risk. Batch size is therefore not a pure optimization knob, but part of the broader score--cost tradeoff.

The resulting scale-down lesson is reflexive. Smaller adapters matter not only because they reduce the cost of one trained model, but because they reduce the cost of searching for better recipes. Tiny adapters whose best runs are already competitive but whose means are unreliable require exactly the kind of repeated experimentation that cheaper adapters enable: seed-heavy evaluation, stronger initialization, cleaner schedules, tighter KL control, and rank-specific hyperparameters. \Cref{fig:rank-seed-reliability} illustrates this feedback loop. Cheaper runs make broader evidence collection possible, and broader evidence collection is what turns lucky low-rank runs into reliable ones.

% \begin{figure}[tbp]
% \centering
% \includegraphics[width=0.74\textwidth]{figures/scale_down/rank_reduction_seed_reliability.pdf}
% \caption{Seed-level reliability. Cheaper adapter configurations make seed-heavy evaluation feasible, which is necessary for separating stable operating regimes from lucky runs.}
% \label{fig:rank-seed-reliability}
% \end{figure}

For a population of personal models, this distinction becomes system-level. A reduction in rank may appear modest for a single adapter, but it compounds across repeated updates, optimizer states, checkpoints, serving-time adapter loads, and the millions of persistent model instances that share a common prior. The objective is not to claim that rank $1$ is already sufficient. It is to identify a path along which the per-instance adaptive boundary can become smaller without becoming brittle, so that the population as a whole stays close to the ``share 99.5\%, differ in 0.5\%'' regime that makes personal models economically feasible.

\begin{figure}[ht]
\centering
\begin{subfigure}[t]{0.48\textwidth}
\centering
\includegraphics[width=\linewidth]{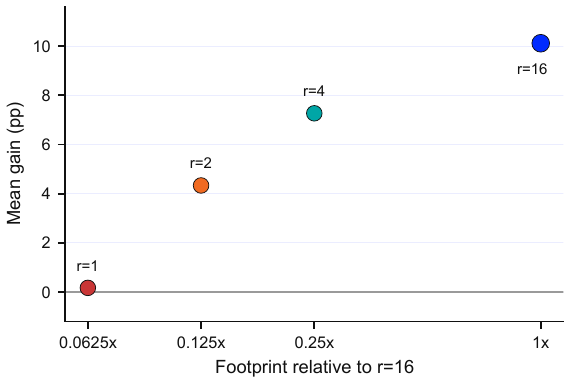}
\caption{Minimal footprint versus gain.}
\label{fig:rank-tiny-efficiency}
\end{subfigure}\hfill
\begin{subfigure}[t]{0.48\textwidth}
\centering
\includegraphics[width=\linewidth]{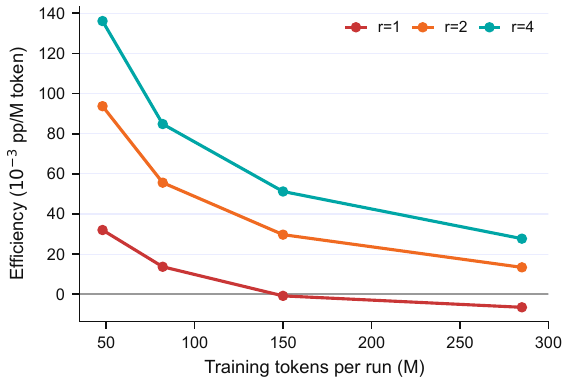}
\caption{Low-rank token efficiency.}
\label{fig:rank-tiny-budget}
\end{subfigure}
\caption{Tiny-adapter objective. The low-rank research target is not maximum single-run score alone, but stable learning signal under minimal trainable state and token budget.}
\label{fig:rank-tiny-adapter}
\end{figure}

\Cref{fig:rank-tiny-adapter} reframes this objective explicitly. The low-rank frontier should be evaluated not only by maximum single-run score, but also by whether useful learning signal can be made reliable under minimal trainable state and token budget. Under the current evidence, ranks $16$ to $32$ remain the practical default, ranks $1$ to $4$ define the research frontier where best-run gains are already there but mean reliability is not, and ranks above $64$ warn against treating larger adapters as the default direction of progress. Rank reduction is therefore an operating-regime problem in which expressivity, optimization, variance, and budget must be solved together rather than independently. The most immediate lever is the initialization geometry of the few directions a tiny adapter does have, which the next subsubsection addresses directly.

% \begin{figure}[tbp]
% \centering
% \includegraphics[width=0.84\textwidth]{figures/scale_down/rank_reduction_operating_regime.pdf}
% \caption{Rank reduction as an operating-regime problem. Expressivity, variance control, optimization, and budget must be solved together rather than optimized as independent knobs.}
% \label{fig:rank-operating-regime}
% \end{figure}

\subsubsection{RL-Native Initialization for Stability}
\label{subsubsec:initialization}

The rank-sweep results (\Cref{fig:rank-overview-efficiency} (b)) above show that extremely small adapters are not uniformly useless, but
they are fragile. This fragility is most visible at rank \(r=1\). A rank-one LoRA adapter has only
one adaptive direction per weight matrix. If this direction is poorly chosen, the adapter cannot
redistribute learning across alternative components. Standard LoRA initializes this direction
randomly, which is often sufficient at moderate ranks, but becomes unreliable when the adaptive
subspace collapses to a single dimension. The natural question is therefore not only how small
the rank can be, but how the few available directions should be initialized.

A natural first attempt is to use the geometry of the pretrained weight matrix. If the adapter has
only one or a few directions, those directions should not be wasted on arbitrary random axes.
The pretrained matrix already contains a spectral structure, and its singular vectors provide
candidate directions that may be more meaningful than random Gaussian rows. From this
perspective, geometry-aware initialization is an attractive way to rescue the rank-one regime:
Instead of increasing the rank, we try to make the available direction more useful.

\begin{figure}[tbp]
\centering
\includegraphics[width=0.7\linewidth]{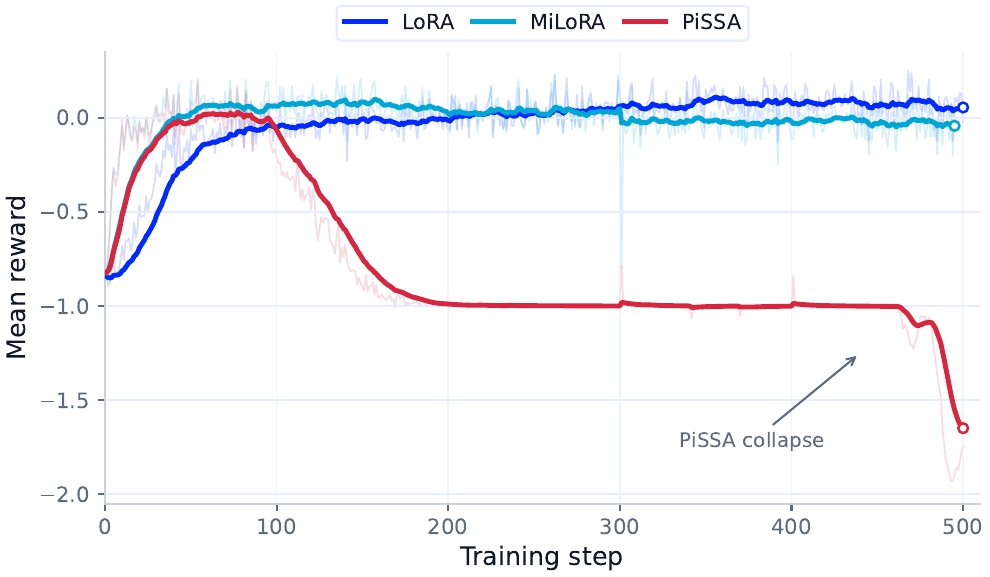}
\caption{Training reward curves for DAPO 1.5B. PiSSA, MiLoRA, and LoRA are compared over 500 training steps using the same reward metric.}
\label{fig:pissa_milora}
\end{figure}

However, a recent study \citep{yin2025evaluatingparameterefficientmethods} shows that existing SVD-based LoRA initializations cannot be transferred
from supervised fine-tuning to reinforcement learning naively. PiSSA \citep{meng2025pissa} initializes adapters using
the principal singular directions, while MiLoRA\citep{wang2025milora} uses the minor singular directions. Both methods
use pretrained spectral information and can be effective in supervised fine-tuning, where dense
token-level supervision rewards fast convergence. In RL with verifiable rewards, however, these
same methods may underperform standard LoRA and exhibit training collapse. We replicate this phenomenon as shown in \Cref{fig:pissa_milora}.

% The issue is not simply
% that they use geometry. Rather, they combine subspace selection with singular-value scaling,
% which can make early updates too aggressive for RL's local policy-improvement regime. By using cosine learning rate decay, we observe 

 % We train DeepSeek-R1-Distill-Qwen-1.5B with DAPO on the DAPO-Math-17k dataset for 500 steps using a constant learning rate of $1 \times      
 %  10^{-5}$, with per-device batch size 4, gradient accumulation steps 8 (effective batch size 32), and 8 rollout generations per prompt. We    
 %  compare three PEFT adapters applied to all attention and MLP projection layers (${$\texttt{q, k, v, o, up, down, gate_proj}$}$): LoRA        
 %  ($r{=}16$, $\alpha{=}32$), PiSSA ($r{=}32$, $\alpha{=}64$), and MiLoRA ($r{=}32$, $\alpha{=}64$). 

This motivates an RL-native initialization criterion. A useful initialization should expose a
meaningful low-dimensional learning direction, but it should not make the first policy updates
too large. In supervised fine-tuning, initialization mainly affects convergence speed. The loss is
dense, token-level, and anchored to fixed target sequences, so structured initializations can be
useful because the optimizer receives many reliable gradient steps. In RL with verifiable rewards,
the optimization geometry is more restrictive. The model learns from sampled responses, and
practical objectives rely on token-level surrogates around the rollout policy. These surrogates
remain meaningful only when the updated policy stays close to the policy that generated the
samples. KL penalties, clipping, and trust-region mechanisms therefore do more than regularize:
they define the local region in which RL updates are trustworthy.

The reason is easiest to see from the sequence-level objective. The RLVR objective can be
written as \(J(\theta)=\mathbb{E}_{x\sim\mathcal{D},\,y\sim\pi_\theta(\cdot\mid x)}[R(x,y)]\),
where \(R(x,y)\in\{0,1\}\) is assigned to a complete response. Since
\(\pi_\theta(y\mid x)=\prod_{t=1}^{T}\pi_\theta(y_t\mid x,y_{<t})\), the REINFORCE
gradient \citep{williams1992reinforce} distributes the sequence reward across token positions:
\[
    \nabla_\theta J(\theta)
    =
    \mathbb{E}_{y\sim\pi_\theta}
    \biggl[
        R(x,y)
        \sum_{t=1}^{T}
        \nabla_\theta \log \pi_\theta(y_t\mid x,y_{<t})
    \biggr].
\]
In deployed RL systems, responses are often sampled from an inference-side rollout policy
\(\mu\), while updates are computed for \(\pi_\theta\). This gives the importance-sampled form
\[
    \nabla_\theta J(\theta)
    =
    \mathbb{E}_{y\sim\mu}
    \biggl[
        R(x,y)
        \frac{\pi_\theta(y\mid x)}{\mu(y\mid x)}
        \sum_{t=1}^{T}
        \nabla_\theta \log \pi_\theta(y_t\mid x,y_{<t})
    \biggr].
\]
The sequence-level importance weight decomposes as
\[
    \frac{\pi_\theta(y\mid x)}{\mu(y\mid x)}
    =
    \prod_{t=1}^{T}
    \frac{\pi_\theta(y_t\mid x,y_{<t})}
         {\mu(y_t\mid x,y_{<t})}
    =
    \prod_{t=1}^{T}r_t
    =
    \prod_{t=1}^{T}(1+\delta_t),
\]
where \(\delta_t=r_t-1\). This product is exponentially unstable: if \(T=512\) and each
\(r_t=1.01\), the sequence weight is approximately \(1.01^{512}\approx163\). Practical
token-level surrogates therefore depend on the first-order Taylor expansion
\[
    \prod_{t=1}^{T}(1+\delta_t)
    =
    1+\sum_{t=1}^{T}\delta_t
    +
    O(\delta^2),
\]
which is accurate only when the rollout and updated policies remain close. Dropping higher-order
terms gives the approximate token-level objective \citep{zheng2025stabilizingrl}
\[
    \nabla_\theta J(\theta)
    \approx
    \mathbb{E}_{y\sim\mu}
    \biggl[
        R(x,y)
        \sum_{t=1}^{T}
        w_t
        \nabla_\theta \log \pi_\theta(y_t\mid x,y_{<t})
    \biggr],
    \qquad
    w_t=
    \frac{\pi_\theta(y_t\mid x,y_{<t})}
         {\mu(y_t\mid x,y_{<t})}.
\]

This approximation explains why RL fine-tuning is sensitive to early policy movement. To second
order, the KL divergence between the updated and current policies satisfies \citep{zhu2025pathtakenrlvrprovably}

\[
    D_{\mathrm{KL}}
    \bigl(
        \pi_{\theta+\Delta\theta}
        \,\|\,
        \pi_\theta
    \bigr)
    \approx
    \frac{1}{2}
    \Delta\theta^\top F\Delta\theta,
    \qquad
    F=
    \mathbb{E}
    \bigl[
        \nabla_\theta\log\pi_\theta
        \nabla_\theta\log\pi_\theta^\top
    \bigr].
\]
If one update satisfies
\(D_{\mathrm{KL}}(\pi_{\theta^+}\,\|\,\pi_\theta)\le K\) and
\(F(\theta)\succeq mI\) on the update subspace, then a block update obeys
\[
    \|\Delta W\|_F
    \le
    \sqrt{\frac{2K}{m}}
    \bigl(1+o(1)\bigr).
\]
This creates a KL leash introduced by \citep{zhu2025pathtakenrlvrprovably}: a policy-space trust region induces a weight-space movement budget.
RLVR must therefore make progress through small, controlled steps rather than large jumps.
Initialization does not merely choose where optimization starts, but biases the direction of the
entire trajectory inside this narrow local region. A bad initialization can push the optimizer
toward directions where small parameter movements create large policy drift, causing the
first-order approximation above to break down.

This is why SVD-scaled initializations are risky in the tiny-adapter regime. At rank \(r=1\), a
geometry-aware direction is valuable, but singular-value amplification can consume the KL
budget before useful learning stabilizes. The design target is therefore not geometry alone, but
controlled geometry: keep a meaningful pretrained direction while removing the scaling that
makes the early update overly aggressive \citep{zhang2026geometry}.

We instantiate this idea with \textbf{OLoRA-tail}. Let $ W_0 = U\Sigma V^\top$
be the singular value decomposition of a pretrained weight matrix, and let \(U_{-r}\) and
\(V_{-r}\) denote the left and right singular vectors associated with the smallest \(r\) singular
values. OLoRA-tail initializes
\[
    B_0 = U_{-r},
    \qquad
    A_0 = V_{-r}^{\top}.
\]

OLoRA-tail differs from MiLoRA in the detail that matters most for RL stability. MiLoRA uses
\[
    B_0 = U_{-r}\Sigma_{-r}^{1/2},
    \qquad
    A_0 = \Sigma_{-r}^{1/2}V_{-r}^{\top},
\]
which injects singular-value scaling into both LoRA factors. OLoRA-tail keeps the same gentle tail
subspace but removes this scaling. Thus, it preserves pretrained geometry while avoiding the
extra singular value scaling that can destabilize RL. 

\begin{figure}[t]                         \centering                         \includegraphics[width=\linewidth]{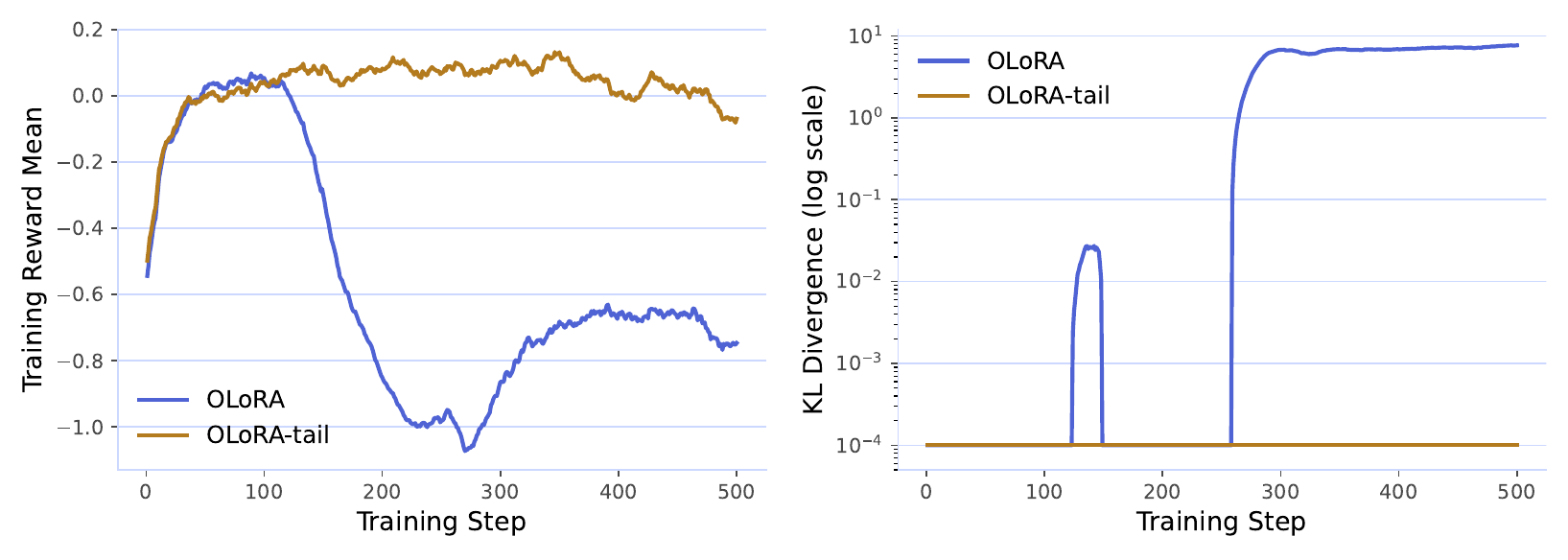}                       \caption{
          Training reward (left) and KL divergence from the reference model (right, log scale)                                
          for OLoRA and OLoRA-tail on DeepSeek-R1-Distill-Qwen-1.5B trained with DAPO.
          OLoRA collapses around step 100, with reward dropping to $-1.0$ and KL divergence
          exploding to ${\sim}8$, while OLoRA-tail remains stable throughout 500 steps.
      }
      \label{fig:olora_vs_olora_tail}
  \end{figure}

 Although both OLoRA \citep{buyukakyuz2024oloraorthonormallowrankadaptation} and OLoRA-tail apply orthogonal initialization to the adapter matrices,
  they differ in {which} singular subspace is used.
  OLoRA initializes adapters from the principal singular vectors of the pre-trained weights,
  whereas OLoRA-tail initializes from the minor singular vectors.
  As shown in \Cref{fig:olora_vs_olora_tail}, OLoRA undergoes training collapse under the
  DAPO objective: its reward deteriorates sharply after step 100 and its KL divergence from the
  reference model grows by several orders of magnitude, eventually saturating near $8$.
  We attribute this instability to the fact that the principal subspace captures the dominant
  directions of the pre-trained representation, and perturbing these directions through on-policy
  RL updates induces large distributional shifts that the DAPO clipping mechanism cannot
  fully contain.
  By contrast, the minor subspace spans directions that are comparatively inert in the
  pre-trained model, providing a safer optimization landscape: the adapter updates remain
  orthogonal to the most sensitive weight directions, keeping KL divergence near zero and
  training stable throughout.
  Based on this observation, we adopt OLoRA-tail for
  all subsequent scale-down experiments.

\begin{figure}[tbp]                  
      \centering                   \includegraphics[width=0.62\linewidth]{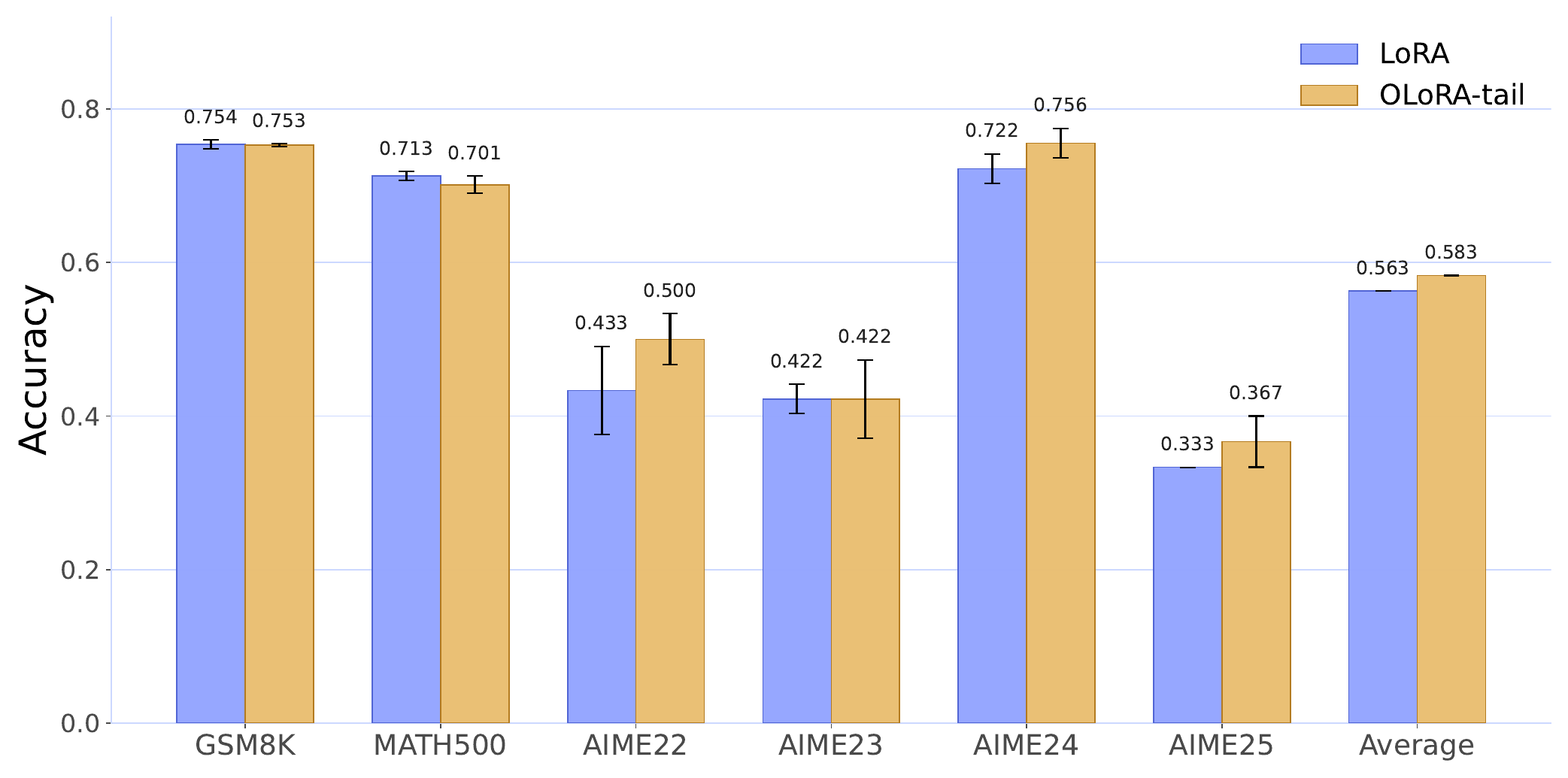}            
      \caption{   
          Comparison of LoRA and OLoRA-tail on mathematical reasoning benchmarks. OLoRA-tail consistently matches or outperforms LoRA,
   achieving a higher average accuracy ($58.3\%$ vs.\ $56.3\%$).
      }
      \label{fig:lora_vs_olora_minor}
  \end{figure}

 We train DeepSeek-R1-Distill-Qwen-1.5B with DAPO on the DAPO-Math-17k dataset for 500 steps, using a constant learning rate 
  of $1\times10^{-5}$ and an effective batch size of 32. Both LoRA and OLoRA-tail are applied to all attention and MLP        
  projection layers with rank $r{=}16$ and scaling factor $\alpha{=}32$. We evaluate LoRA and OLoRA-tail on six mathematical reasoning benchmarks using                  
DeepSeek-R1-Distill-Qwen-1.5B trained with DAPO. As shown in \Cref{fig:lora_vs_olora_minor}, OLoRA-tail achieves a higher average accuracy of $58.3 \%$  compared to LoRA's $56.3\%$, demonstrating that a better geometric initialization can provide an advantage. 

Having established that OLoRA-tail's minor-subspace initialization yields consistent gains over standard LoRA at rank       
  $r{=}16$, a natural question is whether this geometric advantage persists as we push the adapter to its most extreme
  compression. Rank-one adaptation represents the smallest possible trainable footprint—a single outer-product update per     
  weight matrix, yet it is notoriously sensitive to initialization: with only one direction available, a poor choice of
  singular vector can permanently misalign the adapter with the task signal. The stability benefits we observed at rank 16    
  suggest that anchoring the adapter in the minor subspace may be precisely what is needed to make rank-one adaptation viable.
   This gives a clean scale-down test: can a better one-dimensional geometry make rank-one adaptation usable without
  increasing the trainable parameter budget?

We evaluate OLoRA-tail at the extreme compression of rank $r{=}1$ across two model
  scales: Qwen3-8B and Qwen3-30B-A3B-Instruct, under the same experimental settings in Section \Cref{subsubsec:rank-reduction}

\begin{figure}[t]
      \centering
      \includegraphics[width=0.9\linewidth]{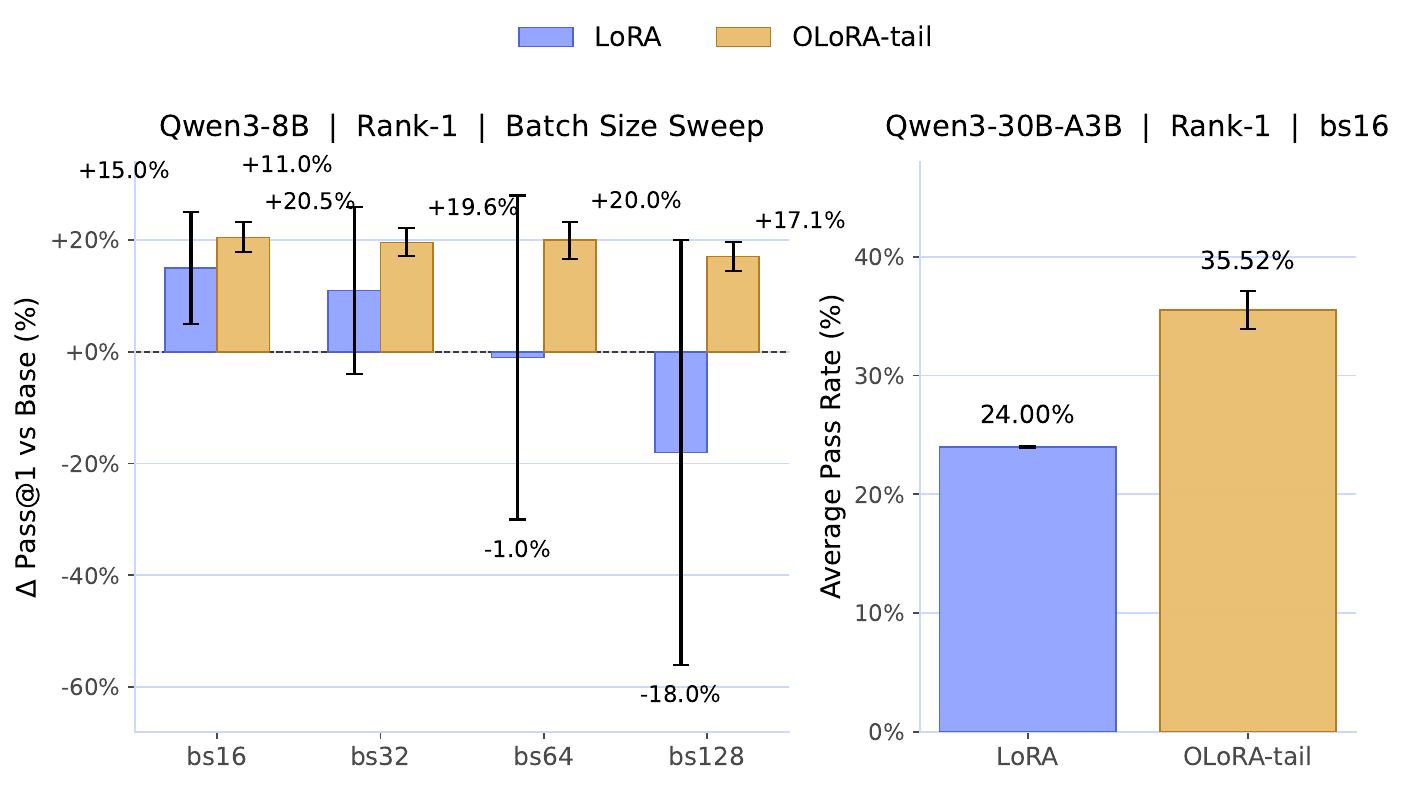}
      \caption{
          Rank-1 OLoRA-tail versus LoRA on Qwen3-8B and Qwen3-30B-A3B-Instruct trained
          with DAPO.
          \textbf{Left}: $\Delta$Pass@1 over the base model across batch sizes on Qwen3-8B;
          OLoRA-tail maintains a consistent gain of ${\sim}{+}20\%$ regardless of batch size,
          while LoRA degrades from $+15\%$ at bs16 to $-18\%$ at bs128.
          \textbf{Right}: Average pass rate on Qwen3-30B-A3B-Instruct;
          OLoRA-tail ($35.5\%$) surpasses LoRA ($24.0\%$) by $11.5$ percentage points.
          Error bars show standard deviation across 6 random seeds.
      }
      \label{fig:rank1_8b_30b}
  \end{figure}
  As shown in \Cref{fig:rank1_8b_30b}, the results demonstrate that this distinction
  is substantial.
  On Qwen3-8B, OLoRA-tail delivers a consistent ${\sim}{+}20\%$ gain over the base model
  across all batch sizes, while standard LoRA degrades sharply with increasing batch size:
  from $+15\%$ at bs16 to $-18\%$ at bs128, with collapse risk reaching $67\%$.
  The geometric advantage of minor-subspace initialization does not merely narrow the
  performance gap, but it eliminates the sensitivity to batch size entirely.
  On Qwen3-30B-A3B-Instruct, OLoRA-tail achieves an average pass rate of
  $35.5\%{\pm}1.6\%$, surpassing the LoRA baseline of $24.0\%$ by $11.5$ percentage
  points ($+48\%$ relative).

  These results refine the interpretation of rank.
  Rank determines the number of available directions, but initialization determines whether
  those directions are usable.
  In the tiny-adapter regime, adding rank is one way to increase the probability of finding
  a useful direction. Choosing the direction geometrically is another, and our results show
  it can be strictly more parameter-efficient.
  OLoRA-tail improves the adapter's usable capacity without increasing trainable weights,
  optimizer state, checkpoint size, or serving-time footprint, which is precisely the
  scale-down property that matters for sustaining large populations of personal models.
  The practical implication is not that every task should use $r{=}1$, but that the lower
  boundary of viable adaptation can be pushed further when initialization is designed for
  the KL-constrained, on-policy operating regime of RL fine-tuning.

\subsubsection{Reusable Hyperparameters for Lower Training Effort}\label{subsubsec:hyperparameter-reuse}

Hyperparameter transfer becomes a scaling bottleneck once adapters are trained at population scale. LoRA exposes at least three tightly coupled knobs: rank $r$, scale $\alpha$, and learning rate $\eta$. In practice, users often change rank because of memory, speed, or capacity constraints and then retune the learning rate independently. The more useful question, however, is not \textit{what is the best LoRA learning rate?}  but \textit{how should the learning-rate search move when rank changes?} The answer depends directly on how $\alpha$ scales with rank.

LoRA commonly parameterizes the update as
\begin{equation}
 \Delta W = \frac{\alpha_r}{r}BA.
\end{equation}
Under standard initialization, $A$ is random and $B=0$, so the first effective movement comes from updating $B$. With learning rate $\eta$, the first update to $B$ scales as $\eta \alpha_r/r$, and substituting that back into $\Delta W$ yields an early-step perturbation proportional to
\begin{equation}
 \eta \frac{\alpha_r^2}{r}.
\end{equation}

\begin{figure}[tbp]
\centering
\includegraphics[width=0.85\textwidth]{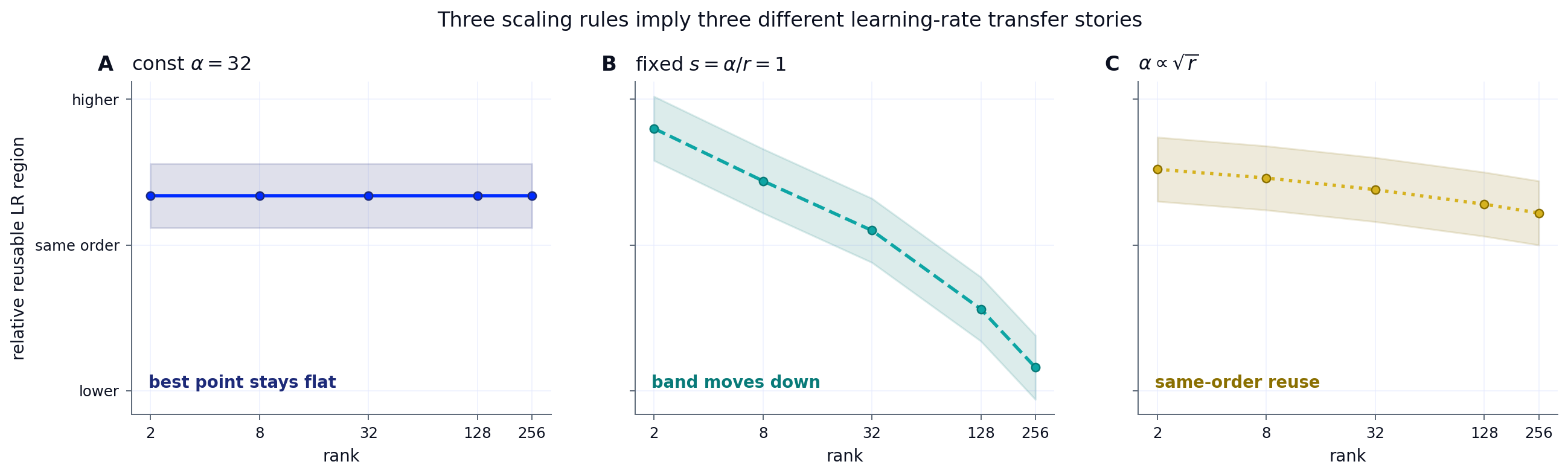}
\caption{The practical summary is that learning-rate transfer depends on the chosen alpha-scaling rule.}\label{fig:triquetra-summary}
\end{figure}

Although this expression does not fully characterize AdamW dynamics, it captures the leading dependence of early update magnitude on rank and scaling convention. If $\alpha/r$ is fixed, then $\alpha_r \propto r$ and the effective update grows with rank, so higher rank should push the search toward smaller learning rates. If $\alpha$ is fixed, the effective update shrinks with rank, so higher rank does not force a smaller learning rate. If $\alpha \propto \sqrt{r}$, rank dependence cancels, yielding the simplest theoretical path to same-order learning-rate reuse.

\begin{figure}[tbp]
\centering
\begin{subfigure}[t]{0.49\textwidth}
\centering
\includegraphics[width=\linewidth]{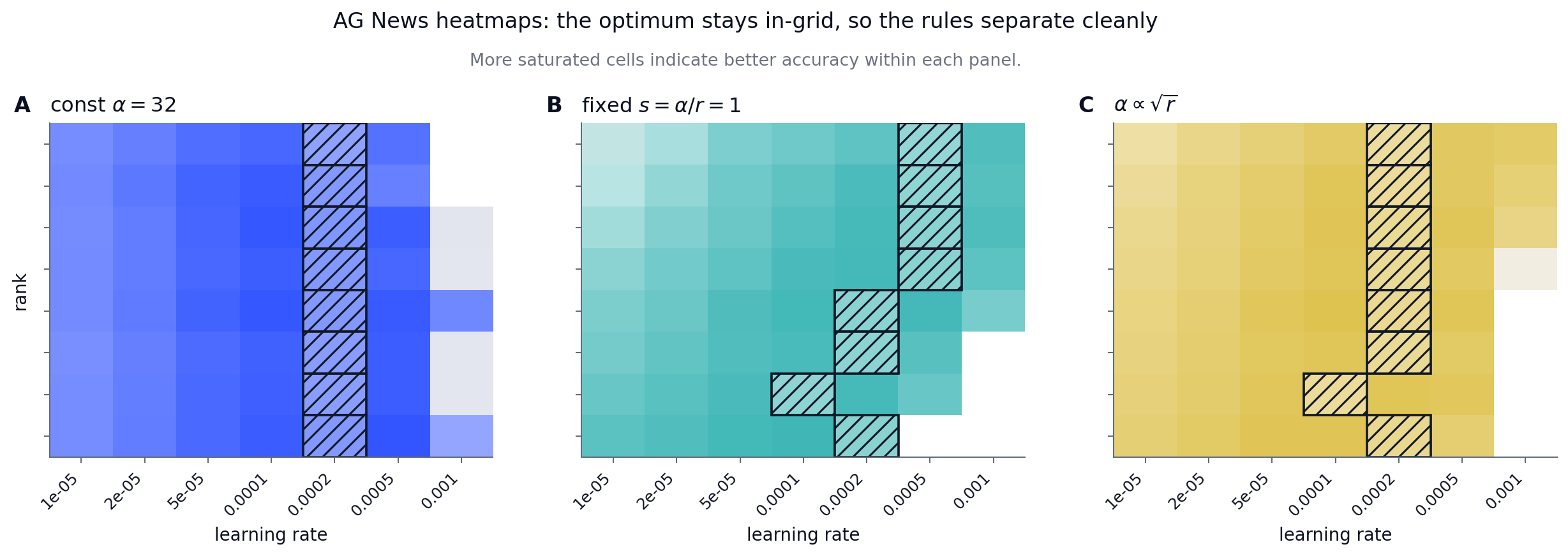}
\caption{Rank-by-learning-rate heatmaps show how the good region moves under different alpha rules.}\label{fig:triquetra-agnews-heatmap}
\end{subfigure}\hfill
\begin{subfigure}[t]{0.49\textwidth}
\centering
\includegraphics[width=\linewidth]{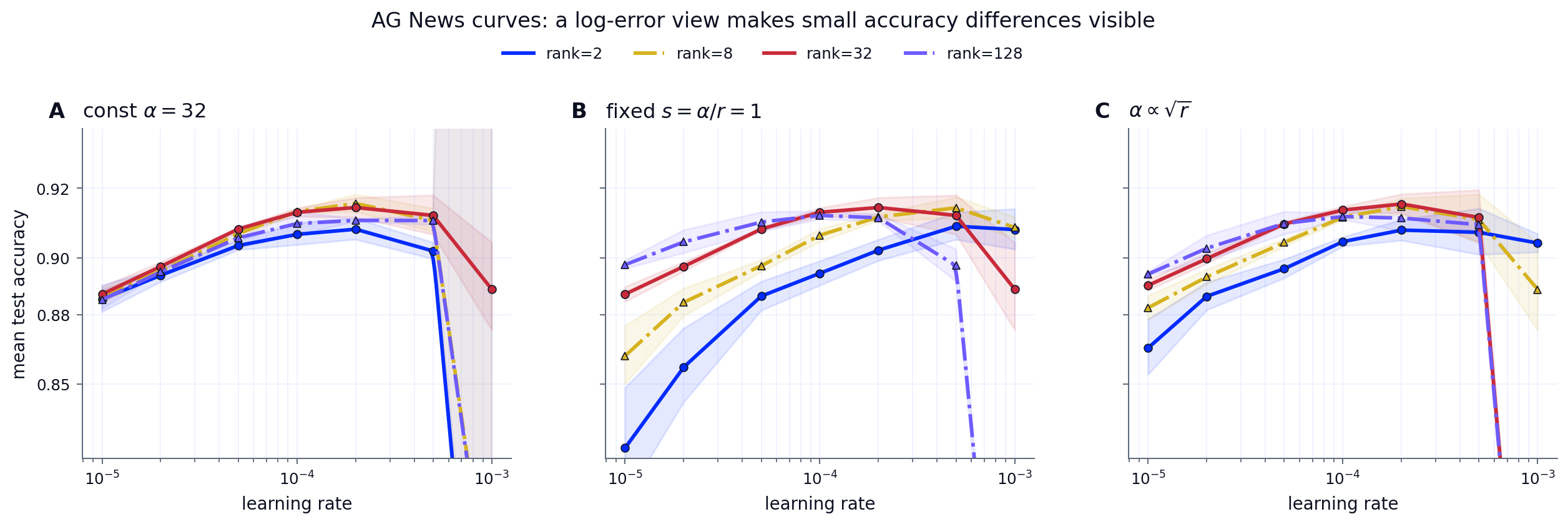}
\caption{Curves separate best-point reuse from the quality of transferred settings.}\label{fig:triquetra-agnews-curves}
\end{subfigure}
\caption{AG News learning-rate transfer across ranks. Heatmaps locate the good learning-rate region, while curves distinguish best-point reuse from transfer quality under each alpha rule.}\label{fig:triquetra-agnews}
\end{figure}

Empirical sweeps show the same split. On AG News \citep{zhang2015charcnn} with DistilBERT \citep{sanh2020distilbert}, ranks 2 through 256 were tested under constant $\alpha=32$, fixed $\alpha/r=1$, and $\alpha \propto \sqrt{r}$ with $\alpha/\sqrt{r}=8$. Fixed $\alpha/r$ shifted the good learning-rate region downward as rank increased. Constant $\alpha$ produced a flatter best-learning-rate location and was easiest to tune in that simple cross-rank setting. The square-root rule preserved same-order reuse in line with the early-training proxy, though constant $\alpha$ was slightly flatter in that particular experiment.

A harder Qwen3-4B MATH transfer setting \citep{yang2025qwen3,hendrycks2021math} makes the transfer problem more consequential. The useful region collapses into a very small learning-rate range, and the scaling rules diverge at high rank. Constant $\alpha$ keeps the best point flat, but it is not the strongest rule overall. Fixed $\alpha/r$ can perform well at low rank, especially with aggressive ratios, but deteriorates quickly as rank grows. The square-root rule is the most balanced: it preserves best-point reuse and delivers stronger high-rank behavior. This distinction matters because best-point reuse and transfer quality are not identical. A rule can keep the same nominal best learning rate while producing weaker performance when reused across ranks.

For population-scale PEFT, the practical object is not one magic learning rate but a reusable band. Fixed $\alpha/r$ pushes the band downward with rank, constant $\alpha$ often keeps it flatter, and $\alpha \propto \sqrt{r}$ preserves same-order reuse while remaining stronger in the harder reasoning setting. A world with millions of adapters is impossible if every new adapter requires a full hyperparameter sweep. Reusable bands and rank-stable parameterizations are therefore not conveniences. They are prerequisites for adapter populations.

The square-root scaling rule was proposed by rsLoRA \citep{kalajdzievski2023rslora} as a theoretical fix for the rank-dependent update magnitude in standard LoRA. Triquetra's contribution is not to introduce this rule, but to validate it empirically in the harder RL fine-tuning setting and to show that the choice of alpha rule has practical consequences for transfer quality, not just for theoretical update magnitude. On simple classification tasks, constant $\alpha$ can be equally flat; on harder reasoning tasks such as Qwen3-4B MATH, the square-root rule is more robust when the reusable band is narrow and transfer quality matters.

\begin{figure}[tbp]
\centering
\begin{subfigure}[t]{0.49\textwidth}
\centering
\includegraphics[width=\linewidth]{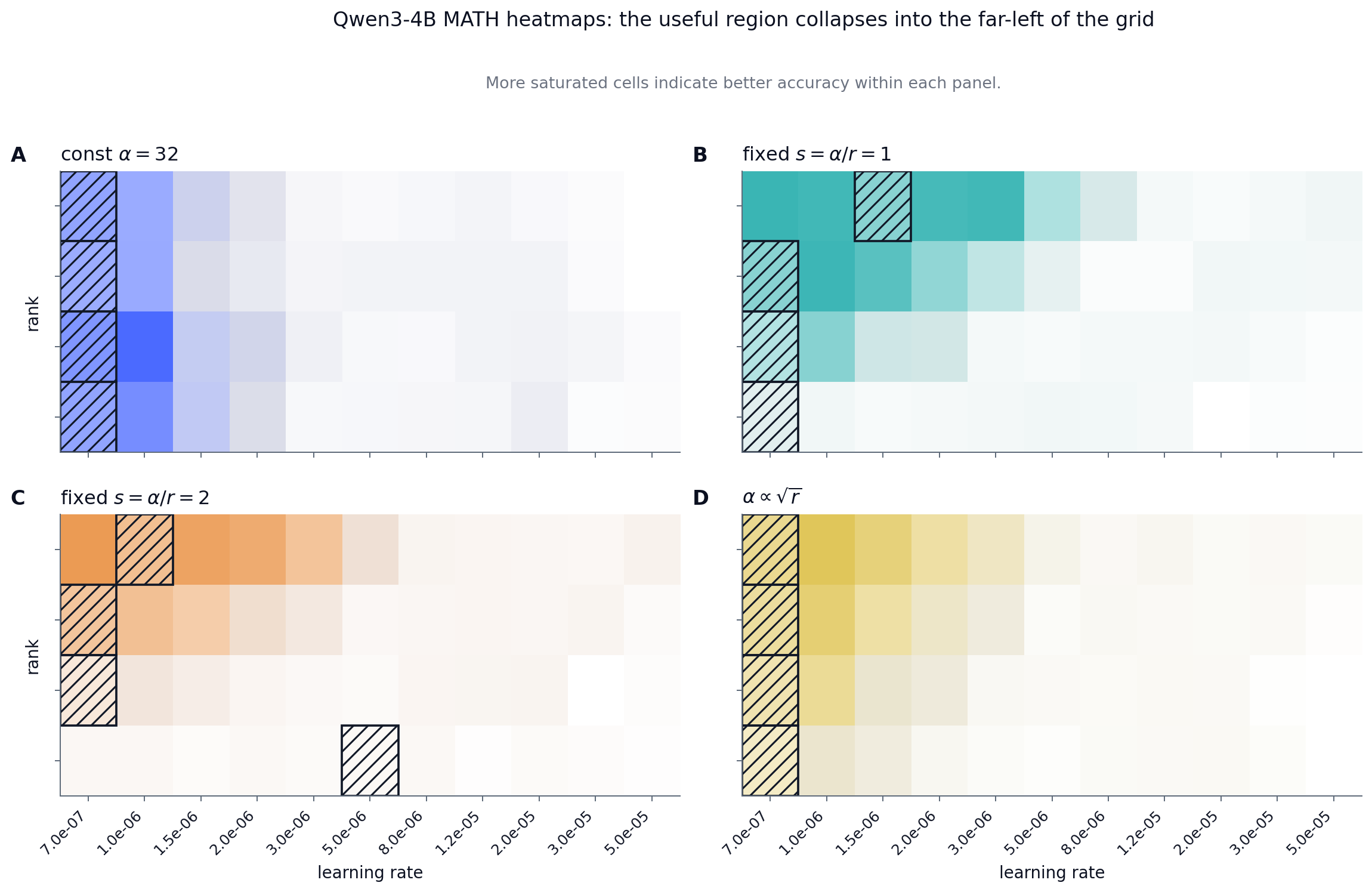}
\caption{Heatmaps show that harder reasoning settings make the reusable band narrower and more consequential.}\label{fig:triquetra-qwen-heatmap}
\end{subfigure}\hfill
\begin{subfigure}[t]{0.49\textwidth}
\centering
\includegraphics[width=\linewidth]{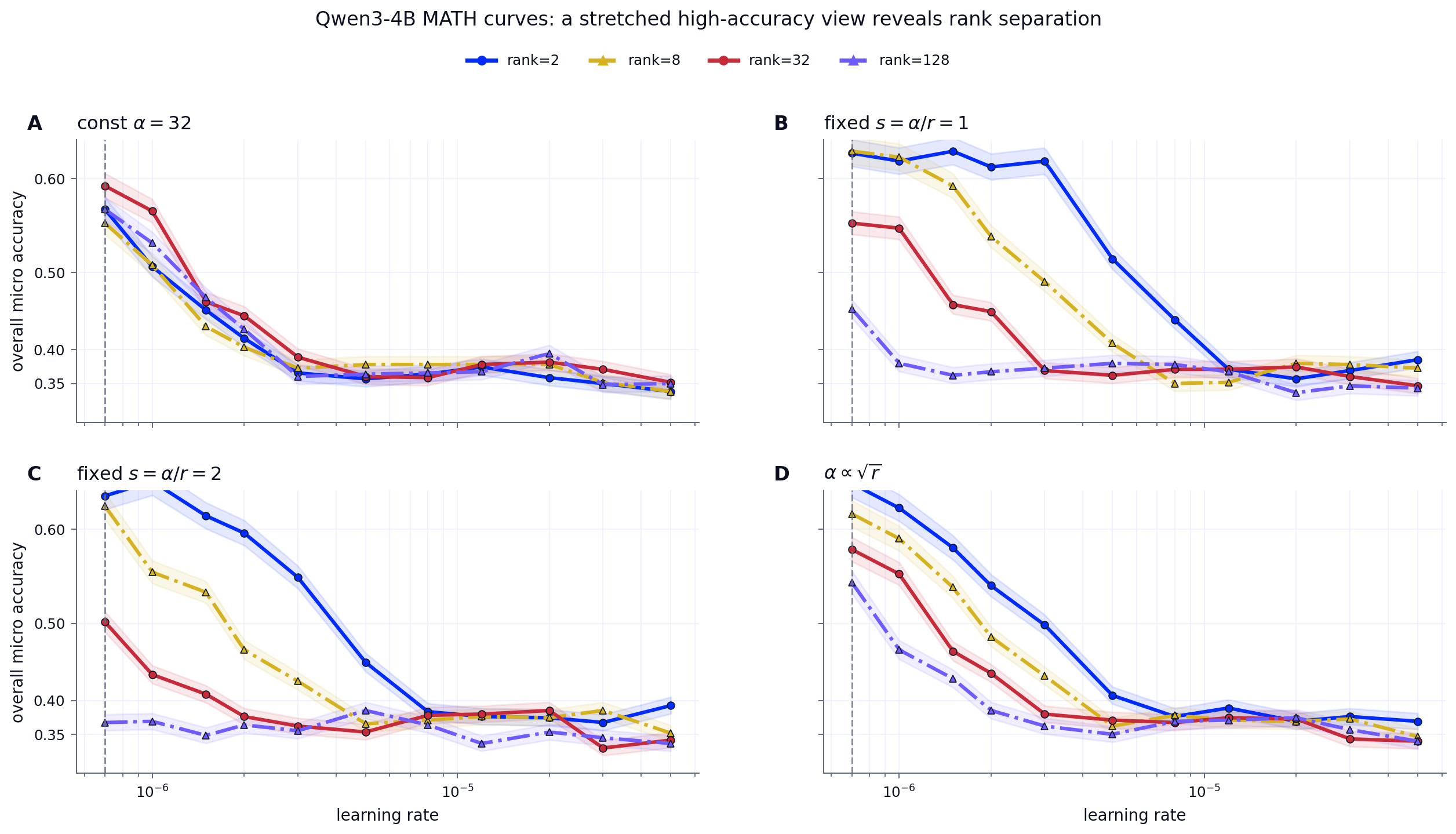}
\caption{Curves motivate square-root alpha scaling as a practical path to same-order learning-rate reuse.}\label{fig:triquetra-qwen-curves}
\end{subfigure}
\caption{Qwen3-4B MATH learning-rate transfer across ranks. The reusable band is narrow, and the square-root alpha rule preserves same-order reuse while remaining stronger at high rank.}\label{fig:triquetra-qwen}
\end{figure}

Hyperparameter transfer is a hidden cost of PEFT because many recipes are under-specified. One implementation may keep alpha fixed, another may keep alpha divided by rank fixed, and a third may scale alpha with the square root of rank. Two runs can therefore share the same apparent learning rate and rank while having very different early update magnitudes. Without recording the scaling convention, the recipe is not portable across papers, codebases, model families, or serving platforms.

For personal models, this under-specification becomes a platform problem. A service that trains thousands or millions of adapters cannot ask each user or product team to rediscover the correct learning-rate band. It needs configuration laws that predict how a safe band moves when rank changes for cost reasons, when model size changes for prior reasons, or when task difficulty changes. The Triquetra framing supplies the beginning of such a law, complementing broader hyperparameter-transfer work \citep{yang2022tensorprograms}: track the joint effect of rank, alpha, and learning rate on the first effective weight movement, then validate the resulting transfer rule empirically.

The square-root rule is especially interesting because it makes theoretical and operational goals coincide. The early-step proxy becomes rank-invariant, so the same-order learning-rate band can be reused. The empirical results suggest that this is not always the flattest rule on simple tasks, but it is robust in the harder reasoning setting where transfer quality matters. This is precisely the regime personal-model infrastructure must handle: not one easy benchmark, but many heterogeneous learning problems under limited tuning budgets.

Triquetra also changes how adapter configuration should be documented. A rank number without alpha and learning rate is incomplete, an alpha rule without a transfer policy is incomplete, and a learning-rate recommendation without rank context is not portable. A population of adapters needs configuration laws, not just configuration values. This is why Scale Down includes training-effort efficiency, not only parameter efficiency.

% \begin{figure}[tbp]
% \centering
% \includegraphics[width=0.9\textwidth]{figures/scale_down/triquetra-for-lora-s-entangled-knobs-when-lora-rank-changes-how-should-learning-rate-move_figure_03.png}
% \caption{Early-step theory surfaces explain why $\eta\alpha_r^2/r$ is a useful proxy for cross-rank transfer.}\label{fig:triquetra-theory}
% \end{figure}

\subsection{Beyond LoRA: Toward a Spectrum of Adapters}\label{subsec:beyond-lora}

The preceding sections study Scale Down within the standard LoRA setting: the adaptive state is a static low-rank parameter update, and the main question is how small this update can become while remaining expressive, stable, and reusable. This view is essential, but it does not exhaust the design space of efficient adaptation. Complementary lines of work enlarge the static-LoRA setting in other directions, including hierarchical or layer-wise rank allocation \citep{zhou2025rankadaptor,zhou2026lara}, intra- and inter-layer parameter sharing \citep{zhou2025bslora}, joint optimization of LoRA rank with mixed-precision quantization \citep{zhou2025balancing,zhou2026autoqra}, and combined low-rank plus sparsity compression \citep{zhou2025largecompression}. The remainder of this section pursues a different question: if PEFT is to support persistent personal models, the adaptive unit must not only be small after training; it must also be writable, reusable, and capable of changing with interaction history. Scale Down therefore extends from parameter reduction to state design: how compact, stable, and dynamic can the local adaptive state become?

\paragraph{From static adapters to writable local state.}
Ordinary LoRA stores adaptation directly in low-rank weights. Once trained, the same parameter update is applied regardless of the model's previous interactions. This is suitable for task or domain specialization, but it is less suitable for personal models that must accumulate preferences, corrections, task progress, and long-term behavioral patterns over time. A broader adapter view treats PEFT as an interface between a frozen shared backbone and a local adaptive state. The important design question is then not only how many trainable parameters the adapter introduces, but also what kind of state it maintains, how that state is updated, and how it influences inference.

\begin{figure}[t]
    \centering
    \includegraphics[width=0.95\textwidth]{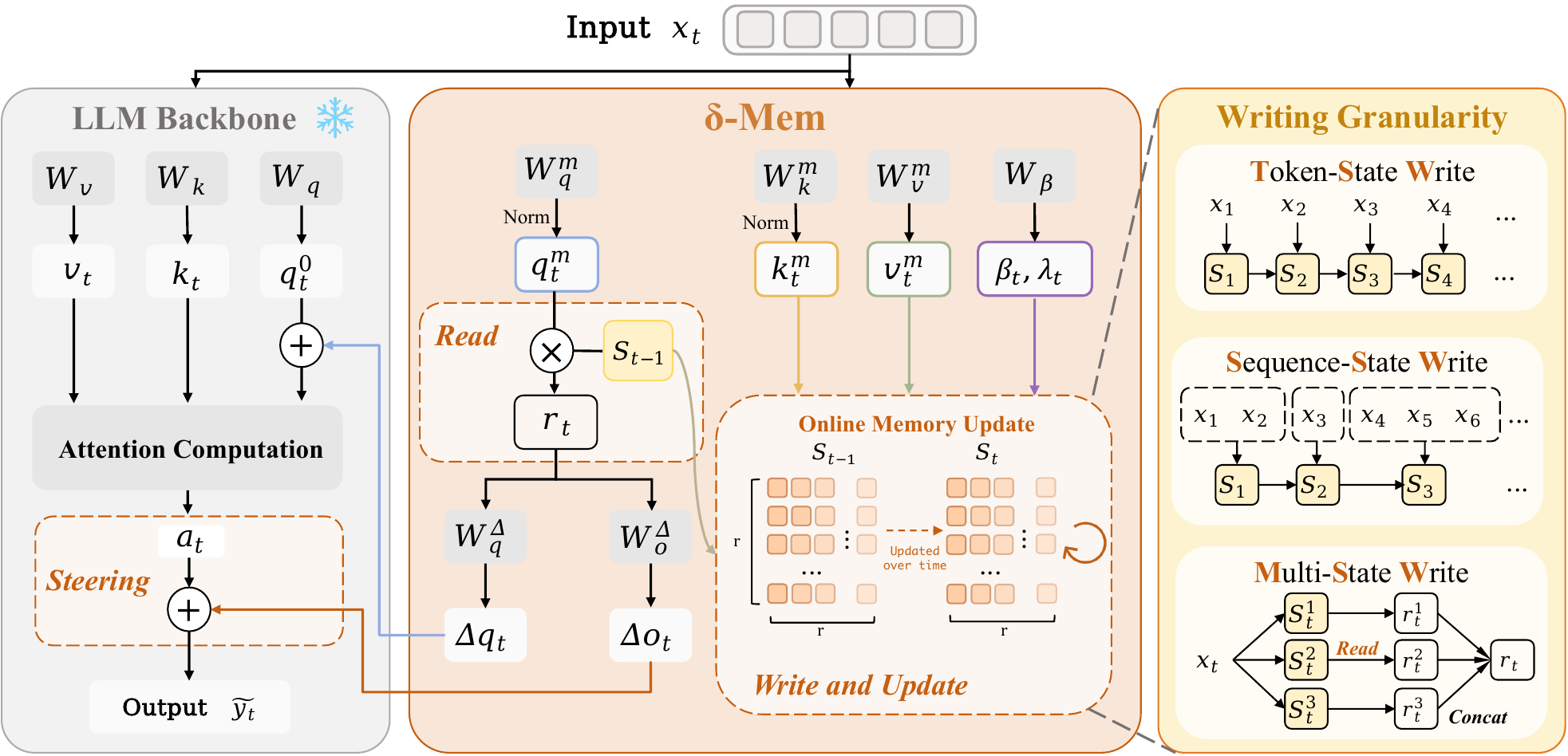}
    \caption{Overview of $\delta$-mem. A compact online memory state is read to produce history-dependent low-rank corrections to the frozen attention computation, and is then updated with current key--value information through delta-rule learning.}
    \label{fig:delta-mem-arch}
\end{figure}

\paragraph{A representative stateful adapter: $\delta$-mem.}
A concrete example of this direction is $\delta$-mem~\citep{lei2026deltamem}, which augments a frozen full-attention Transformer with a compact online associative-memory state. Unlike ordinary LoRA, whose low-rank update is fixed after training, $\delta$-mem maintains a state that evolves as tokens are processed. At each position, the model reads from the previous memory state, uses the readout to generate low-rank corrections to the backbone attention computation, and then writes the current information back into memory through a delta-rule update, as illustrated in \Cref{fig:delta-mem-arch}.

Concretely, $\delta$-mem maintains a low-dimensional state
\[
    \mathbf S_t \in \mathbb{R}^{r\times r},
\]
which stores key--value associations in the adapter space. Given a memory key $\mathbf k_t^m$ and value $\mathbf v_t^m$, the update can be written as
\[
    \mathbf S_t
    =
    \mathrm{Diag}(\boldsymbol\lambda_t)\mathbf S_{t-1}
    +
    \mathrm{Diag}(\boldsymbol\beta_t)
    \left(
    \mathbf v_t^m-\mathbf S_{t-1}\mathbf k_t^m
    \right)
    (\mathbf k_t^m)^\top .
\]
Here $\boldsymbol\lambda_t$ controls retention and $\boldsymbol\beta_t$ controls write strength. The important point is that the state writes the prediction residual rather than simply accumulating all new information. Already predictable associations induce small updates, while novel or mispredicted associations modify the state. In this sense, $\delta$-mem preserves the low-rank steering interface of LoRA, but replaces a static parameter patch with a history-conditioned correction.

\paragraph{Writing granularity as part of the adaptive state.}
The $\delta$-mem design also shows that the adaptive unit is not defined only by parameter count. It studies different writing granularities: token-level writing preserves fine-grained local information, sequence-level writing reduces redundant token noise, and multi-state writing separates information across several parallel states to reduce interference. For Scale Down, the lesson is that two adapters with similar trainable budgets can behave very differently depending on when they write, what they write, and how their state is organized.

\paragraph{Evidence for compact dynamic memory.}
\begin{table}[tbp]
    \centering
    \footnotesize
    \setlength{\tabcolsep}{4pt}
    \renewcommand{\arraystretch}{1.15}
    \caption{Main benchmark results comparing different memory mechanisms on Qwen3-4B-Instruct. All values report the task-specific metrics. For the final average score, HotpotQA is counted using Exact Match~(EM).}
    \label{tab:delta-mem-baseline}
    \fittowidth{%
    \begin{tabular}{@{}cccccccccccccccc@{}}
    \toprule
    \multirow{2}{*}{\appkey{Model}} & \multirow{2}{*}{IFEval} & \multicolumn{2}{c}{HotpotQA} & \multirow{2}{*}{GPQA-D} & \multicolumn{5}{c}{Memory Agent Bench} & \multicolumn{5}{c}{LoCoMo} & \multirow{2}{*}{Avg.} \\
    \cmidrule(lr){3-4} \cmidrule(lr){6-10} \cmidrule(lr){11-15}
    & & EM & F1 & & Avg. & AR & TTL & LRU & SF & Avg. & Multi & Temp & Open & Single \\
    \midrule
    \appkey{Qwen3-4B-Instruct} & 81.89 & 42.35 & 56.00 & 39.39 & 29.54 & 35.30 & 26.14 & \textbf{47.08} & 14.37 & 40.79 & 38.39 & 32.89 & 10.77 & 48.05 & 46.79 \\
    \appgroup{16}{Textual Memory}
    \quad + BM25 RAG     & -- & 40.35 & 52.83 & -- & 24.49 & 32.20 & 9.74 & 37.86 & 15.00 & 36.68 & 38.12 & 20.34 & 9.99  & 45.47 & 44.56 \\
    \quad + LLMLingua-2  & -- & 36.93 & 50.03 & -- & 15.63 & 21.45 & 1.43 & 38.45 & 8.62  & 40.98 & 39.07 & 30.13 & 10.98 & 49.19 & 42.96 \\
    \quad + MemoryBank   & -- & --    & --    & -- & 17.65 & 22.65 & 7.67 & 36.36 & 9.88  & 38.14 & 37.88 & 21.76 & 13.35 & 47.31 & 43.88 \\
    \appgroup{16}{Parametric Memory}
    \quad + Context2LoRA & 76.71 & 37.85 & 50.88 & 29.29 & 32.53 & 40.00 & 29.86 & 25.15 & \textbf{17.75} & 48.11 & 37.95 & 34.99 & 16.75 & \textbf{60.11} & 44.90 \\
    \quad + MemGen       & 39.37 & 5.36  & 16.27 & 38.89 & 29.61 & 34.85 & 28.45 & 44.30 & 14.38 & 40.05 & 32.93 & 33.30 & 12.67 & 48.13 & 30.66 \\
    \appgroup{16}{Outside-channel Memory}
    \quad + MLP Memory   & 24.95 & 10.94 & 25.83 & 22.73 & 28.80 & 35.35 & 26.00 & 31.19 & 14.38 & 26.85 & 32.87 & 16.72 & 8.81  & 30.75 & 22.85 \\
    \midrule
    \appgroup{16}{$\delta$-Mem}
    \quad + $\delta$-Mem~(SSW) & 81.70 & 49.22 & 63.43 & \textbf{41.41} & 37.84 & 41.50 & \textbf{50.50} & 43.02 & 16.50 & 47.05 & 41.00 & 36.48 & 14.08 & 56.88 & 51.44 \\
    \quad + $\delta$-Mem~(TSW) & \textbf{82.99} & \textbf{49.41} & \textbf{63.66} & 40.40 & 36.48 & 42.45 & 40.64 & 46.08 & 15.88 & 46.53 & 42.14 & 37.20 & 13.35 & 55.36 & \textbf{51.66} \\
    \quad + $\delta$-Mem~(MSW) & 81.52 & 46.86 & 60.47 & 37.37 & \textbf{38.85} & \textbf{44.40} & 47.29 & 41.55 & 17.00 & \textbf{49.12} & \textbf{42.57} & \textbf{39.31} & \textbf{18.12} & 58.59 & 50.74 \\
    \bottomrule
    \end{tabular}%
    }
\end{table}

The empirical results in \Cref{tab:delta-mem-baseline} support this adaptive-state view. On a Qwen3-4B-Instruct backbone, the best $\delta$-mem variant improves the average score from $46.79\%$ to $51.66\%$, while outperforming static or textual memory baselines such as Context2LoRA~\citep{back2026understanding,hu2021lora}. The gains are especially visible on memory-intensive benchmarks. On MemoryAgentBench~\citep{hu2025evaluating}, $\delta$-mem improves the average score from $29.54\%$ to $38.85\%$. On LoCoMo~\citep{maharana2024longmemory}, the multi-state variant reaches the strongest average score, suggesting that multiple compact states can help reduce memory interference in long-context personal-memory settings. On HotpotQA~\citep{yang2018hotpotqa}, token-level writing improves EM/F1 from $42.35\%/56.00\%$ to $49.41\%/63.66\%$. These results suggest that the online state provides useful historical signals that are not captured by static parameter patches or purely textual memory mechanisms.

\paragraph{Efficiency of the stateful interface.}
The efficiency profile is also consistent with the Scale Down argument. The token-state and sequence-state variants introduce only $4.87$M trainable parameters, about $0.12\%$ of the Qwen3-4B backbone, while the multi-state variant uses $19.47$M parameters, about $0.48\%$. These overheads are substantially smaller than heavier auxiliary-memory baselines such as MemGen and MLP Memory. Because the recurrent state has fixed size, its storage cost does not grow linearly with interaction history length. Inference is not free: each decoding step must read from and update the online state. However, this places $\delta$-mem at a useful operating point: a small recurrent computation cost is exchanged for persistent, history-dependent steering.

\paragraph{Implications for PEFT scaling.}
The broader implication is that PEFT scaling should not be equated with static rank reduction alone. In ordinary LoRA, Scale Down asks how low the rank can be while preserving task performance. In stateful adapters such as $\delta$-mem, Scale Down also asks how small a writable local state can be while still carrying useful historical information. This distinction matters for personal models. The frozen backbone supplies the shared foundation prior, the adapter parameters define the read--write interface, and the online state stores instance-specific experience. Future PEFT mechanisms may therefore combine three ingredients: a strong shared prior, a compact low-rank steering interface, and a small dynamically writable state. Their role is not merely to reduce the cost of fine-tuning, but to make local adaptive state persistent, compact, and directly coupled to the model's forward computation.

\section{\texorpdfstring{\pony{Scale Out: From Individual Adaptation to Population-Scale Personalization}}{Scale Out: From Individual Adaptation to Population-Scale Personalization}}\label{sec:scale-out}

\begin{figure}[tbp]
    \centering
    \includegraphics[width=\linewidth]{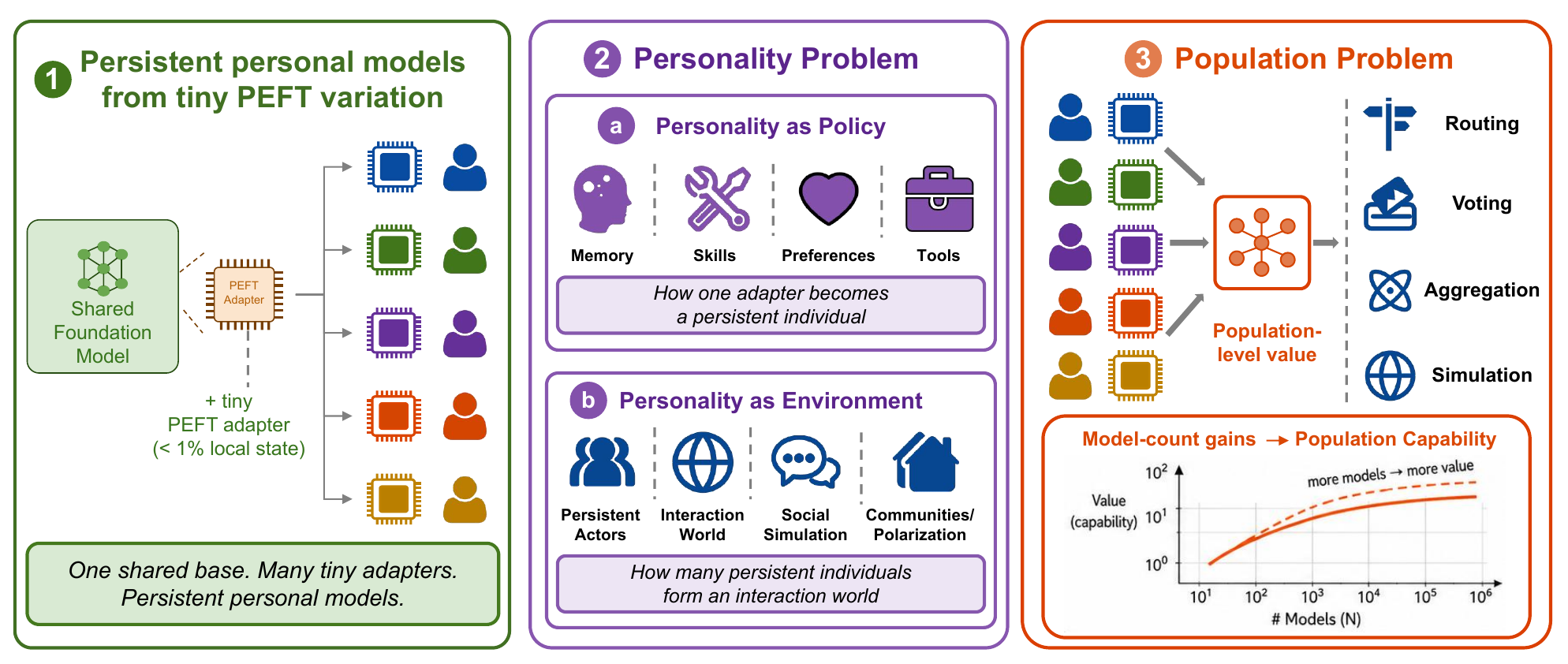}
    \caption{\pony{Scale Out maps small stable variation to population-level value. A shared base model supports persistent personal models through lightweight PEFT adapters. Scale Out requires both preserved individuality and population-level utility.}}
    \label{fig:scale-out-teaser}
\end{figure}

\pony{Scale Down identified the operating regime in which the local adaptive state becomes small, stable, and cheap enough to be trained, stored, updated, and served repeatedly. Scale Out begins when such adaptation becomes routine. The question is no longer only how to improve one adapted model, but what becomes possible when many persistent personal model instances can exist at the same time. PEFT makes this question concrete: a shared foundation model can support many isolated adapters, each carrying a different local state.}

\pony{Adapter count alone, however, is not a scaling law. If many adapters collapse toward the same policy, model count is only redundancy. If adapters remain different but their differences cannot be simulated, composed, selected, or aggregated, diversity remains local personalization. Scale Out therefore asks three connected questions, each addressed by a subsection below: (1) how personal models preserve continuity for individuals, including what to memorize, how to write memory into adapter parameters, and how to govern learned capabilities (Section~\ref{subsec:personality-adapters}); (2) how persistent user-conditioned policies can support user simulators and agent environments (Section~\ref{subsubsec:personality-as-environment}); and (3) how diversity among adapted models can become a source of collective performance (Section~\ref{subsec:population-capability}).}

\pony{A note on Context Learning: it appears in Section~\ref{subsec:personality-adapters} as the \emph{write policy} for LoRA-as-memory, the mechanism that decides which context-time improvements should be stabilized into adapter parameters. It is not a standalone Scale Out algorithm. Its role is to make memory-signal efficiency tractable: without a principled write policy, LoRA memory capacity is a hard ceiling, but with one, repeated interaction can selectively fill that capacity with behaviorally useful state.}

\subsection{\pony{Personal Models for Individuals}}\label{subsec:personality-adapters}

\pony{A personal model is not just a universal assistant with a longer prompt. It needs persistent adaptive state, so that repeated experience can shape future behavior. A task adapter can be disposable: it solves a benchmark or domain and can be replaced when the task changes. A personal adapter instead carries part of the enduring state for one user, agent, role, or workflow: memories, preferences, skills, and tool habits. The adapter stores part of the learned behavioral state, not the entire user history.}

\pony{Personal adaptation has two roles in Scale Out. As a user-conditioned policy, it explains how a shared base model becomes a persistent personal model for one user. As a simulated environment component (\Cref{subsubsec:personality-as-environment}), it explains how many such persistent policies can form user simulators or agent environments. The final population subsection then asks a separate question: whether differences among adapted models can be aggregated into measurable system-level performance.}

\paragraph{From adapter to personal policy.}
\pony{The first function of local adaptive state is policy specialization. A personal model should not merely answer in a preferred tone. It should help decide what to remember, which tools to prefer, how to ask questions, when to avoid action, how to recover from failures, and which behaviors are natural or unacceptable for a particular user. In this sense, personalization is not decoration. It is the policy layer that determines how a shared base model becomes useful for one person. Adapter isolation turns personalization from a global update problem into a local policy problem: the base model remains shared, while experience and behavior become user-specific. The rest of this subsection develops this view in four steps: memory is a precondition for continuity, memory requires a capacity law, bounded capacity requires a write policy, and once adapters shape tools and skills, personalization becomes an agentic capability.}

\paragraph{Memory as the precondition of personality.} 
Personality requires memory because a model cannot remain itself without preserving experience across interactions. Preferences, habits, past failures, recurring tasks, relationships, and learned workflows all depend on accumulated state. Prompt-based personalization is transient; retrieval-based memory is external and must be reinterpreted at every turn; parametric memory offers the stronger possibility that selected experience becomes part of the model's own policy. This motivates the LoRA-as-memory question: can a lightweight adapter serve as a bounded memory substrate for personal state? Since conversational and agent memory require more than recognition accuracy \citep{packer2024memgpt,chhikara2025mem0,maharana2024longmemory}, a personal adapter must support recall, addressing, reasoning, overwrite, conflict resolution, and behavioral application.

\paragraph{LoRA as memory: capacity scaling law.}
To support millions of personal models, we need scaling laws linking adapter size to capacity, stability, and retrieval accuracy. Key questions include how capacity correlates with LoRA rank, target modules, training tokens, and base-model scale, and when storage becomes detrimental interference.

% 这一段其实是介绍我们提出了报菜名的benchmark，如果太长可以放到附录
% Central to establishing these laws is a robust evaluation standard. A useful LoRA memory benchmark should cover four dimensions. (1) \textbf{Content coverage} asks whether facts, events, preferences, constraints, and procedures have been included. (2) \textbf{Structural coverage} asks whether temporal order, hierarchy, causality, and cross-segment dependencies are represented. (3)\textbf{ Difficulty coverage} asks whether tasks include direct recall, multi-hop reasoning, comparison, overwrite, and conflict resolution. (4) \textbf{Distributional coverage} asks whether generated queries resemble the downstream questions that the personal model will actually face. A benchmark that only converts context into convenient QA can overstate memory while missing the behaviors needed for personal models. We therefore introduce DishNameBenchmark as a controlled benchmark for isolating the core operations of LoRA memory. Although the surface task is simple, the abstraction is intentionally general: a dish name is a memory slot, a list of dishes is a local write, multiple turns create a sequential memory stream, a position query tests addressing, an adjacency query tests local relation, and a correction tests overwrite. This design lets us vary memory length, update frequency, query type, and correction pattern while keeping the underlying memory object interpretable. 

Establishing such laws requires a benchmark that measures more than convenient context-to-QA conversion. A useful LoRA memory benchmark should test content, structure, difficulty, and downstream query distribution, including recall, addressing, relation, overwrite, and conflict resolution. We therefore introduce DishNameBenchmark as a controlled benchmark for isolating core LoRA-memory operations. It abstracts memory into interpretable slot-writing and slot-querying tasks, allowing us to vary memory length, update frequency, query type, and correction pattern while keeping the stored object simple and measurable.

\begin{figure}[tbp]
    \centering
    \includegraphics[width=\linewidth]{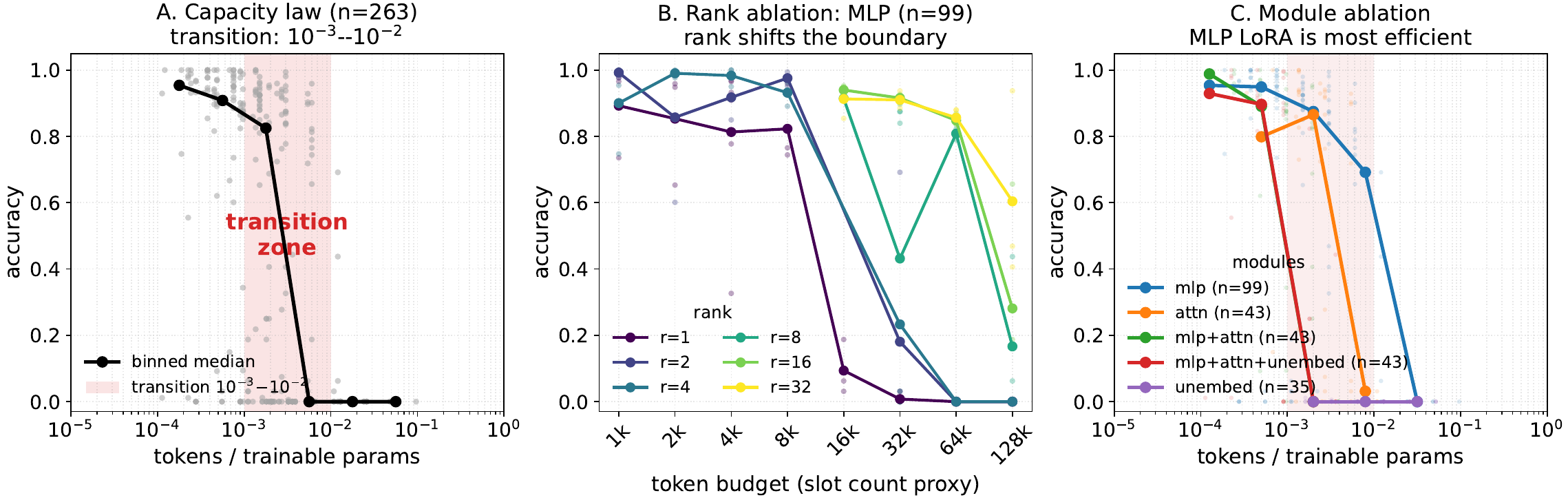}
    \caption{DishNameBenchmark reveals a bounded LoRA memory law: usable capacity lies around $10^{-3}-10^{-2}$ tokens per trainable parameter, degrades predictably after saturation, and is most parameter-efficient when memory is written into MLP LoRA.}
    \label{fig:lora-as-memory-scaling-law}
\end{figure}

We evaluate LoRA memory capacity on DishNameBenchmark using Qwen3-series models. We report capacity efficiency as the ratio between memory tokens and trainable parameters, and measure whether the model can correctly recover the stored slot values under position, adjacency, and correction-style queries. The results are shown in \Cref{fig:lora-as-memory-scaling-law}: 

\begin{itemize}
    \item The first result is \textbf{a sharp capacity transition}. Across 263 runs, accuracy stays close to one when capacity efficiency is below roughly \(10^{-3}\), begins to degrade in the transition region between \(10^{-3}\) and \(10^{-2}\), and collapses toward zero once the memory load exceeds about \(10^{-2}\) tokens per trainable parameter. This suggests that LoRA memory has a measurable capacity ceiling: \textbf{the empirical upper limit in this setting lies between \(10^{-3}\) and \(10^{-2}\) memory tokens per trainable parameter.}
    \item For a fixed parameter budget, \textbf{performance drops approximately linearly after the adapter reaches its capacity limit.} In the rank ablation, increasing the token budget initially preserves near-perfect accuracy, but once the memory load crosses the adapter's effective capacity, accuracy falls rapidly as the number of required slots increases. Rank mainly shifts the capacity boundary by increasing the number of trainable parameters. At low memory budgets, even small-rank adapters can match larger-rank adapters. At high memory budgets, larger ranks survive longer because they provide more total storage surface. This supports the interpretation that rank buys capacity, but does not remove the underlying capacity law.
    \item The third result concerns where memory should be written. In the module ablation, training MLP LoRA provides the best parameter efficiency. At matched parameter budgets, MLP adapters maintain high accuracy at larger capacity-efficiency values than attention-only adapters, combined MLP+attention adapters, or unembedding adapters. Attention-only and full-module training can store memory, but they are less efficient per trainable parameter. Unembedding-only training performs worst and collapses quickly. The observed ordering is approximately
    \[
    \mathrm{MLP} > \mathrm{Attention} \approx \mathrm{All} \gg \mathrm{Unembed}.
    \]
    From a scale-out perspective, this matters because personal adapters must be cheap to train, store, and serve. If the goal is to maximize memory capacity per parameter, \textbf{the most efficient design in this benchmark is to train MLP LoRA only}.
\end{itemize}

% 从scaling law说明，容量是由上限的，不能什么都记住，所以引入到下一章节，哪些需要被记住
These results make LoRA memory a scarce resource rather than an unlimited store. Diversity therefore cannot be produced by writing everything into every adapter. If usable capacity is bounded and module choice strongly affects parameter efficiency, the next question is not only how much a personal adapter can memorize, but what kind of information deserves to become stable local variation.

\paragraph{What to memorize: behavioral state, not raw facts.}
The next question is what deserves LoRA memory. Because LoRA memory is harder to inspect and more expensive to edit, it should not function as a general fact store. A personal model instead needs a memory hierarchy (\Cref{tab:memory-hierarchy}): editable facts should remain in retrieval, inspectable external reality should remain in tools, and LoRA should be reserved for persistent behavioral adaptation.

\begin{table}[tbp]
\centering
\small
\setlength{\tabcolsep}{6pt}
\renewcommand{\arraystretch}{1.20}
\caption{A practical personal model should decide which memory layer should store each kind of state.}
\label{tab:memory-hierarchy}
\fittowidth{%
\begin{tabular}{@{}M{2.8cm}M{4.2cm}M{6.2cm}@{}}
\toprule
\apphead Memory layer & Example & Best suited for \\
\midrule
\appkey{Context}          & Current conversation              & Short-term reasoning and local task state \\
\addlinespace[2pt]
\appkey{Retrieval memory} & Notes, documents, user facts      & Editable factual recall and large evidence stores \\
\addlinespace[2pt]
\appkey{Tool state}       & Calendars, files, databases       & External reality that should remain inspectable \\
\addlinespace[2pt]
\appkey{LoRA memory}      & Skills, habits, policy, persona   & Persistent behavioral adaptation \\
\bottomrule
\end{tabular}%
}
\end{table}

This hierarchy clarifies the role of LoRA memory. A rare document should stay in retrieval, a calendar event should remain tool state, and a recurring workflow should become skill memory. The memories that matter for Scale Out are not isolated facts, but behavior-shaping structures that make one adapted model reliably different from another. Skills are the natural target because they are reusable, procedural, and policy-shaping. Following the intuition of SKILL-0 \citep{skill02026}, the scale-out implication is that a personal adapter should become a compact library of learned behavioral state: workflows, tool-use habits, reasoning templates, domain heuristics, safety routines, and action policies.

\begin{figure}[t]
\centering
\includegraphics[width=0.80\textwidth]{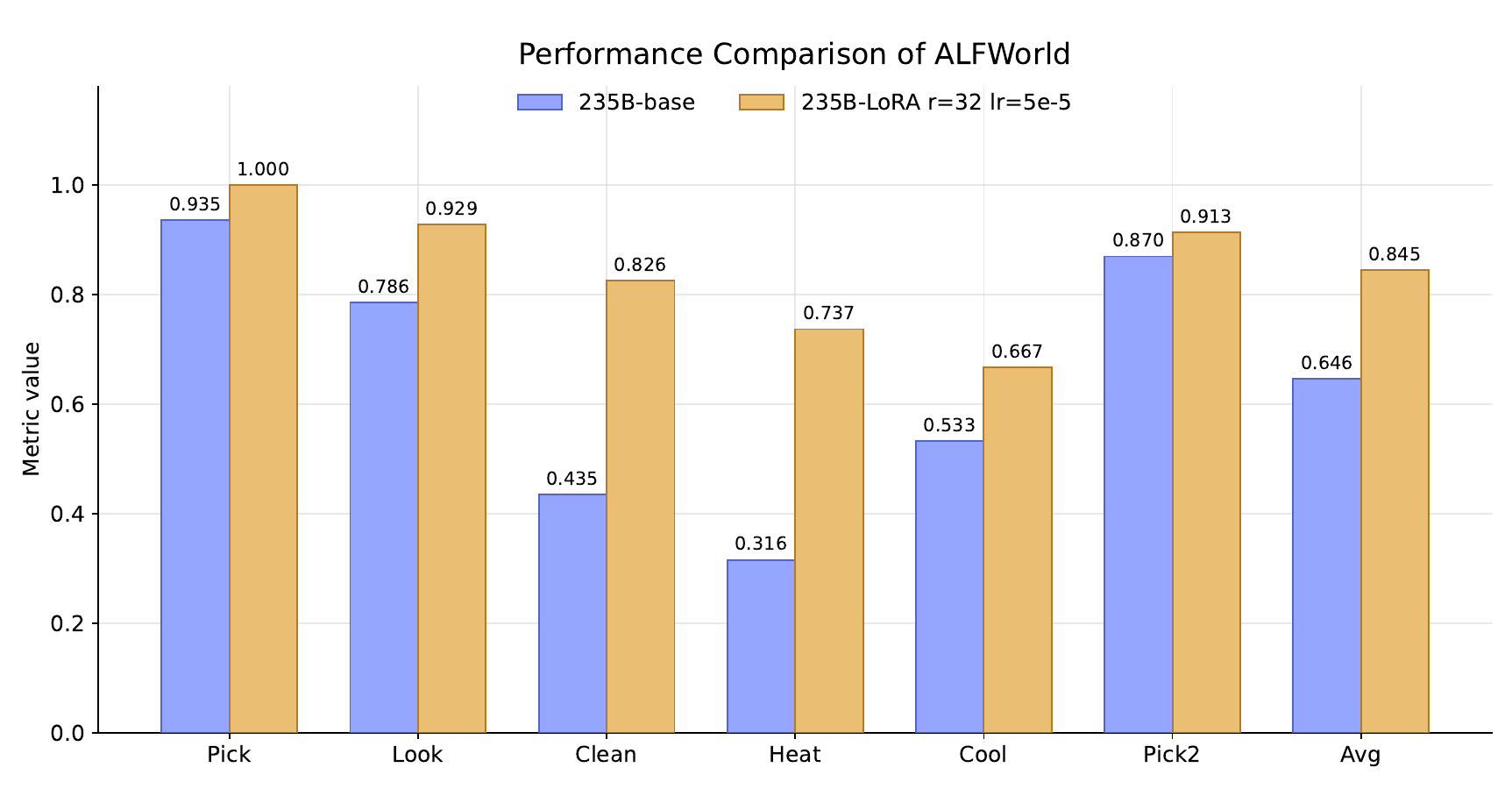}
\caption{LoRA skill memory on ALFWorld. Starting from Qwen3-235B, a rank-32 LoRA adapter trained with the Skill-0/MinT recipe improves ALFWorld validation performance over the base model across task categories, raising the average score from \(0.646\) to \(0.845\).}
\label{fig:skill0-alfworld-lora}
\end{figure}

We test this skill-memory interpretation by applying LoRA to a large-prior agent setting. Using Qwen3-235B as the shared base model, we train a rank-32 LoRA adapter with the Skill-0/MinT ALFWorld recipe and evaluate it on the ALFWorld benchmark. As shown in \Cref{fig:skill0-alfworld-lora}, the adapted model improves over the base model on all reported task categories, with the average metric increasing from \(0.646\) to \(0.845\). This result is not evidence that LoRA should store arbitrary facts. Rather, it supports the narrower claim that LoRA can store reusable skill-like behavioral state: the adapter changes how the model acts in a procedural environment.

% \textcolor{blue}{
% We will reproduce this line of evidence with LoRA: skill demonstrations will be encoded into adapters and compared against prompt-only, retrieval-only, LoRA-skill, and LoRA-plus-retrieval systems. The expected comparison is not whether an adapter can memorize a string, but whether it improves success rate, consistency, tool-use correctness, and generalization when a skill must be applied repeatedly.
% }

\paragraph{How to memorize: Context Learning as write policy.}
% 说完改记住什么东西以后，说一下如何记住这些信息
The memory hierarchy specifies where different states should live; Context Learning specifies how experience moves between layers. If Scale Down makes local variation cheap to store, Context Learning decides which temporary experiences should be stabilized into that variation. Context Engineering selects, retrieves, and arranges information to improve the current response. Context Learning asks which parts of that context-time improvement should become durable model state. In this sense, it is not merely better prompting, but the write policy of the personal model: it decides when repeated usefulness justifies converting context, retrieval, tool outcomes, or demonstrations into adapter parameters.

\begin{lstlisting}[
  language=Python,
  float=tbp,
  caption={Context Distillation as an on-policy context-to-parameters transfer. Context Learning as repeated Context Distillation.},
  label={lst:context-distill},
  basicstyle=\ttfamily\footnotesize,
  keywordstyle=\color{blue!70!black}\bfseries,
  commentstyle=\color{gray!70!black}\itshape,
  stringstyle=\color{green!45!black},
  numbers=left,
  numberstyle=\tiny\color{gray},
  numbersep=6pt,
  frame=single,
  rulecolor=\color{gray!35},
  backgroundcolor=\color{gray!6},
  xleftmargin=2.4em,
  xrightmargin=0.5em,
  framexleftmargin=2.0em,
  breaklines=true,
  breakatwhitespace=true,
  tabsize=4,
  showstringspaces=false,
  columns=fullflexible,
  keepspaces=true,
]
def context_distill(model, query, build_context, rl_update):
    # Step 1: query-only produces an on-policy rollout.
    out = model.sample(query)

    # Step 2: query + context scores the rollout.
    ctx = build_context(query)
    r_tok = model.token_reward(query, ctx, out)

    # Step 3: update from the query-only rollout and rewards.
    return rl_update(model, query, out, r_tok)

def context_learning(model, queries, build_context, rl_update, steps):
    for _ in range(steps):
        query = next(queries)
        model = context_distill(model, query, build_context, rl_update)
    return model
\end{lstlisting}

\begin{figure}[tbp]
\centering
\resizebox{0.96\textwidth}{!}{\begin{tikzpicture}[x=1cm,y=1cm,>=Latex,line cap=round,line join=round,font=\footnotesize\sffamily]
  \tikzset{
    panel/.style={draw=mindlabfg, very thick, rounded corners=1.4mm, fill=white},
    softpanel/.style={draw=mindlabfg, very thick, rounded corners=1.4mm, fill=mindlabblue!10},
    stage/.style={draw=mindlabfg, very thick, rounded corners=1.4mm, fill=white},
    box/.style={
      draw=mindlabfg, thick, rounded corners=0.8mm, fill=white, align=center,
      inner sep=2pt, minimum width=1.55cm, minimum height=0.55cm,
      font=\footnotesize\sffamily
    },
    boxblue/.style={box, fill=mindlabbluepale!55},
    boxaccent/.style={box, fill=mindlabblue!22},
    boxsoft/.style={box, fill=mindlabfg!7},
    op/.style={
      draw=mindlabfg, thick, rounded corners=0.8mm, fill=mindlabblue!16, align=center,
      inner sep=2pt, minimum width=1.60cm, minimum height=1.30cm,
      text width=1.45cm, font=\footnotesize\sffamily
    },
    bigarrow/.style={->, line width=0.9pt, draw=mindlabfg, shorten >=2pt, shorten <=2pt},
    softarrow/.style={->, semithick, draw=mindlabfg!55, dashed, shorten >=2pt, shorten <=2pt},
    flowarrow/.style={->, line width=0.9pt, draw=mindlabfg, shorten >=3pt, shorten <=3pt},
    title/.style={font=\bfseries\footnotesize\sffamily, text=mindlabfg, align=center, fill=white, inner sep=2pt},
    stagetitle/.style={font=\bfseries\footnotesize\sffamily, text=mindlabfg, align=center, fill=white, inner sep=2pt},
    figtitle/.style={font=\bfseries\small\sffamily, text=mindlabfg, align=center},
    elabel/.style={font=\scriptsize\sffamily, text=mindlabfg, align=center, fill=white, inner sep=1.2pt},
    plainlabel/.style={font=\scriptsize\sffamily, text=mindlabfg!72, align=center, inner sep=0pt}
  }

  \path[use as bounding box] (-0.20,-3.45) rectangle (16.20,3.50);
  \node[figtitle] at (8.00,3.10) {Context Distillation: writing context gains into parameters};

  % Stage panels share top y = 2.20 and bottom y = -2.10
  % Column anchors (panel centers):
  %   S1 x=2.20  S2 x=6.55  S3 x=11.30  R x=14.95
  \def\Stop{2.20}
  \def\Sbot{-2.10}

  \node[stage] (s1) at (2.20,0.05)  [minimum width=3.30cm, minimum height=4.30cm] {};
  \node[stage] (s2) at (6.55,0.05)  [minimum width=4.20cm, minimum height=4.30cm] {};
  \node[stage] (s3) at (11.30,0.05) [minimum width=4.20cm, minimum height=4.30cm] {};
  \node[stage] (sR) at (14.95,0.05) [minimum width=1.70cm, minimum height=4.30cm] {};

  \node[stagetitle] at (2.20,2.05)  {Step 1\\Query-only rollout};
  \node[stagetitle] at (6.55,2.05)  {Step 2\\Context scoring \& rewards};
  \node[stagetitle] at (11.30,2.05) {Step 3\\RL-style parameter update};
  \node[stagetitle] at (14.95,2.05) {Result};

  % Stage 1 internals: aligned column at x=2.20, rows y=0.85,-0.10,-1.05
  \node[boxaccent] (q1)    at (2.20,0.85)  {Query};
  \node[boxblue]   (th1)   at (2.20,-0.10) {$\theta_t$};
  \node[boxsoft]   (out1)  at (2.20,-1.05) {Output};
  \draw[flowarrow] (q1.south) -- (th1.north);
  \draw[flowarrow] (th1.south) -- (out1.north);

  % Stage 2: inputs column x=5.45, op column x=7.65
  \node[boxaccent] (q2)   at (5.45,0.85)  {Query};
  \node[boxblue]   (ctx2) at (5.45,-0.10) {Context};
  \node[boxsoft]   (out2) at (5.45,-1.05) {Output};
  \node[op]        (rm)   at (7.65,-0.10) {Reward /\\scoring\\function};
  \draw[flowarrow] (q2.east)  -- (q2.east -| rm.west);
  \draw[flowarrow] (ctx2.east) -- (rm.west);
  \draw[flowarrow] (out2.east) -- (out2.east -| rm.west);

  % Stage 3: inputs column x=10.20, op column x=12.40
  \node[boxaccent] (q3)   at (10.20,0.85)  {Query};
  \node[boxblue]   (th3)  at (10.20,-0.10) {$\theta_t$};
  \node[boxsoft]   (rew3) at (10.20,-1.05) {$\{r_i\}$};
  \node[op]        (opt)  at (12.40,-0.10) {Optimizer /\\update\\function};
  \draw[flowarrow] (q3.east)  -- (q3.east -| opt.west);
  \draw[flowarrow] (th3.east) -- (opt.west);
  \draw[flowarrow] (rew3.east) -- (rew3.east -| opt.west);

  % Result
  \node[boxaccent] (thN) at (14.95,-0.10) [minimum width=1.30cm] {$\theta_{t+1}$};
  \node[plainlabel] at (14.95,-1.00) {distilled\\knowledge};

  % Stage-to-stage arrows (horizontal, between panels)
  \draw[bigarrow] (s1.east) -- (s2.west);
  \draw[bigarrow] (s2.east) -- (s3.west);
  \draw[bigarrow] (s3.east) -- (sR.west);

  % Carry-over: only one curve, from reward fn down/over to {r_i} input of stage 3.
  \draw[softarrow] (rm.south) .. controls (7.65,-2.85) and (10.20,-2.85) .. (rew3.south);
  \node[elabel, text=mindlabfg!75] at (8.92,-2.95) {token rewards $\{r_1,\dots,r_n\}$};
\end{tikzpicture}}
\caption{Context Distillation updates the query-only policy using a query-plus-context teacher signal.}\label{fig:context-distillation}
\end{figure}
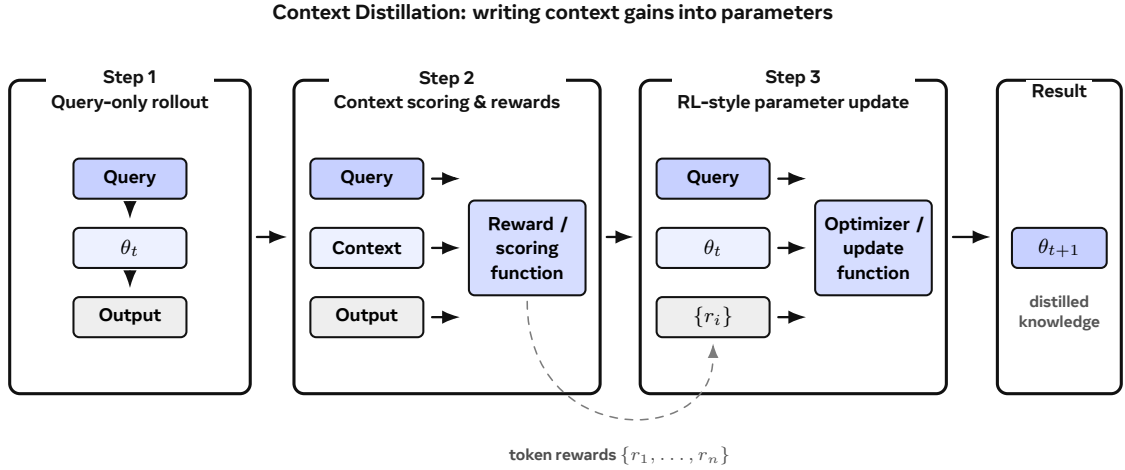

% 介绍一下context distillation是如何实现的
The mechanism begins with Context Distillation, summarized in \Cref{lst:context-distill} and \Cref{fig:context-distillation}. A query-only policy first produces an on-policy rollout. A stronger query-plus-context system then evaluates that rollout using retrieved evidence, demonstrations, tool outputs, execution outcomes, or slower verification. The resulting token-level or trajectory-level signal drives an RL-style update. Crucially, the update is applied to the query-only rollout, so the model learns to perform better without requiring the same context to be present at inference time. This differs from off-policy context distillation, where targets are produced with context and then learned without context through supervised fine-tuning \citep{snell2022distillingcontext}. Here, context is not used as a supervised target source; it is used as privileged information for scoring an output already produced by the query-only policy.

% 介绍一下context learning是如何实现的
Repeating this operation turns Context Distillation into Context Learning. At step \(t\), policy \(\pi_t\) answers a query without privileged context; the context system retrieves evidence, runs tools, observes outcomes, or performs slower verification; and the adapter is updated from this signal to produce \(\pi_{t+1}\). Future query-only behavior now starts from a stronger internal state. Related self-distillation methods also use additional information to create stable parameter updates \citep{hubotter2026selfdistillation,shenfeld2026continualselfdistillation}. Here, however, the loop is explicitly personalized: each iteration decides what part of a user's context, tools, outcomes, or demonstrations should become adaptive state.

% \begin{figure}[tbp]
% \centering
% \includegraphics[width=0.8\textwidth]{figures/scale_out/figure_06.png}
% \caption{Context Distillation updates the query-only policy using a query-plus-context teacher signal.}\label{fig:context-distillation}
% \end{figure}

% 介绍RAG2LoRA为一个例子
RAG2LoRA illustrates this memory-layer transfer. The point is not that all retrieved facts should become parameters, but that retrieval can provide a teacher signal when it repeatedly improves behavior. Over time, recurring facts, preferences, and procedures can be partially internalized into the adapter, reducing dependence on perfect retrieval while leaving editable evidence in external memory. This is the memory-signal efficiency required for Scale Out: each user can maintain a personal adapter whose value grows with interaction while the base model remains shared.

\paragraph{Agentic personal models.}
The write-policy problem becomes most concrete in agentic systems, where the memory being written is often a workflow, tool habit, or post-failure correction. In this setting, the adapter is no longer only improving text generation; it is shaping how the model acts over files, messages, calendars, documents, code, tools, and long-running workflows. Personality becomes operational when memory determines continuity, skills determine competence, and policy determines safe action. This is where diversity begins to produce activity: different memories and skills lead agents to choose different tools, recover from different failures, and follow different workflows under the same external task. Related agent-learning work also treats skills and reflection as reusable learning objects \citep{shinn2023reflexion,xia2026skillrl}.

MindClaw \citep{li2026mindclaw} illustrates this transition from prompt-space skill growth to parametric skill consolidation. Textual skill libraries are useful during cold start, but over time they create skill drift and retrieval dependence: the library grows, prompts get longer, and the agent may still fail to trigger the right capability. The scale-out goal is therefore not a larger skill library, but a higher probability that useful skills are learned, triggered, and applied consistently.

Thus, personality as policy is not merely memory. It is the construction of a persistent local policy: a personal model that remembers at the right level, internalizes reusable skills, and turns user-specific experience into stable action tendencies. This completes the individual side of Scale Out. The next question is what happens when many such policy-bearing adapters coexist and interact.

\subsection{\pony{User Simulators and Agent Environments}}\label{subsubsec:personality-as-environment}

\paragraph{\pony{From personal policy to user simulation.}}
\pony{A personal model is not only an interface for one user. In an agent environment, each persistent user-conditioned policy can become part of the environment experienced by other agents. Once adapters carry distinct memory and policy states, the relevant object is no longer only the distribution of requests served by one model, but the distribution of histories, preferences, reactions, and interactions among persistent model instances. For an agent, the user is part of the environment, and a good user simulator therefore acts as a world model for user reactions, preferences, constraints, and long-term behavior.}

\paragraph{\pony{The collapse problem of prompt-based user simulation.}}
\pony{LLM social simulators often instantiate many agents by combining one shared model with different persona prompts. Systems such as OASIS make it possible to run large-scale social-media simulations with LLM agents and recommender dynamics \citep{oasis2024}. Generative Agents \citep{park2023generative} demonstrated that LLM-based agents can produce plausible human-like behavior through prompt-driven memory and reflection. However, persona prompting has a structural limitation: it changes the description of the agent, but not the learned policy that generates behavior. Agents may sound different at the surface level, but repeated interaction can underrepresent durable heterogeneity and drift toward the base model's average stance, style, and action prior. This is precisely the limitation that per-user LoRA adapters are designed to address: rather than describing a persona in a prompt, each adapter carries a distinct learned policy shaped by a different interaction history.}

\begin{table}[tbp]
\centering
\footnotesize
\setlength{\tabcolsep}{4pt}
\renewcommand{\arraystretch}{1.20}
\caption{EvoBot reports that learned social agents better match real group-opinion dynamics than prompt-only LLM baselines \citep{kong-etal-2025-enhancing}. Lower bias and diversity difference indicate closer alignment with real data.}
\label{tab:evobot-group-opinion}
\fittowidth{%
\begin{tabular}{@{}ccccccccc@{}}
\toprule
\apphead \multirow{2}{*}{Method}
& \multicolumn{4}{c}{COVID-19}
& \multicolumn{4}{c}{Russian-Ukrainian Conflict} \\
\cmidrule(lr){2-5} \cmidrule(lr){6-9}
& Mean & Std & $\Delta_{\mathrm{mean}}$ & $\Delta_{\mathrm{std}}$
& Mean & Std & $\Delta_{\mathrm{mean}}$ & $\Delta_{\mathrm{std}}$ \\
\midrule
\appkey{Real}   & $-0.017$ & 0.472 & /     & /     & $-0.239$ & 0.670 & /     & /     \\
\appkey{BC}     & $-0.041$ & 0.389 & 0.089 & 0.112 & $-0.304$ & 0.104 & 0.104 & 0.554 \\
\appkey{Lorenz} & $+0.084$ & 0.725 & 0.107 & 0.264 & $-0.811$ & 0.105 & 0.572 & 0.565 \\
\appkey{Llama}  & $-0.053$ & 0.368 & 0.098 & 0.105 & $-0.324$ & 0.405 & 0.202 & 0.265 \\
\appkey{GPT}    & $+0.032$ & 0.342 & 0.081 & 0.083 & $-0.256$ & 0.435 & 0.135 & 0.238 \\
\appkey{EvoBot} & $+0.010$ & 0.428 & 0.072 & 0.052 & $-0.237$ & 0.480 & 0.101 & 0.194 \\
\bottomrule
\end{tabular}%
}
\end{table}

\Cref{tab:evobot-group-opinion} provides related evidence for this concern: learned social agents better match real group-opinion dynamics than prompt-only LLM baselines, especially in opinion diversity \citep{kong-etal-2025-enhancing}. This does not by itself prove collapse in every prompt-based simulator, but it supports the need for learned, user-conditioned policies when simulation depends on stable heterogeneity. This collapse is especially problematic for social phenomena that require durable disagreement and persistent behavioral difference. Echo chambers, minority capitulation, preference cascades, group polarization, community norms, and coordination failures all depend on agents that carry stable histories and react differently to the same exposure. If every agent shares one mutable policy, then the simulation is not a population of persistent individuals. It is one model role-playing many social actors. This motivates the PEFT formulation of social simulation: a shared base model should provide the common prior, while per-user LoRA adapters provide the persistent, isolated policy states that make a population of agents behave like many individuals rather than one model role-playing many roles. In this view, realistic activity requires stable diversity: agents must not only receive different prompts, but carry different histories and action priors.

\begin{table}[tbp]
    \centering
    \small
    \setlength{\tabcolsep}{6pt}
    \renewcommand{\arraystretch}{1.20}
    \caption{Per-user LoRA produces monotonic structural scaling in OASIS as population size increases.}
    \label{tab:oasis-lora-scaling}
    \fittowidth{%
    \begin{tabular}{@{}M{5.4cm}M{1.8cm}M{1.8cm}M{1.8cm}M{2.6cm}@{}}
    \toprule
    \apphead Metric & $N{=}128$ & $N{=}256$ & $N{=}512$ & $128 \rightarrow 512$ \\
    \midrule
    \appgroup{5}{Identity persistence}
    \quad Final polarization distance      & $+0.388$ & $+0.335$ & $+0.319$ & -- \\
    \quad Supportive stance std.\          & $0.340$  & $0.346$  & $0.357$  & -- \\
    \quad Skeptical stance std.\           & $0.416$  & $0.339$  & $0.393$  & -- \\
    \appgroup{5}{Population topology}
    \quad Effective interaction communities & $9.21$   & $11.77$  & $14.85$  & $+61\%$ \\
    \quad Co-engagement modularity         & $0.502$  & $0.561$  & $0.716$  & $+43\%$ \\
    \quad Within-community side-homophily  & $0.670$  & $0.644$  & $0.583$  & $-13\%$ \\
    \appgroup{5}{Attention cascade}
    \quad Cascade Gini on likes            & $0.913$  & $0.866$  & $0.884$  & -- \\
    \quad Top-10\% post like share         & $0.859$  & $0.731$  & $0.781$  & -- \\
    \bottomrule
    \end{tabular}%
    }
\end{table}

\paragraph{LoRA versus shared-base agents: structural scaling in OASIS.}
We test whether isolated personal adapters change the structure of a simulated social environment by comparing per-user LoRA agents against shared-base agents in OASIS. The population is sampled from the c8 game-development community, with \(N \in \{128,256,512\}\). In the LoRA condition, each user receives a rank-4 LoRA adapter trained on 80 historical tweets. In the control condition, all agents sample decisions from the same shared Qwen3-4B-Instruct base model. The OASIS setup, recommender, decision prompt, follow graph, stance seed posts, and initial polarization distance are held fixed. Prompt-level cross-side exposure remains approximately \(0.16\)-\(0.18\) in every cell, so downstream differences are not explained by systematically different feeds. The recommender controls exposure, while the adapter controls reaction.

The LoRA population differs from the shared-base population along the expected scale-out chain: diversity, activity, and topology (\Cref{tab:oasis-lora-scaling}). First, it preserves identity-level heterogeneity. At every \(N\), final polarization distance remains higher under per-user LoRA, and within-side stance dispersion is consistently larger: supportive-user standard deviation is \(2.18\times\)--\(2.45\times\) that of the base condition, while skeptical-user dispersion is \(1.32\times\)--\(2.01\times\). Second, this diversity produces a richer action ecology. Compared with the shared-base condition, LoRA produces substantially more comments and original posts, while shared-base agents collapse toward a narrower action prior with no original posts and very few comments. Third, activity compounds into population topology. Effective interaction communities grow monotonically from \(9.21\) to \(14.85\), co-engagement modularity grows from \(0.502\) to \(0.716\), and within-community side-homophily falls from \(0.670\) to \(0.583\). These normalized metrics show that larger LoRA populations do not simply produce more events. They produce more effective micro-communities, tighter co-attention structure, and communities that increasingly cross the original supportive/skeptical split.

\begin{table}[tbp]
\centering
\small
\setlength{\tabcolsep}{6pt}
\renewcommand{\arraystretch}{1.20}
\caption{Controlled comparison between per-user LoRA and shared-base agents. Values above one indicate a LoRA advantage for ratios, while negative values indicate lower side-homophily under LoRA.}
\label{tab:oasis-lora-base-comparison}
\fittowidth{%
\begin{tabular}{@{}M{2.4cm}M{5.8cm}M{1.8cm}M{1.8cm}M{1.8cm}@{}}
\toprule
\apphead Comparison & Metric & $N{=}128$ & $N{=}256$ & $N{=}512$ \\
\midrule
\appkey{LoRA / Base} & Effective interaction communities & $1.48\times$ & $2.19\times$ & $1.47\times$ \\
\addlinespace[2pt]
\appkey{LoRA / Base} & Co-engagement modularity          & $1.20\times$ & $0.95\times$ & $1.20\times$ \\
\addlinespace[2pt]
\appkey{LoRA $-$ Base} & Within-community side-homophily & $-0.089$     & $-0.193$     & $-0.197$ \\
\addlinespace[2pt]
\appkey{LoRA / Base} & Supportive stance std.\           & $2.28\times$ & $2.18\times$ & $2.45\times$ \\
\addlinespace[2pt]
\appkey{LoRA / Base} & Skeptical stance std.\            & $2.01\times$ & $1.32\times$ & $1.57\times$ \\
\addlinespace[2pt]
\appkey{Base / LoRA} & Top-10\% post like share          & $1.16\times$ & $1.37\times$ & $1.28\times$ \\
\addlinespace[2pt]
\appkey{LoRA $-$ Base} & Comments in DB                  & $+70$        & $+247$       & $+493$ \\
\addlinespace[2pt]
\appkey{LoRA $-$ Base} & Original posts in DB            & $+139$       & $+153$       & $+306$ \\
\bottomrule
\end{tabular}%
}
\end{table}

The controlled comparison to shared-base agents (\Cref{tab:oasis-lora-base-comparison}) shows that these effects are not an automatic consequence of OASIS or of increasing \(N\). Across population sizes, LoRA produces more effective interaction communities, lower within-community side-homophily, broader stance dispersion, and a heavier long tail of attention. The shared-base condition, by contrast, concentrates likes more sharply and produces much less content-creation signal.

These results should be read as evidence for a scale-out regime rather than a universal social-simulation law: they are measured on one community, one recommender mechanism, and one seed per cell. Still, they isolate the relevant mechanism for this paper. The shared base model primarily simulates exposure to content, while the LoRA population simulates persistent actors with different histories, stances, and behavioral priors. This is the environment-level role of personality in Scale Out.

\paragraph{\pony{Personality as environment.}}
\pony{The OASIS result matters because it changes the ontology of simulation. With personality-bearing adapters, the environment is no longer a feed shown to interchangeable samples from one shared policy. It is composed of persistent actors that react differently to the same exposure and whose differences compound into population topology. The unit of scale is therefore diversity, not only throughput: scaling the number of users means scaling isolated histories, preferences, action priors, and behavioral attractors. Such populations can support product-policy testing, recommender intervention studies, echo-chamber analysis, and multi-agent RL environments populated by more realistic users. The environment result establishes the first value of diversity: persistent adapter differences can generate activity and structure interaction. The next subsection asks whether the same kind of diversity can also be aggregated into direct task performance.}

\subsection{\pony{Diversity as a Source of Collective Intelligence}}\label{subsec:population-capability}
% 开场承接personality到population：Personality explains difference; Population explains value
% 差异可以通过 routing/voting/debate/tool-use aggregation 变成计算资源。
\pony{Personal adaptation explains how adapters remain different. Collective intelligence asks why those differences matter. Once many adaptive instances coexist, diversity itself can become a computational resource. Different personal models may accumulate distinct histories, specialize along different trajectories, make different errors, prefer different tools, and solve problems through different reasoning paths. Scale Out therefore asks not only whether many adapters can be trained and served, but whether increasing the number of distinct adapted models produces measurable system-level value.}

% 实验设计：same 30B base, same RL recipe, Math17k, AIME24. Collaboration vs Repetition.
\paragraph{A Controlled Model-Count Experiment.}
A controlled model-count experiment makes this question measurable. We start from the same base model, Qwen3-30B, keep the same RL recipe and evaluation target, train many LoRA variants that differ only by training-data permutation and masking, and scale collaboration by increasing the number of models \(k\). Answers are aggregated by majority vote over 200 collaboration evaluation sources, using random subset sampling for each \(k\) and 30 random samples per \(k\). Empty extracted answers do not vote. Here, population activity is not social interaction but collective inference: each adapted model contributes an answer, and value emerges only when differentiated answers can be aggregated.

The design controls away other sources of scale: the base model family, Math17k \citep{hendrycks2021math} training task, AIME24 \citep{maa2024aime} evaluation, optimization recipe, answer extraction, and correctness pipeline are fixed. In the figure and analysis, \emph{Collaboration} denotes the former different-model setting, where votes are aggregated across distinct LoRA variants; \emph{Repetition} denotes the former same-model setting, where votes are aggregated from repeated samples of one model. If Collaboration improves more than Repetition, the gain cannot be explained only by stochastic decoding; it indicates complementary policies learned by different LoRA trajectories.

% 主结果：Collaboration 随 k 提升。Repetition saturates earlier。Collaboration vs Repetition gap at large k.
\paragraph{Model count produces predictable gains.}

\begin{figure}[tbp]
\centering
\includegraphics[width=0.52\textwidth]{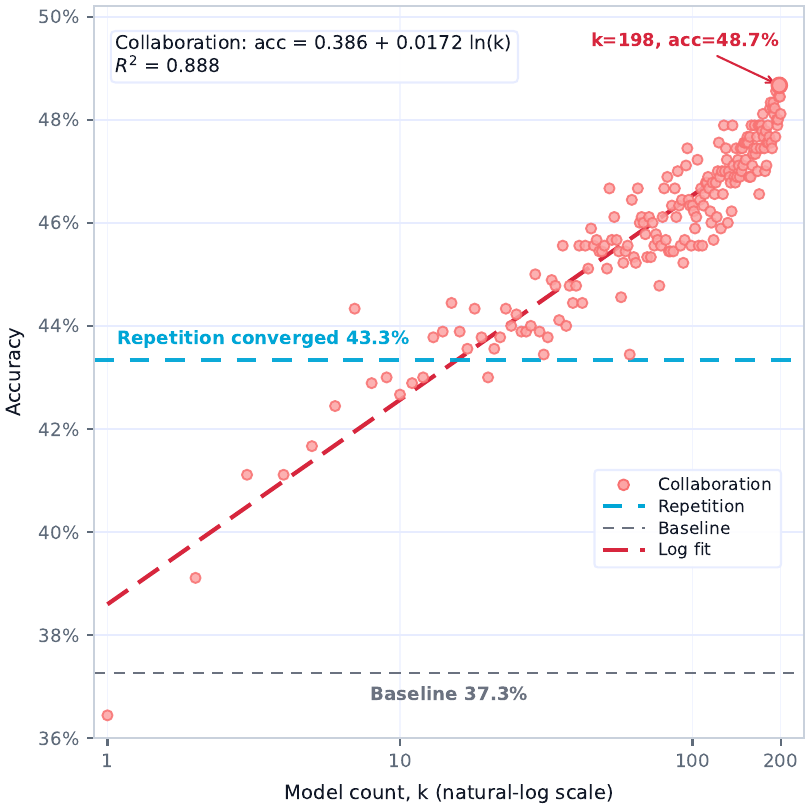}
\caption{Model-count scaling by majority vote. Collaboration across distinct LoRA models improves beyond repeated sampling from one model, and the Collaboration curve is approximately linear in $\ln(k)$ over the measured range.}
\label{fig:model-count-majority-vote}
\end{figure}

The main result is shown in \Cref{fig:model-count-majority-vote}: Collaboration improves steadily as \(k\) increases. Accuracy rises from \(0.3644\) at \(k=1\) to \(0.4267\) at \(k=10\), \(0.4633\) at \(k=100\), and a best observed \(0.4867\) at \(k=198\). Relative to the final baseline accuracy of \(0.3727\), the largest gain is about \(+0.1140\). Repetition improves early but saturates sooner, reaching a best accuracy of \(0.4378\) at \(k=24\). At large \(k\), the Collaboration advantage reaches about \(+0.0533\).
Fitting the Collaboration curve against \(\ln(k)\) gives
\begin{equation}
 \mathrm{accuracy} \approx 0.386 + 0.0172 \ln(k),
\end{equation}
with \(R^2 \approx 0.888\) over the observed range. The discovered behavior is not linear in \(k\); it is approximately linear in \(\ln(k)\) (shown in \Cref{fig:model-count-majority-vote}). Adding more collaborating models gives diminishing but predictable returns. We treat this as an empirical law in one controlled regime, not as a universal theorem of model populations. Its importance is that it makes model count a measurable scale-out variable: accuracy can be studied as a function of the number of distinct LoRA-adapted models.

% 非 universal theorem，而是 controlled regime 下的 empirical law。
% \paragraph{An empirical log scaling law.}

% 介绍diversity的好处，为什么性能会更高，而不是噪音
\paragraph{Why diversity is not sampling noise.}

The comparison between Collaboration and Repetition shows that adapter diversity is not equivalent to sampling noise. Repeated sampling from one model helps at small \(k\), because stochastic decoding produces varied answers, but it saturates earlier. Voting across different LoRA instances continues to improve at larger \(k\), suggesting that the population aggregates complementary policies. Importantly, this diversity does not come from unrelated architectures or pretraining corpora. It comes from small PEFT runs following different trajectories under data ordering, masking, stochasticity, and checkpoint timing. Even this modest, cheaply constructed adapter diversity is useful.

% 收尾总结、从一个模型到很多模型，介绍使用lora的好处，再提出future research direction
\paragraph{From one model to a distribution of models.}
Operationally, this experiment would be difficult without PEFT. Training 200 full checkpoints, serving them, extracting answers, and evaluating many random subsets would be expensive and cumbersome. LoRA turns each model into a lightweight variant of the same prior, making controlled population experiments feasible. This is the concrete mechanism behind Scale Out: \textbf{once adapter creation and serving are cheap, researchers can optimize not only a model but a distribution of models.}

The observed log law is not a universal theorem of model populations, but it establishes a research object: accuracy as a function of model count. Future systems may route among adapters, vote across them, cluster them by experience, distill from successful subpopulations, or feed aggregate lessons back into a shared prior. This is why the scale-out thesis emphasizes populations of personal models rather than a single increasingly personalized assistant. Individual adaptation creates value, but populations create a second-order resource: diversity of histories, skills, failures, and successes. The resulting system is not one universal assistant with ever more context, but an ecology of persistent, partially specialized agents.

\section{\pony{Infrastructure for PEFT Populations}}\label{sec:infrastructure}

\subsection{\pony{Why the Three Axes Need a Systems Layer}}\label{subsec:infra-three-axes}

\pony{The three scaling axes define a practical architecture only if the adapted state can survive as an operational object. Without a systems layer, each axis fails in a characteristic way. Scale Up without lifecycle management produces a strong prior that can be adapted once but not repeatedly. Each LoRA RL run becomes an isolated event, adapter state is lost between runs, and the trained behavior cannot be reliably transferred to the serving runtime that will actually deploy it. Scale Down without mobility management produces a small adapter that is efficient during training but checkpoint-centric during deployment. Every variant requires a full merged model artifact, and the population scales with base-model copies rather than with local adaptive state. Scale Out without addressability and residency control produces a large catalog of adapter files that cannot be selectively loaded, evicted, evaluated, or rolled back. Model count grows, but adapted identities do not persist.}

\pony{PEFT makes individuality compact; managing compact individuality requires a lifecycle like the one instantiated by MinT \citep{mindlab2026mint}. Its mechanisms and measurements illustrate what the three axes require in practice. MinT keeps expensive dense or MoE base models resident and treats LoRA adapters as behavior-carrying policy state. It does not make PEFT important by itself. Rather, it supplies an implementation example of the lifecycle that the three axes require. Large-prior rollout and training remain semantically connected. Small adapters move as exported revisions instead of full checkpoints. Many policy revisions remain addressable while only a bounded working set occupies local cache or GPU batch slots.}

\pony{The unit managed by such infrastructure is not a single file. A continuing adapter-based policy includes adapter tensors, optimizer state, rollout records, evaluation results, and serving placement. These facts change at different time scales. A trainer may hold mutable optimizer state, a sampler may need a fixed adapter revision, a serving actor may evict adapter bytes without deleting the policy, and an evaluation score must name the revision that produced it. Population-scale personalization therefore needs a systems layer because scale is not just the number of adapters stored on disk. It is the number of adapted identities that can be resumed, scored, selected, served, and governed over time.}

\subsection{\pony{Policy Identity: From Adapter Weights to Adapter Revisions}}\label{subsec:infra-policy-identity}

\pony{The adapter boundary becomes a system object when MinT names it as a policy record and exports fixed adapter revisions from it.} Raw adapter weights are not enough: they do not say which base model they attach to, which rank and target modules define their shape, which rollout records generated the latest update, which exported version was evaluated, or where the adapter currently resides. MinT separates these concerns through policy records and adapter revisions.

\begin{table}[ht]
\centering
\small
\setlength{\tabcolsep}{6pt}
\renewcommand{\arraystretch}{1.20}
\caption{MinT separates policy identity from temporary compute residency.}
\label{tab:mint-policy-state}
\fittowidth{%
\begin{tabular}{@{}M{2.6cm}M{5.4cm}M{5.4cm}@{}}
\toprule
\apphead State object & What it stores & Why it matters for PEFT scaling \\
\midrule
\appkey{Base deployment}    & A resident dense or MoE foundation model. & Keeps the expensive shared prior loaded while many adapters change around it. \\
\addlinespace[2pt]
\appkey{Policy record}      & Base version, LoRA rank, target modules, checkpoints, rollout records, and exported revisions. & Makes an adapted behavior resumable, auditable, and rollbackable. \\
\addlinespace[2pt]
\appkey{Policy session}     & A temporary trainer restoration with adapter tensors, optimizer moments, scheduler position, gradients, and rollout metadata. & Allows time-sliced multi-LoRA training without duplicating the base model. \\
\addlinespace[2pt]
\appkey{Adapter revision}   & A fixed exported PEFT adapter in serving tensor layout. & Defines the behavior selected by rollout, evaluation, serving, and rollback. \\
\addlinespace[2pt]
\appkey{Serving residency}  & Whether an adapter is in shared storage, CPU cache, or a GPU batch slot. & Lets a large catalog remain addressable while only a bounded working set is active. \\
\bottomrule
\end{tabular}%
}
\end{table}

A policy record is the durable identity of one adapted behavior over a compatible base. It records the base version, adapter shape, training checkpoint state, rollout records, and exported adapter revisions. A policy session is a temporary restoration of that record on a trainer. An adapter revision is a fixed serving/evaluation object in the layout expected by the sampler. Serving residency is only a placement fact: the revision may be active in a GPU batch, warm in a CPU cache, or only present in shared storage. This distinction turns a personal model from an anonymous LoRA file into a recoverable and auditable policy instance.

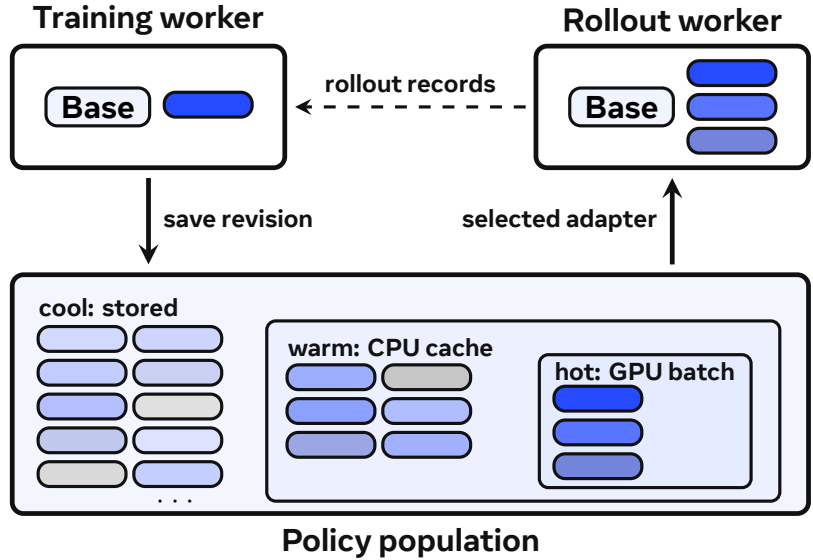
\begin{figure}[t]
\centering
\resizebox{0.72\textwidth}{!}{\begin{tikzpicture}[x=1cm,y=1cm,>=stealth,font=\small\sffamily]
  \colorlet{loraone}{mindlabblue!85}
  \colorlet{loratwo}{mindlabblue!68}
  \colorlet{lorathree}{mindlabblue!52}
  \colorlet{lorafour}{mindlabline!78}
  \colorlet{lorafive}{mindlabline!60}
  \colorlet{lorasix}{mindlabblue!38}
  \colorlet{loraseven}{mindlabline!42}
  \colorlet{loraeight}{mindlabfg!32}
  \colorlet{coldtier}{mindlabbluepale!35}
  \colorlet{warmtier}{mindlabbluepale!55}
  \colorlet{hottier}{mindlabbluepale!75}
  \colorlet{coolone}{mindlabblue!18}
  \colorlet{cooltwo}{mindlabblue!24}
  \colorlet{coolthree}{mindlabblue!30}
  \colorlet{coolfour}{mindlabline!25}
  \colorlet{coolfive}{mindlabfg!16}
  \colorlet{warmone}{mindlabblue!38}
  \colorlet{warmtwo}{mindlabblue!45}
  \colorlet{warmthree}{mindlabline!40}
  \colorlet{warmfour}{mindlabfg!24}
  \colorlet{hotone}{mindlabblue!85}
  \colorlet{hottwo}{mindlabblue!65}
  \colorlet{hotthree}{mindlabline!56}
  \tikzstyle{frame}=[draw=mindlabfg, very thick, rounded corners=1mm, fill=white]
  \tikzstyle{tier}=[draw=mindlabfg, thick, rounded corners=0.8mm, fill=white]
  \tikzstyle{base}=[draw=mindlabfg, thick, rounded corners=1mm, fill=mindlabbluepale!45, align=center, inner sep=2pt, minimum width=0.92cm, minimum height=0.30cm]
  \tikzstyle{adapter}=[draw=mindlabfg, thick, rounded corners=1mm, minimum width=0.78cm, minimum height=0.22cm, inner sep=0pt]
  \tikzstyle{arrow}=[->, very thick, draw=mindlabfg, shorten >=2pt, shorten <=2pt]
  \tikzstyle{returnarrow}=[->, thick, dashed, draw=mindlabfg, shorten >=2pt, shorten <=2pt]
  \tikzstyle{title}=[font=\bfseries\small\sffamily, text=mindlabfg, align=center]
  \tikzstyle{label}=[font=\scriptsize\sffamily, text=mindlabfg, align=center, fill=white, inner sep=1.2pt]
  \tikzstyle{smalllabel}=[font=\scriptsize\sffamily, text=mindlabfg, align=center, fill=white, inner sep=1pt]
  \tikzstyle{plainlabel}=[font=\scriptsize\sffamily, text=mindlabfg, align=center, inner sep=1pt]
  \tikzstyle{tierlabel}=[font=\scriptsize\sffamily, text=mindlabfg, align=center, inner sep=0pt]

  \path[use as bounding box] (1.05,-3.18) rectangle (8.95,2.18);

  \node[frame, minimum width=2.42cm, minimum height=1.08cm] (trainer) at (2.95,1.00) {};
  \node[title] at (2.95,1.77) {Training worker};
  \node[base] at (2.50,1.00) {Base};
  \node[adapter, fill=hotone] at (3.48,1.02) {};

  \node[frame, minimum width=2.42cm, minimum height=1.08cm] (rollout) at (7.60,1.00) {};
  \node[title] at (7.60,1.77) {Rollout worker};
  \node[base] at (7.15,1.00) {Base};
  \node[adapter, fill=hotone] at (8.13,1.30) {};
  \node[adapter, fill=hottwo] at (8.13,1.00) {};
  \node[adapter, fill=hotthree] at (8.13,0.70) {};

  \node[frame, fill=coldtier, minimum width=7.07cm, minimum height=2.12cm] (population) at (5.275,-1.55) {};
  \node[tierlabel, anchor=west] at (1.97,-0.78) {cool: stored};
  \foreach \x/\y/\c in {
	    2.36/-1.06/coolone,2.36/-1.36/cooltwo,2.36/-1.66/coolthree,2.36/-1.96/coolfour,2.36/-2.26/coolfive,
	    3.21/-1.06/cooltwo!82,3.21/-1.36/coolfour!86,3.21/-1.66/coolfive!82,3.21/-1.96/coolone!74,3.21/-2.26/coolthree!76
  } {
    \node[adapter, fill=\c, minimum width=0.78cm] at (\x,\y) {};
  }
  \node[plainlabel] at (3.21,-2.50) {$\cdots$};

  \node[tier, fill=warmtier, minimum width=4.555cm, minimum height=1.60cm] (host) at (6.2775,-1.70) {};
  \node[tierlabel, anchor=west] at (4.18,-1.13) {warm: CPU cache};
  \foreach \x/\y/\c in {
	    4.57/-1.40/warmone,4.57/-1.70/warmtwo,4.57/-2.00/warmthree,
	    5.42/-1.40/warmfour,5.42/-1.70/warmone!82,5.42/-2.00/warmtwo!82
  } {
    \node[adapter, fill=\c, minimum width=0.78cm] at (\x,\y) {};
  }

  \node[tier, fill=hottier, minimum width=1.88cm, minimum height=1.18cm] (gpu) at (7.36,-1.79) {};
  \node[tierlabel, anchor=west] at (6.55,-1.35) {hot: GPU batch};
  \node[adapter, fill=hotone] at (6.94,-1.59) {};
  \node[adapter, fill=hottwo] at (6.94,-1.89) {};
  \node[adapter, fill=hotthree] at (6.94,-2.19) {};
  \node[title] at (5.275,-2.88) {Policy population};

  \draw[arrow] (trainer.south) -- node[label, right, pos=0.50, xshift=2pt] {save revision} (trainer.south |- population.north);
  \draw[arrow] (population.north -| rollout.south) -- node[label, left, pos=0.50, xshift=-2pt] {selected adapter} (rollout.south);
  \draw[returnarrow] (rollout.west) -- node[label, above, yshift=2pt] {rollout records} (trainer.east);
\end{tikzpicture}}
\caption{MinT policy lifecycle. Training updates adapter state over a resident base, export saves fixed adapter revisions into the policy population, and rollout or serving selects revisions through bounded residency tiers.}
\label{fig:mint-policy-lifecycle}
\end{figure}

This identity layer connects directly to Scale Out. A population of personal models must preserve differences across time, not merely instantiate many prompts or adapters once. Memories, skills, preferences, and permissions need a place to accumulate, but they also need revision boundaries so that behavior can be evaluated, served, rolled back, or retired. The adapter remains the local adaptive state, and the policy record is the system object that lets that state persist as one member of a larger population.

\subsection{\pony{Scale Up Requires Computation Provenance}}\label{subsec:infra-scale-up-provenance}

Scale Up depends on keeping a strong prior available to repeated LoRA RL without turning every run into a separate full-model deployment. In MinT, the base deployment is the resident object: dense and MoE bases stay loaded across model-parallel trainer groups, rollout engines, and serving actors, while policy records restore only the adapter and training state needed for the selected policy. This is the systems form of the Scale Up argument. A stronger prior increases adapter leverage only if rollout, scoring, export, and serving continue to refer to the same policy over the same compatible base.

Large-prior LoRA RL also needs computation provenance. In an MoE model, selected expert ids determine part of the computation that produced each rollout token. If training-time scoring routes that token through different experts, the update no longer scores the sampled policy. MinT records route information when the backend exposes it and replays selected expert paths when the training layout can map them. If route ids are missing or unmappable, the corresponding token is removed from the replayed policy-gradient term rather than treated as equivalent.

Dynamic sparse attention introduces another provenance boundary. In GLM-5-style DSA, the indexer and top-$k$ path determine which tokens enter sparse attention. MinT fixes implementation mismatches where the cause is exposed, including indexer RoPE layout, normalized query/key inputs, deterministic top-$k$ behavior, frozen indexer defaults, long-context THD/CP support, and LoRA loading for DSA target modules. When probability mismatch remains, IcePop-style correction masks token-level ratios outside the trusted band. This does not reconstruct every DSA indexer decision, but it preserves the failure signal by excluding unsafe scoring terms from the update.

The relevant Scale Up property is semantic continuity, not only GPU capacity. The policy that generated trajectories, the policy whose probabilities are scored during training, and the policy later served as an adapter revision must remain the same adapted behavior over a compatible base. Router replay, DSA correction, distributed export, and policy-record resolution are the MinT mechanisms that keep a large prior usable as a repeated adaptation substrate rather than as a static checkpoint.

\begin{figure}[t]
\centering
\includegraphics[width=0.82\textwidth]{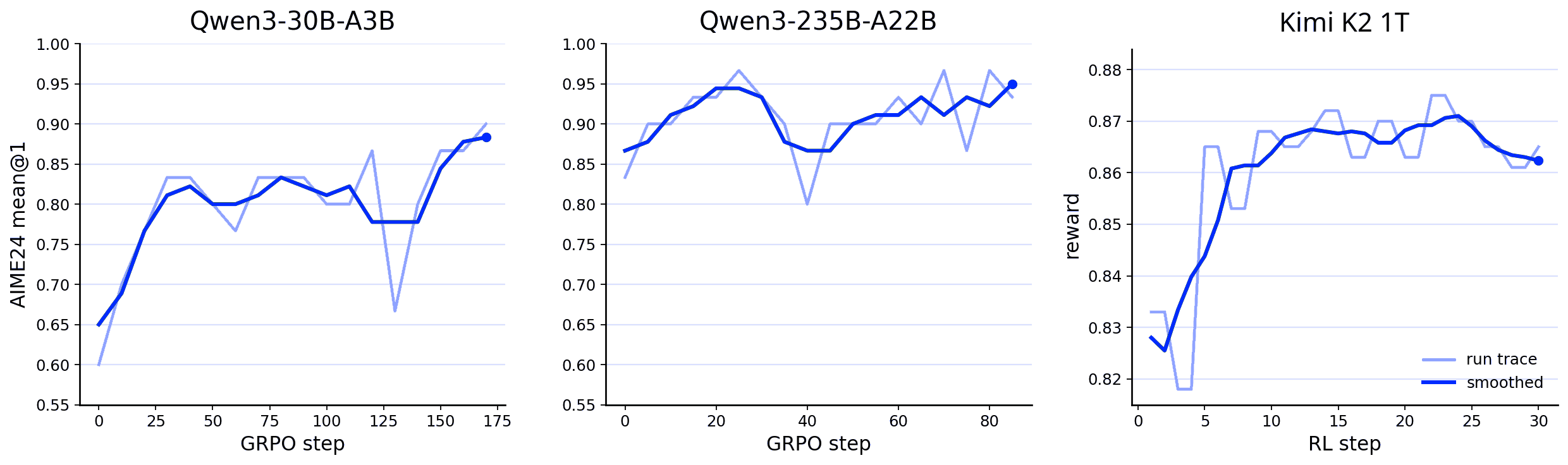}
\caption{MoE LoRA RL curves from the MinT evaluation. The 30B and 235B panels use AIME24 mean@1; the Kimi K2 panel reports the end-to-end LoRA RL reward curve for a 1T-class countdown-task run.}
\label{fig:mint-moe-scale-curves}
\end{figure}

The large-model evidence exercises this lifecycle on dense and sparse paths, including Qwen3-235B-A22B GRPO and a Kimi K2 1T-class countdown-task LoRA RL path over a 1.04T-parameter MoE with 32.6B active parameters. The evidence boundary is specific: these runs show that large-prior LoRA RL can be made operational when adapter state, rollout records, route metadata, export layout, and serving revision are managed as one policy lifecycle. They do not by themselves prove a universal frontier-model scaling law. They show the infrastructure condition under which the Scale Up axis can be tested.

\subsection{\pony{Scale Down Requires Adapter-Only Mobility}}\label{subsec:infra-scale-down-mobility}

Scale Down asks whether the adaptive state can remain small, stable, and reusable across many updates. MinT turns that algorithmic requirement into a handoff boundary: the exported adapter revision, not a merged full checkpoint, is the artifact that moves from trainer to sampler. If every trained adapter has to be merged into a full checkpoint before serving, the update may be parameter-efficient during training but checkpoint-centric during deployment. The population would then scale with repeated base-model artifacts rather than with local adaptive state.

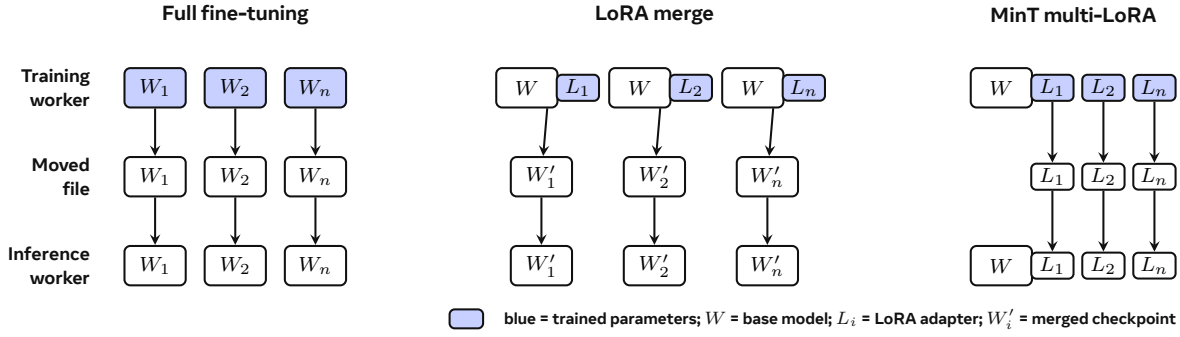
\begin{figure}[t]
\centering
\resizebox{0.96\textwidth}{!}{\begin{tikzpicture}[x=1cm,y=1cm,>=stealth,font=\footnotesize\sffamily]
  \tikzstyle{title}=[font=\bfseries\small\sffamily, text=mindlabfg, align=center]
  \tikzstyle{row}=[font=\bfseries\footnotesize\sffamily, text=mindlabfg, align=right]
  \tikzstyle{block}=[draw=mindlabfg, thick, rounded corners=0.8mm, fill=white, align=center, inner sep=2pt, minimum width=0.84cm, minimum height=0.54cm]
  \tikzstyle{trainblock}=[block, fill=mindlabblue!22]
  \tikzstyle{adapter}=[draw=mindlabfg, thick, rounded corners=0.7mm, fill=white, align=center, inner sep=1.8pt, minimum width=0.58cm, minimum height=0.34cm]
  \tikzstyle{trainadapter}=[adapter, fill=mindlabblue!22]
  \tikzstyle{note}=[font=\scriptsize\sffamily, text=mindlabfg, align=center]
  \tikzstyle{arrow}=[->, thick, draw=mindlabfg]

  \path[use as bounding box] (-1.45,-4.34) rectangle (14.95,0.42);

  \node[title] at (1.75,0.10) {Full fine-tuning};
  \node[title] at (7.50,0.10) {LoRA merge};
  \node[title] at (13.25,0.10) {MinT multi-LoRA};

  \node[row, anchor=east] at (-0.12,-0.90) {Training\\worker};
  \node[row, anchor=east] at (-0.12,-2.12) {Moved\\file};
  \node[row, anchor=east] at (-0.12,-3.34) {Inference\\worker};

  \foreach \x/\i in {0.65/1,1.75/2,2.85/n} {
    \node[trainblock] (ftw\i) at (\x,-0.90) {$W_{\i}$};
    \node[block] (ftc\i) at (\x,-2.12) {$W_{\i}$};
    \node[block] (fts\i) at (\x,-3.34) {$W_{\i}$};
    \draw[arrow] (ftw\i) -- (ftc\i);
    \draw[arrow] (ftc\i) -- (fts\i);
  }

  \foreach \x/\i in {5.95/1,7.50/2,9.05/n} {
    \node[block] (mb\i) at (\x-0.20,-0.90) {$W$};
    \node[trainadapter] (ml\i) at (\x+0.50,-0.90) {$L_{\i}$};
    \node[block] (mc\i) at (\x,-2.12) {$W'_{\i}$};
    \node[block] (ms\i) at (\x,-3.34) {$W'_{\i}$};
    \coordinate (join\i) at (\x+0.08,-1.18);
    \draw[arrow] (join\i) -- (mc\i);
    \draw[arrow] (mc\i) -- (ms\i);
  }
  \node[block] (mtw) at (12.26,-0.90) {$W$};
  \foreach \x/\i in {12.96/1,13.66/2,14.36/n} {
    \node[trainadapter] (mtl\i) at (\x,-0.90) {$L_{\i}$};
    \node[adapter] (mal\i) at (\x,-2.12) {$L_{\i}$};
    \draw[arrow] (mtl\i) -- (mal\i);
  }

  \node[block] (miw) at (12.26,-3.34) {$W$};
  \foreach \x/\i in {12.96/1,13.66/2,14.36/n} {
    \node[adapter] (msl\i) at (\x,-3.34) {$L_{\i}$};
    \draw[arrow] (mal\i) -- (msl\i);
  }

  \node[trainadapter, minimum width=0.46cm, minimum height=0.26cm] at (4.92,-4.07) {};
  \node[note, anchor=east] at (14.78,-4.07) {blue = trained parameters; $W$ = base model; $L_i$ = LoRA adapter; $W'_i$ = merged checkpoint};
\end{tikzpicture}}
\caption{Training-to-serving artifacts under full fine-tuning, merge-based LoRA, and MinT. Full fine-tuning and merge-based LoRA move full model checkpoints for each variant. MinT moves exported LoRA adapter revisions to an inference engine that already holds the compatible base.}
\label{fig:mint-handoff-paths}
\end{figure}

The exported revision carries adapter tensors, rank, target modules, tensor layout, and base compatibility. Optimizer state, accumulated gradients, and rank-local training files remain on the training side. Evaluation and serving select the fixed revision, so a score or deployed behavior is attributed to a concrete adapter state rather than to an implicit worker-local checkpoint. This is the system-level counterpart of the low-rank and initialization results in Scale Down: reducing trainable state matters most when the reduced state is also the object that can be moved, evaluated, served, and rolled back.

\begin{table}[ht]
\centering
\scriptsize
\setlength{\tabcolsep}{3pt}
\renewcommand{\arraystretch}{1.08}
\caption{Adapter-only handoff keeps the moved artifact much smaller than a merged or full checkpoint while preserving the sampling path used for evaluation. Cold first sample is the first request wall time; total sample speed includes that first request, while warm speed excludes it.}
\label{tab:mint-handoff}
\fittowidth{%
\begin{tabular}{@{}M{1.8cm}M{1.6cm}M{2.3cm}M{1.8cm}M{2.4cm}M{2.0cm}M{2.5cm}@{}}
\toprule
\apphead Model & Path & Moved artifact & File size & Materialization or load & Cold first sample & \makecell{sample speed\\total/warm} \\
\midrule
\appgroup{7}{Qwen3-4B}
\appkey{Qwen3-4B}  & Adapter & rank-32 LoRA & 252 MiB  & 0.036\,s & 4.114\,s & 15.568/15.567 tok/s \\
\addlinespace[2pt]
\appkey{Qwen3-4B}  & Merge   & full model   & 8.061 GB & 71.820\,s & 55.704\,s & 4.697/20.595 tok/s \\
\addlinespace[2pt]
\appgroup{7}{Qwen3-30B}
\appkey{Qwen3-30B} & Adapter & rank-16 LoRA & 1.692 GB & 46.455\,s & 117.304\,s & 1.874/5.700 tok/s \\
\addlinespace[2pt]
\appkey{Qwen3-30B} & Merge   & full model   & 61.084 GB & 402.245\,s & 156.074\,s & 1.573/6.904 tok/s \\
\bottomrule
\end{tabular}%
}
\end{table}

The MinT handoff measurements illustrate why this boundary matters for the Scale Down axis. On Qwen3-4B, a rank-32 adapter is 252 MiB, while the merged full checkpoint is 8.061 GB. On Qwen3-30B, a rank-16 adapter is 1.692 GB, while the merged full checkpoint is 61.084 GB. These numbers do not define a universal adapter ratio, because rank, target modules, dtype, and tensor layout change the size. They show the systems invariant: the crossing artifact can remain the local adaptive state rather than a full copy of the shared prior. In MinT, Scale Down is therefore not just a smaller training update. It is a smaller object that can move across training, rollout, evaluation, serving, and rollback without re-materializing the base model.

MinT also supports Scale Down through time-sliced training over resident bases. A policy session restores only the selected adapter tensors, optimizer moments, scheduler position, accumulated gradients, and rollout records. The base remains loaded while different policies take turns on compatible trainer workers. This does not make each individual update mathematically smaller, but it reduces the system cost of maintaining many small updates: repeated learning becomes a schedulable service operation rather than a sequence of isolated full-checkpoint jobs.

\subsection{\pony{Scale Out Requires Bounded Residency}}\label{subsec:infra-scale-out-residency}

Scale Out asks what changes when the number of adapted instances becomes a scaling variable. MinT supports this axis by separating addressability from residency. A million personal models does not mean that one inference engine keeps a million adapters in GPU memory. It means that a large population of adapter revisions can be named, selected, loaded, served, evicted, and later recovered while each serving actor keeps only a bounded local working set. Multi-tenant LoRA serving systems \citep{sheng2023slora,chen2023punica,zhou2025dynamic} establish the batching and operator-level machinery for sharing one base model across many adapters; the MinT-specific question is how to extend that view into a lifecycle of named, persistent policy revisions.

\begin{figure}[t]
\centering
\resizebox{0.92\textwidth}{!}{\begin{tikzpicture}[x=1cm,y=1cm,>=stealth,font=\footnotesize\sffamily]
  \tikzstyle{actor}=[draw=mindlabfg, very thick, rounded corners=1mm, fill=white]
  \tikzstyle{box}=[draw=mindlabfg, thick, rounded corners=0.8mm, fill=mindlabbluepale!45, align=center, inner sep=4pt]
  \tikzstyle{store}=[draw=mindlabfg, thick, rounded corners=0.8mm, fill=white, align=center, inner sep=4pt]
  \tikzstyle{slotbox}=[draw=mindlabfg, thick, rounded corners=0.8mm, fill=white, align=center, inner sep=3pt]
  \tikzstyle{adapter}=[draw=mindlabfg, thick, rounded corners=1mm, align=center, minimum width=0.86cm, minimum height=0.23cm, inner sep=0pt, fill=mindlabfg!8]
  \tikzstyle{selected}=[draw=mindlabfg, thick, rounded corners=1mm, align=center, font=\scriptsize\sffamily, minimum width=1.14cm, minimum height=0.30cm, inner sep=1.2pt, fill=mindlabblue!38]
  \tikzstyle{arrow}=[->, very thick, draw=mindlabfg, shorten >=2pt, shorten <=2pt]
  \tikzstyle{hot}=[->, very thick, draw=hotpath, shorten >=2pt, shorten <=2pt]
  \tikzstyle{cold}=[->, thick, dashed, draw=coldpath, shorten >=2pt, shorten <=2pt]
  \tikzstyle{label}=[font=\footnotesize\sffamily, align=center, fill=white, inner sep=1.5pt, text=mindlabfg]
  \tikzstyle{smalllabel}=[font=\scriptsize\sffamily, align=center, fill=white, inner sep=1.1pt, text=mindlabfg]
  \tikzstyle{title}=[font=\bfseries\footnotesize\sffamily, align=center, text=mindlabfg]
  \colorlet{hotpath}{mindlabblue}
  \colorlet{coldpath}{mindlabline}

  \path[use as bounding box] (0.20,-2.42) rectangle (14.05,2.18);

  \node[box, minimum width=1.92cm, minimum height=0.72cm] (map) at (2.92,0.40) {Serving map\\policy $\to r$};

  \node[actor, minimum width=7.02cm, minimum height=3.70cm] (actor) at (8.53,-0.18) {};
  \node[title] at (8.53,1.44) {vLLM sampler actor};

  \node[slotbox, minimum width=1.86cm, minimum height=1.28cm] (gpu) at (6.64,0.40) {};
  \node[smalllabel, anchor=south] at (6.64,1.09) {GPU LoRA slots};
  \node[adapter] at (6.64,0.72) {};
  \node[selected] (selected) at (6.64,0.40) {$L_r$};
  \node[adapter] at (6.64,0.08) {};

  \node[box, minimum width=2.10cm, minimum height=0.76cm] (infer) at (10.30,0.40)
    {vLLM\\generation};
  \node[store, minimum width=2.10cm, minimum height=0.78cm] (base) at (10.30,-1.35)
    {resident\\model};
  \node[label, anchor=west] (tokens) at (12.54,0.40) {response tokens};

  \node[store, minimum width=1.86cm, minimum height=0.78cm] (warm) at (6.64,-1.35)
    {CPU adapter\\cache};
  \node[store, minimum width=2.50cm, minimum height=0.78cm] (catalog) at (2.63,-1.35)
    {shared adapter\\storage};

  \draw[arrow] (0.62,0.40) -- node[smalllabel, above, pos=0.42] {request} (map.west);
  \draw[hot] (map.east) -- node[smalllabel, above, yshift=3pt, pos=0.50, text=hotpath] {hit: $L_r$ resident} (gpu.west);
  \draw[hot] (gpu.east) -- node[smalllabel, above, yshift=3pt, pos=0.50, text=hotpath] {selected $L_r$} (infer.west);
  \draw[arrow] (base.north) -- (infer.south);
  \draw[arrow] (infer.east) -- (tokens.west);

  \draw[cold] (catalog.east) -- node[smalllabel, above, pos=0.50, text=coldpath] {fetch $r$} (warm.west);
  \draw[cold] (warm.north) -- node[smalllabel, right, xshift=6pt, pos=0.44, text=coldpath] {admit $L_r$} (gpu.south);

  \draw[hot, -] (12.42,-1.09) -- (12.88,-1.09);
  \node[smalllabel, anchor=west, text=hotpath] at (12.94,-1.09) {hot path};
  \draw[cold, -, dashed] (12.42,-1.61) -- (12.88,-1.61);
  \node[smalllabel, anchor=west, text=coldpath] at (12.94,-1.61) {cold path};
\end{tikzpicture}}
\caption{Shared-base multi-LoRA serving in MinT. A request resolves to an exported adapter revision. The hot path uses an adapter already in a GPU slot, while the cold path fetches the revision from shared storage into the CPU cache and then admits it into a GPU batch slot before decoding.}
\label{fig:mint-serving-blocks}
\end{figure}
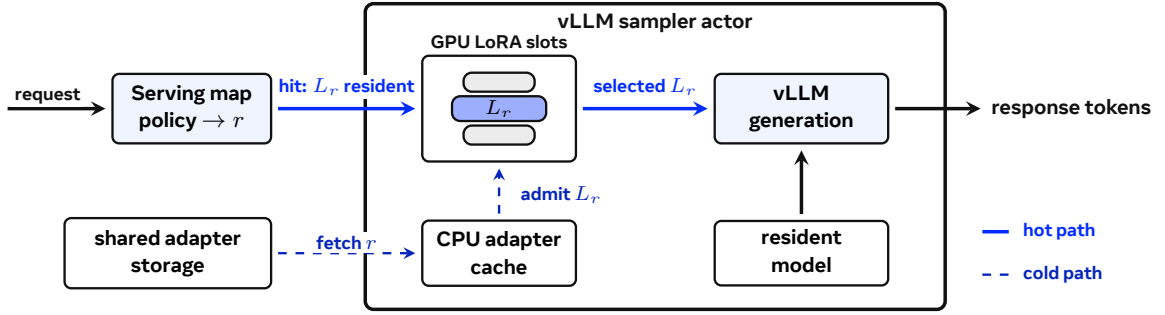

MinT separates serving state into three tiers. The durable catalog names adapter revisions in shared storage. The CPU cache holds adapter bytes near one serving actor. The GPU batch contains the smaller set of adapters active in the current decoding step. A request may hit the GPU batch, promote a CPU-cached adapter, or enter a cold-load path from shared storage. The policy identity is stable across all three placements, and only residency changes.

\begin{table}[ht]
\centering
\small
\setlength{\tabcolsep}{6pt}
\renewcommand{\arraystretch}{1.20}
\caption{Policy-population serving separates addressability from live residency.}
\label{tab:mint-serving-bounds}
\fittowidth{%
\begin{tabular}{@{}M{2.6cm}M{5.6cm}M{5.2cm}@{}}
\toprule
\apphead Resource tier & Evidence or bound & Interpretation \\
\midrule
\appkey{Addressable catalog}     & Built and audited a $10^6$-entry packed adapter catalog; serving experiments select bounded working sets from it. & Catalog size is an addressability scale, not simultaneous GPU residency. \\
\addlinespace[2pt]
\appkey{CPU adapter cache}       & 369 loaded adapters at a 512-adapter hotset; 550 loaded adapters under 2048-adapter weak-locality pressure. & Local CPU memory absorbs recurring traffic before requests touch shared storage. \\
\addlinespace[2pt]
\appkey{GPU batch slots}         & 64 distinct adapters in the tested same-batch window. & Batch execution has the smallest adapter-diversity window. \\
\addlinespace[2pt]
\appkey{Cold loading}            & 16 distinct cache misses form a 1.375--23.267\,s load staircase. & Different missing policies must be admitted as explicit service work. \\
\addlinespace[2pt]
\appkey{Packed MoE LoRA loading} & 37,248 tensor objects reduced to 672; live loading becomes 8.5--8.7$\times$ faster. & Representation controls cold-load overhead even when adapter bytes are modest. \\
\addlinespace[2pt]
\appkey{Readiness gate} & Newly registered adapters become user-visible only after activation/prewarm. & Registration and serving readiness are separate lifecycle states. \\
\bottomrule
\end{tabular}%
}
\end{table}

The MinT serving measurements preserve the distinction between addressability, local residency, and active GPU use. The catalog evidence is an addressability result: MinT builds and audits a $10^6$-entry packed adapter catalog. The local-residency evidence is smaller by design: on one Qwen3-30B rank-1 serving actor, repeated-adapter traffic reaches hundreds of CPU-cached adapters, while the tested same-batch adapter window is 64 distinct adapters. Weak-locality traffic and broad rollout waves expose the cold path: 16 distinct cache misses form a 1.375--23.267 second load staircase, while repeated requests for the same missing policy can share one load. These numbers are the evidence for the residency boundary that Scale Out needs: policy count can be large only if catalog lookup, CPU cache, GPU batch slots, and cold activation are treated as different service scales.

\begin{table}[ht]
\centering
\scriptsize
\setlength{\tabcolsep}{3.2pt}
\renewcommand{\arraystretch}{1.10}
\caption{Hot-reload and readiness measurements on the MinT serving path. Admission protects old warm tenants by moving cold activation into a scheduled path; two-phase readiness exposes new adapters only after activation, so zero load time applies to ready-path user requests rather than registration time.}
\label{tab:mint-serving-readiness}
\fittowidth{%
\begin{tabular}{@{}M{3.0cm}M{3.4cm}M{4.6cm}M{3.4cm}@{}}
\toprule
\apphead Policy & Existing warm traffic & New-adapter path & Interpretation \\
\midrule
\appgroup{4}{Expose before readiness}
\appkey{Admission off}
& post TTFT p95 24.03\,s; $>$20\,s stalls: 10
& e2e p95 59.18\,s; user TTFT p95 22.19\,s; load p95 47.37\,s
& Fast exposure, but cold first-touch disrupts warm tenants. \\
\addlinespace[3pt]
\appkey{Admission on}
& post TTFT p95 9.71\,s; $>$20\,s stalls: 0
& e2e p95 314.79\,s; user TTFT p95 10.68\,s; load p95 294.96\,s
& Admission protects warm tenants, but new users wait behind activation. \\
\addlinespace[3pt]
\appgroup{4}{Expose after readiness}
\appkey{Two-phase readiness}
& post TTFT p95 9.63\,s; $>$20\,s stalls: 0
& ready-path TTFT p95 4.60\,s; load p95 0.00\,s; prewarm span 409.04\,s
& First user requests arrive after activation, so they do not load adapters. \\
\bottomrule
\end{tabular}%
}
\end{table}

Readiness is the operational control that prevents Scale Out from degrading existing warm users. Immediate exposure lets new adapters become selectable quickly, but cold first-touch disrupts warm tenants: without admission, existing warm traffic reaches 24.03 s post-reload TTFT p95 and records 10 stalls above 20 s. Admission removes those stalls but shifts waiting to new cold requests. Two-phase readiness changes the user-visible contract: the adapter is registered and prewarmed first, then exposed after activation. In the measured row, old warm TTFT p95 stays at 9.63 s with no stalls above 20 s, and the first ready-path request to the new adapter has 0.00 s load p95 and 4.60 s TTFT p95 after a 409.04 s prewarm span.

\begin{table}[ht]
\centering
\scriptsize
\setlength{\tabcolsep}{4pt}
\renewcommand{\arraystretch}{1.10}
\caption{Packed MoE LoRA loading reduces cold-load overhead by removing tensor fanout. The byte-size change is small; the speedup comes from replacing many tiny tensor objects with a compact serving representation.}
\label{tab:mint-packed-loader}
\fittowidth{%
\begin{tabular}{@{}M{4.8cm}M{2.4cm}M{2.4cm}M{3.6cm}@{}}
\toprule
\apphead Metric & Original & Packed & Effect \\
\midrule
\appgroup{4}{Adapter-file shape}
\appkey{File size} & 110.75 MB & 105.58 MB & 1.05$\times$ smaller \\
\addlinespace[2pt]
\appkey{Tensor objects} & 37{,}248 & 672 & 55.4$\times$ fewer \\
\appgroup{4}{Cold-load slice}
\appkey{Read tensors} & 0.3669\,s & 0.0067\,s & 54.8$\times$ faster \\
\addlinespace[2pt]
\appkey{Build loader objects} & 0.7540\,s & 0.0256\,s & 29.5$\times$ faster \\
\appgroup{4}{Live engine loading}
\appkey{$N{=}4$ live load} & 1.363\,s & 0.156\,s & 8.7$\times$ faster \\
\addlinespace[2pt]
\appkey{$N{=}8$ live load} & 1.361\,s & 0.159\,s & 8.6$\times$ faster \\
\addlinespace[2pt]
\appkey{$N{=}16$ live load} & 1.388\,s & 0.164\,s & 8.5$\times$ faster \\
\bottomrule
\end{tabular}%
}
\end{table}

Adapter representation is the second Scale Out control surfaced by the MinT measurements because the cold path can be object-bound rather than byte-bound. In the measured MoE rank-1 setting, the raw adapter is moderate in bytes but fragmented into 37,248 tensor objects. Packing reduces this to 672 tensor objects with nearly unchanged bytes, improves the direct loader slices by 29.5--54.8$\times$, and makes live engine loading 8.5--8.7$\times$ faster with packed medians below 0.2 seconds. The important conclusion is therefore not that cold loads are slow in one probe. It is that policy count creates several service scales that must be controlled separately: catalog registration names durable revisions, routing preserves locality, CPU cache absorbs recurring policies, GPU batch slots bound active diversity, cold activation is scheduled work, and readiness decides when a newly registered adapter becomes user-visible. Scale Out is a controlled lifecycle for many policy revisions, not a promise that every revision is simultaneously resident.

\subsection{\pony{The Lifecycle of a Personal Model}}\label{subsec:infra-personal-lifecycle}

The personal-model thesis becomes concrete when experience has a durable path into local adaptive state. A user interaction, tool outcome, evaluation trace, or social-simulation event produces records that can be scored or distilled. Training updates adapter state over a shared prior. Export freezes a revision. Serving selects that revision under bounded residency. Later experience resumes from the policy record rather than from an anonymous adapter file. The individual model is therefore not a full checkpoint, but a continuing adaptive identity over a shared base.

This lifecycle ties MinT back to the preceding Scale Out section. LoRA-as-memory needs a writeable local state. Context Learning needs a path from context-time improvement to future query-only behavior. Skill internalization needs repeated updates from tool outcomes and failures. Personalization as a security boundary needs permission and behavior history to remain attached to the policy that will later act. All of these require identity, provenance, mobility, and residency control.

The resulting architecture is the one implied by the title of the paper. A few trillion-scale priors provide shared competence. Many adapter revisions provide local memory, skill, preference, and policy state. MinT keeps those revisions identifiable and movable while serving only a bounded active set at any moment. It is the concrete infrastructure example that connects the three axes: Scale Up through resident large-prior LoRA RL with computation provenance, Scale Down through adapter-only mobility and time-sliced policy sessions, and Scale Out through addressable catalogs with bounded residency.

\section{Conclusion}\label{sec:conclusion}

\pony{The phrase ``million personal models of trillion parameters'' should not be read as a claim that each user owns and trains a separate trillion-parameter checkpoint. The intended architecture is different. A small number of strong trillion-scale base models provide shared capability, while millions of lightweight adapters provide persistent local adaptive state. The base model carries general reasoning, world knowledge, language competence, and tool-use priors. The adapter carries part of the learned consequences of repeated experience, such as memories, preferences, skills, and policies.}

\pony{This architecture depends on all three scaling axes at once. Scale Up makes the shared base model worth adapting. Scale Down makes each update cheap and stable enough to repeat. Scale Out turns repeated updates into persistent populations. Removing any axis breaks the thesis. Weak base models limit what adapters can learn. Expensive adapters prevent continuous adaptation. Without multiplicity, PEFT remains a local optimization rather than a path toward population-scale personalization.}

\pony{Several research problems remain open, organized by the three axes:}

\begin{enumerate}
    \item \pony{\textbf{Scale Up (RL-native PEFT theory)}: rank, scaling, initialization, KL drift, and off-policiness need a unified account that connects adapter geometry to training stability at large prior scale.}
    \item \pony{\textbf{Scale Up (Tiny-adapter reliability)}: low-rank LoRA already shows signal on strong priors, but must become stable under seed, batch, and task variation before it can serve as a repeatable adaptation substrate.}
    \item \pony{\textbf{Scale Down (Stateful adapter design)}: memory-oriented mechanisms such as $\delta$-mem need controlled comparisons against standard LoRA, retrieval, and full-context baselines to establish when a stateful adapter is the right representation.}
    \item \pony{\textbf{Scale Down (Signal efficiency)}: Context Learning needs benchmarks that measure how much durable improvement is extracted from real interaction traces, and how that improvement degrades or compounds over repeated updates.}
    \item \pony{\textbf{Scale Out (LoRA memory and skill capacity)}: the bounded capacity law for parametric memory observed on simple benchmarks needs to be extended to RL-trained, multi-task adapters, where the open question is which experience deserves to become durable adapter state and which should remain in retrieval or context.}
    \item \pony{\textbf{Scale Out (Persistent user simulation)}: per-user adapters resolve the collapse of prompt-based personas in small populations, but it remains open whether large adapter populations can preserve heterogeneity over long-horizon interaction without drifting back to a base-model average.}
    \item \pony{\textbf{Scale Out (Population as computation)}: aggregation gains from distinct LoRA models suggest a research object beyond per-model accuracy, namely how routing, voting, debate, and distillation across adapter populations scale with model count and adapter diversity.}
\end{enumerate}

\pony{\textbf{Limitations.} The evidence in this paper points to a direction, not a deployed system. Most experiments are run at the scale of controlled benchmarks and simulations, while large-scale empirical validation on our own personal-model deployments remains limited. The bottleneck at this point is compute capacity, not missing methodology. We believe this is the right direction, and that its claims will sharpen as more real-interaction evidence accumulates at scale.}

\pony{The final view is a population architecture rather than one universal assistant with ever more context and ever more centralized control. PEFT makes it possible to scale from one shared foundation model to many persistent personal model instances: first serving individuals, then supporting user simulation, and eventually making diversity among adapted models a source of collective intelligence and creativity. That is the broader reason PEFT matters. It makes adaptation efficient, and through that efficiency it makes persistent individuality scalable.}

\bmhead{Acknowledgements}

We thank all members of Mind Lab for the discussions, experiments, engineering support, writing feedback, and project context that shaped this manuscript. Names are listed alphabetically within each group.

\paragraph{Core Contributors.}
\begin{sloppypar}
Andrew Chen, Steven Chiang, Kyrie Lei, Kieran Liu, Pony Ma, Vincent Wang, Josh Ying, Di Zhang, Ruijia Zhang, Adrian Zhou, Yuhua Zhou.
\end{sloppypar}

\paragraph{Team.}
\begin{sloppypar}
Vin Bo, Song Cao, Vic Cao, Andrew Chen, Kaijie Chen, Cleon Cheng, Steven Chiang, Kaixuan Fan, Hera Feng, Huan Feng, Arthur Fu, Jun Gao, Hongquan Gu, Aaron Guan, Nolan Ho, Mutian Hong, Hailee Hou, Peixuan Hua, Charles Huang, Miles Jiang, Nora Jiang, Yuyi Jiang, Qiuyu Jin, Fancy Kong, Andrew Lei, Kyrie Lei, Alexy Li, Lucian Li, Ray Li, Theo Li, Wenhao Li, Zhihui Li, Allen Lin, Jiayi Lin, Kairus Liu, Kieran Liu, Logan Liu, Xiang Liu, Irvine Lu, Maeve Luo, Runze Lv, Pony Ma, Verity Niu, Anson Qiu, Vincent Wang, Rio Yang, Maxwell Yao, Carrie Ye, Regis Ye, Wenlin Ye, Josh Ying, Danney Zeng, Yuhan Zhan, Anya Zhang, Di Zhang, Ruijia Zhang, Shiyang Zhang, Sueky Zhang, Ya Zhang, Wei Zhao, Ada Zhou, Adrian Zhou, Yuhua Zhou, Xinyue Zhu, Murphy Zhuang.
\end{sloppypar}

\bibliographystyle{plainnat}
\bibliography{references}

\end{document}